\documentclass{article}
\usepackage{jfrExamplee}
\usepackage{graphicx}
\usepackage{apalike}
\usepackage{mathtools}
\usepackage{setspace}

\usepackage[printonlyused]{acronym}
\renewcommand{\vec}[1]{\boldsymbol{#1}}   
\newcommand{\mat}[1]{\boldsymbol{#1}}     
\newcommand{\T}{^{\mathsf{T}}}            
\newcommand{\unit}[1]{\,\mathrm{#1}}      
\newcommand{\unitbold}[1]{\,\mathbf{#1}}  

\acrodef{AscTec}{Ascending Technologies}
\acrodef{ASL}{Autonomous Systems Lab}
\acrodef{DGPS}{Differential Global Positioning System}
\acrodef{DOF}{Degree of Freedom}
\acrodef{EKF}{Extended Kalman Filter}
\acrodef{ENU}{East-North-Up}
\acrodef{ETH}{ETH Zürich}
\acrodef{GUI}{Graphical User Interface}
\acrodef{GPS}{Global Positioning System}
\acrodef{ICP}{Iterative Closest Point}
\acrodef{ICT}{Individual Challenge Trial}
\acrodef{GCT}{Grand Challenge Trial}
\acrodef{IMU}{Inertial Measurement Unit}
\acrodef{Lidar}{Light Detection and Ranging}
\acrodef{MAV}{Micro Aerial Vehicle}
\acrodef{MBZIRC}{Mohamed Bin Zayed International Robotics Challenge}
\acrodef{MSF}{Multi-Sensor Fusion}
\acrodef{RANSAC}{Random Sample Consensus}
\acrodef{ROS}{Robot Operating System}
\acrodef{ROVIO}{Robust Visual Inertial Odometry}
\acrodef{RTK}{Real-Time Kinematic}
\acrodef{SIL}{Software-in-the-Loop}
\acrodef{UAV}{Unmanned Aerial Vehicle}
\acrodef{UWB}{Ultra-Wideband}
\acrodef{SDR}{Software Defined Radio}
\acrodef{UDP}{User Datagram Protocol}
\acrodef{VI-Sensor}{Visual-Inertial Sensor}

\acrodef{CPP}{Coverage Path Planning}
\acrodef{DZ}{Dropping Zone}
\acrodef{EPM}{Electropermanent Magnet}
\acrodef{euRathlon}{European Robotics League}
\acrodef{EuRoC}{European Robotics Challenges}
\acrodef{FSM}{Finite State Machine}
\acrodef{IPP}{Informative Path Planning}
\acrodef{FoV}{Field of View}
\acrodef{LZ}{Landing Zone}
\acrodef{KF}{Kalman Filter}
\acrodef{MCU}{Microcontroller Unit}
\acrodef{NMPC}{Nonlinear Model Predictive Controller}
\acrodef{MSF}{Multi Sensor Fusion}
\acrodef{PCB}{Printed Circuit Board}
\acrodef{PWM}{Pulse-Width Modulation}
\acrodef{RMSE}{root-mean-square error}
\acrodef{SAR}{search and rescue}
\acrodef{SDK}{Software Development Kit}
\acrodef{VIO}{Visual-Inertial Odometry}
\acrodef{VS}{Visual Servoing}
\acrodef{UGV}{Unmanned Ground Vehicle}
\usepackage{siunitx} 
\usepackage[inline,shortlabels]{enumitem} 
\usepackage{subcaption} 
\usepackage{hyperref} 
\usepackage{amssymb} 
\usepackage{booktabs} 
\newcommand{\ra}[1]{\renewcommand{\arraystretch}{#1}}
\title{The ETH-MAV Team in the MBZ International Robotics Challenge}

\author{
Rik B\"ahnemann\thanks{The authors contributed equally to this work. Their names are listed in alphabetical order.} \\
\And
Michael Pantic\footnotemark[1]{} \\
\And
Marija Popovi\'c\footnotemark[1]{} \\
\And
Dominik Schindler\footnotemark[1]{} \\
\And
Marco Tranzatto\footnotemark[1]{} \\
\And
Mina Kamel \\
\And
Marius Grimm \\
\And
Jakob Widauer \\
\And
Roland Siegwart \\
\And
Juan Nieto \\
Autonomous Systems Lab (ASL)\\
ETH Zurich - Swiss Federal Institute of Technology \\
Zurich, Switzerland \\
Corresponding author: \texttt{brik@ethz.ch} \\
}

\begin{document}

\maketitle

\begin{abstract}
This article describes the hardware and software systems of the \ac{MAV} platforms used by the ETH Zurich team in the 2017 Mohamed Bin Zayed International Robotics Challenge (MBZIRC).
The aim was to develop robust outdoor platforms with the autonomous capabilities required for the competition, by applying and integrating knowledge from various fields, including computer vision, sensor fusion, optimal control, and probabilistic robotics.
This paper presents the major components and structures of the system architectures, and reports on experimental findings for the MAV-based challenges in the competition.
Main highlights include securing second place both in the individual search, pick, and place task of Challenge 3 and the Grand Challenge,
with autonomous landing executed in less than one minute and a visual servoing success rate of over $90\%$ for object pickups.
\end{abstract}

\acresetall

\section*{Supplementary Material}
For a supplementary video see: \url{https://youtu.be/DXYFAkjHeho}. \\
For open-source components visit: \url{https://github.com/ethz-asl}.

\section{Introduction}
The \ac{MBZIRC} is a biennial competition aiming to demonstrate the state-of-the-art in applied robotics and inspire its future.
The inaugurating event took place on March 16-18, 2017 at the Yas Marina Circuit in Abu Dhabi, UAE, with total prize and sponsorship money of USD 5M and 25 participating teams.
The competition consisted of three individual challenges and a triathlon-type Grand Challenge combining all three in a single event.
Challenge 1 required an \ac{MAV} to locate, track, and land on a moving vehicle.
Challenge 2 required a \ac{UGV} to locate and reach a panel, and operate a valve stem on it.
Challenge 3 required a collaborative team of \acp{MAV} to locate, track, pick, and deliver a set of static and moving objects on a field.
The Grand Challenge required the \acp{MAV} and \ac{UGV} to complete all three challenges simultaneously.

This paper details the hardware and software systems of the \acp{MAV} used by our team for the relevant competition tasks (Challenges 1 and 3, and the Grand Challenge).
The developments, driven by a diverse team of researchers from the \ac{ASL}, sought to further multi-agent autonomous aerial systems for outdoor applications.
Ultimately, we achieved second place in both Challenge 3 and the Grand Challenge.
A full description of the approach of our team for Challenge 2 is described in \cite{mbzirc_ugv_rsl}.

The main challenge we encountered was building robust systems comprising the different individual functionalities required for the aforementioned tasks.
This led to the development of two complex system pipelines,
advancing the applicability of outdoor \acp{MAV} on both the level of stand-alone modules as well as complete system integration.
Methods were implemented and interfaced in the areas of precise state estimation, accurate position control, agent allocation, object detection and tracking, and object gripping, with respect to the competition requirements.
The key elements of our algorithms leverage concepts from various areas, including computer vision, sensor fusion, optimal control, and probabilistic robotics.
This paper is a systems article describing the software and hardware architectures we designed for the competition.
Its main contributions are a detailed report on the development of our infrastructure
and a discussion of our experimental findings in context of the \ac{MBZIRC}.
We hope that our experiences provide valuable insights into outdoor robotics applications
and benefit future competing teams.

This paper is organized as follows: our \ac{MAV} platforms are introduced in Section \ref{sec:platforms}, while state estimation and control methods are presented in Sections \ref{sec:state} and \ref{sec:control}, respectively.
These elements are common to both challenges.
Sections \ref{sec:ch1} and \ref{sec:ch3} detail our approaches to each challenge.
Section \ref{sec:preparation} sketches the development progress.
We present results obtained from our working systems in Section \ref{sec:results} before concluding in Section \ref{sec:conclusion}.

\section{Platforms}
\label{sec:platforms}
\begin{figure}
  \centering
  \subcaptionbox{The \ac{AscTec} Firefly hexacopter shortly before landing on the moving target (Challenge 1).
  For the landing task it has a downward-facing monocular camera and \ac{Lidar}, and custom landing gear.\label{fig:platform_firefly}}{\includegraphics{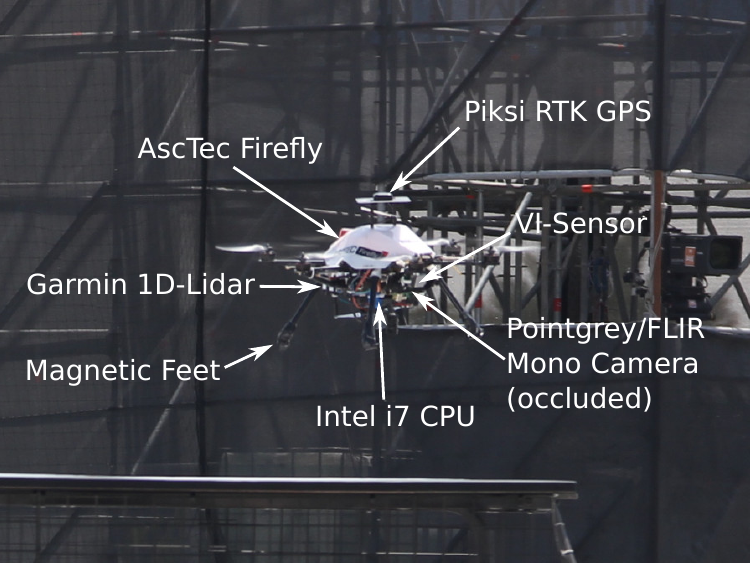}}\hfill%
  \subcaptionbox{
  One of three identical \ac{AscTec} Neo hexacopters grasping a moving object (Challenge 3).
  For aerial gripping each drone has a high resolution color camera and an \ac{EPM} gripper with Hall sensors.\label{fig:platform_neo}}{\includegraphics{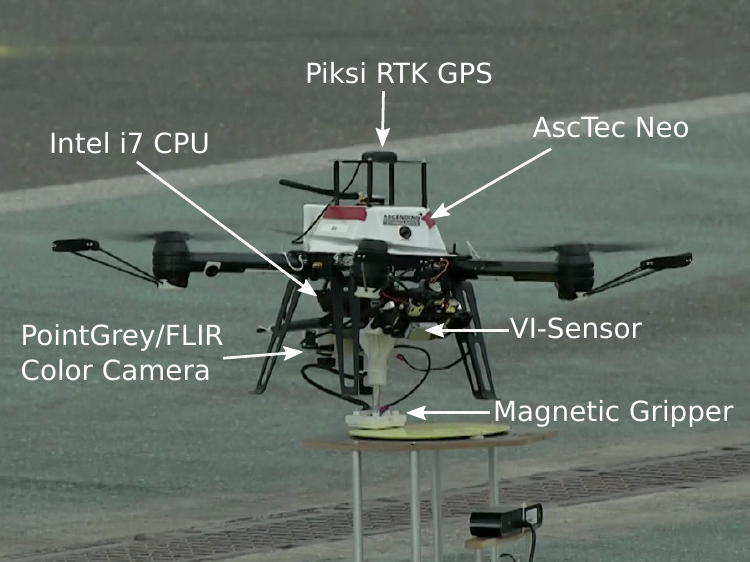}}%

\caption{The \ac{MAV} platforms used by the \ac{ETH} team.
All \acp{MAV} have a \ac{VI-Sensor} and \ac{RTK} \ac{GPS} for state estimation.
Computation is done on-board using an Intel i7 CPU running Ubuntu and \ac{ROS}.}
\label{fig:ch1_ch3_platforms}
\end{figure}

Our hardware decisions were led by the need to create synergies between Challenge 1 and Challenge 3, as well as among previous projects in our group.
Given our previous experience with and the availability of \ac{AscTec} multirotor platforms, we decided to use those for both challenges (\autoref{fig:ch1_ch3_platforms}).
In Challenge 1, we used an \ac{AscTec} Firefly hexacopter with an \ac{AscTec} AutoPilot (\autoref{fig:platform_firefly}).
In Challenge 3, we used three \ac{AscTec} Neo hexacopters with \ac{AscTec} Trinity flight controllers (\autoref{fig:platform_neo}).
Both autopilots provide access to on-board filtered \ac{IMU} and magnetometer data, as well as attitude control.
An integrated safety switch allows for taking over remote attitude control in cases where autonomous algorithms fail.
The grippers have integrated Hall sensors for contact detection, as described in detail in \autoref{sec:ch3_gripper}.

All platforms are equipped with a \emph{perception unit} which uses similar hardware for state estimation and operating the software stack.
This unit is built around an on-board computer with Intel i7 processor running Ubuntu and \ac{ROS} as middleware to exchange messages between different modules.
The core sensors for localization are the on-board \ac{IMU}, a slightly downward-facing \ac{VI-Sensor}, and a Piksi \ac{RTK} \ac{GPS} receiver.
The \ac{VI-Sensor} was developed by the \ac{ASL} and Skybotix AG \cite{nikolic2014synchronized} to obtain fully time-synchronized and factory calibrated \ac{IMU} and stereo-camera datastreams, and a forward-facing monocular camera for \ac{VIO}.
Piksi hardware version V2 is used as the \ac{RTK} receiver \cite{piksi_datasheet_v2}. \ac{RTK} \ac{GPS} is able to achieve much higher positioning precision by mitigating the sources of error typically affecting stand-alone \ac{GPS}, and can reach an accuracy of a few centimeters.

Challenge-specific sensors, on the other hand, are designed to fulfill individual task specifications and thus differ for both platform types.
In Challenge 1, the landing platform is detected with a downward-facing PointGrey/FLIR Chameleon USB 2.0 monocular camera with $1.3 \unit{MP}$ and a fisheye lens \cite{chameleon_mono_camera}.
The distance to the platform is detected with a Garmin LIDAR-Lite V3 \ac{Lidar} sensor \cite{lidar_garmin}.
Moreover, additional commercial landing gear is integrated in the existing platform.
For Challenge 3, we equip each \ac{MAV} with a downward-facing PointGrey/FLIR $3.2 \unit{MP}$ Chameleon USB 3.0 color camera to detect objects \cite{chameleon_color_camera}.
For object gripping, the \acp{MAV} are equipped with modified NicaDrone OpenGrab \ac{EPM} grippers \cite{gripper_datasheet_v3}.

\section{State Estimation}
\label{sec:state}
\begin{figure}
\centering
\includegraphics[width=0.9\linewidth]{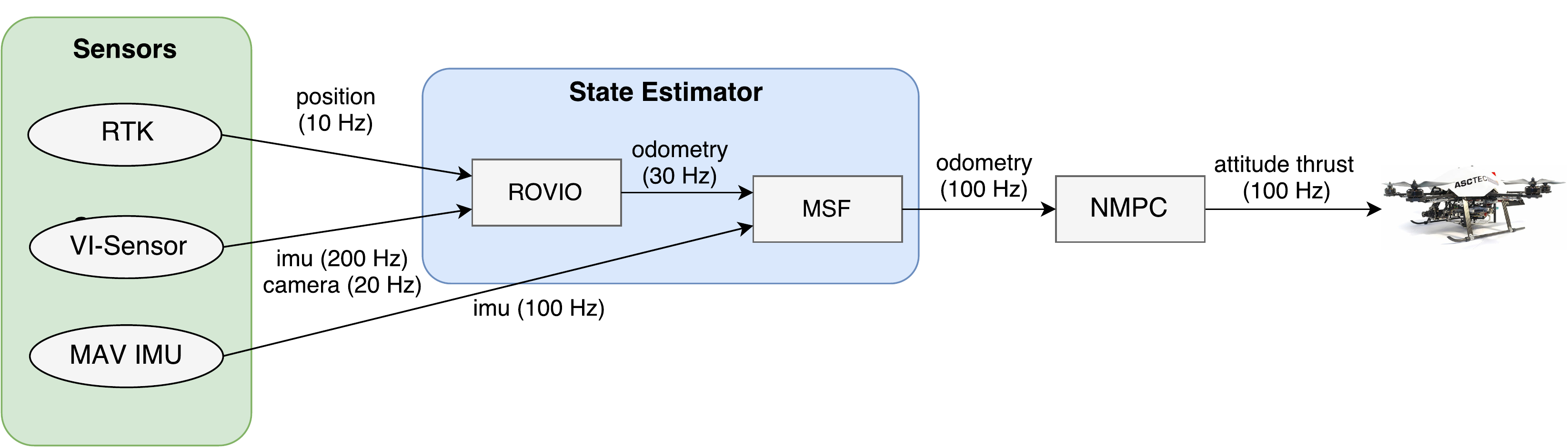}
\caption{The state estimator comprises \acf{ROVIO}, which fuses visual-inertial data with GPS positions, and \acf{MSF} which robustly fuses the pose estimate from \ac{ROVIO} with the on-board \ac{IMU} for control.}
\label{fig:state_estimation_diagram}
\end{figure}

Precise and robust state estimation is a key element for executing the fast and dynamic maneuvers required by both challenges.
This section presents our state estimation pipeline, which consists of a cascade of \acp{EKF}, denoted by the blue box in \autoref{fig:state_estimation_diagram}.
The following subsections detail our architecture and the major design decisions behind it.
The last subsection highlights some insights we attained when integrating \ac{RTK} \ac{GPS}.

\subsection{Conventions and Notations}
In this paper, we utilize the following conventions and notations:
${_A}\vec{p}_{B,C}$ refers to a vector from point $B$ to point $C$ expressed in coordinate frame $A$.
The homogeneous transformation matrix $_A\mat{T}_{B,C} \in {\mathbb{R}}^{4\times4}$ expressed in $A$ converts a homogeneous vector ${_C}\vec{p}_{C,D}$ to the vector ${_A}\vec{p}_{B,D}$, i.e.,
${_A}\vec{p}_{B,D} = {_A}\mat{T}_{B,C} \, {_C}\vec{p}_{C,D}$.
Coordinate frame names are typed lower case italic, e.g., \textit{enu} corresponds to the coordinate frame representing the local tangent plane aligned with \ac{ENU} directions.
\autoref{table_frames_descriptions} presents a complete description of all frames mentioned in the remainder of this work.

\begin{table}[]
\centering
\ra{1.3}
  \begin{tabular}{@{}lllll@{}}
    \toprule
    \textbf{Symbol} & \textbf{Name}            & \textbf{Origin}       	    	& \textbf{Description}                                              &  \\ \midrule
    $A$             & \textit{arena}       & Center of arena       		    & Fixed orientation and origin with respect to \textit{enu} frame  	&  \\
    $B$             & \textit{mav-imu}     & \ac{MAV}-\ac{IMU}     		    & Aligned with \ac{MAV}-\ac{IMU} axes                               &  \\
    $C$             & \textit{camera}      & Focal point camera   		    & $z$-axis pointing out from the lens                        			  &  \\
    $E$             & \textit{enu}         & Center of the arena   		    & Local tangent plane, aligned with \ac{ENU} directions					    &  \\
    $G$             & \textit{gripper}     & Center gripper surface		    & $z$-axis pointing out from gripper surface					              &  \\
    $I$             & \textit{vi-imu}      & \ac{VI-Sensor}-\ac{IMU}      & Aligned with \ac{VI-Sensor} \ac{IMU} axes                         &  \\
    $L$             & \textit{lidar}       & \ac{Lidar} lens             	& $z$-axis pointing out from the lens                        				&  \\
    $O$             & \textit{odom}        & \ac{MAV} starting position   & Aligned with \textit{enu} frame after \ac{ROVIO} reset           &  \\
    $T$             & \textit{target}      & Landing target center        & Platform center estimated by the Challenge 1 tracker              & \\
    \bottomrule
  \end{tabular}
\caption{Description of the frames used.}
\label{table_frames_descriptions}
\end{table}

\subsection{\ac{ROVIO}}
The first block of the pipeline is a monocular \ac{VIO} estimator called \ac{ROVIO} \cite{rovio}.
It achieves accurate tracking performance by leveraging the pixel intensity errors of image patches, which are directly used for tracking multilevel patch features in an underlying \ac{EKF}.
Its software implementation is open-source and publicly available \cite{rovio_github}.
In its standard configuration, \ac{ROVIO} takes as input from the \ac{VI-Sensor} one monocular camera stream synchronized with \ac{IMU} and outputs odometry data.
An odometry message is composed of information about position, orientation, angular, and linear velocities of a certain frame with respect to another.
This filter outputs odometry messages regarding the state of the \ac{VI-Sensor} \ac{IMU} frame, \textit{vi-imu}, with respect to the \textit{odom} frame,
which is gravity-aligned with the origin placed at the \ac{MAV} initialization point.

\subsection{RTK GPS as External Pose in \ac{ROVIO}}
In our configuration, \ac{ROVIO} fuses \ac{VIO} information, which is locally accurate but can diverge in the long-term \cite{scaramuzza_tutorial_vio}, with accurate and precise \ac{RTK} measurements, which arrive at a low frequency but do not suffer from drift \cite{Kaplan2005}.
Our system involves two \ac{RTK} receivers: a base station and at least one \ac{MAV}.
The base station is typically a stationary receiver configured to broadcast \ac{RTK} corrections, often through a radio link.
The \ac{MAV} receiver is configured through the radio pair to receive these corrections from the base station, and applies them to solve for a centimeter-level accurate vector between the units.

\ac{RTK} measurements, expressed as $(\mathit{latitude}, \mathit{longitude}, \mathit{altitude})$, are first converted to a local Cartesian coordinate system and then transmitted to \ac{ROVIO} as external position measurements.
\ac{ENU} is set as the local reference frame \textit{enu} with an arbitrary origin close to the \ac{MAV} initialization position.
This fusion procedure prevents \ac{ROVIO} odometry from drifting, as it is continuously corrected by external \ac{RTK} measurements.
Moreover, the output odometry can be seen as a ``global'' state, since it conveys information about the \textit{vi-imu} frame with respect to the \textit{enu} frame.
This ``global'' property is due to the independent origin and orientation of the \textit{enu} frame with respect to the initial position and orientation of the \ac{MAV}.
In order to successfully integrate \ac{VIO} and \ac{RTK}, the \textit{odom} and \textit{enu} frames must be aligned so that the previous two quantities are expressed in the same coordinate system.
To do so, we exploit orientation data from the on-board \ac{IMU}.

Given the transformations from the \textit{vi-imu} frame $I$ to the \textit{mav-imu} frame $B$ ${_B}\mat{T}_{B,I}$ obtained using the Kalibr framework \cite{kalibr_github}
and the transformation from \textit{mav-imu} frame $B$ to the \textit{enu} frame $E$ ${_E}\mat{T}_{E,B}$ obtained from \ac{RTK} measurements, the aforementioned alignment operation is performed by imposing:
\begin{align}
{_O}\mat{T}_{O,I} = {_E}\mat{T}_{E,B} \, {_B}\mat{T}_{B,I}\texttt{.}
\end{align}
The resulting local \textit{enu} positions, expressed by $(\mathit{east}, \mathit{north}, \mathit{up})$, can be directly used by \ac{ROVIO}. Note that, while the orientation of these two frames is the same, they can have different origins.

\subsection{MSF}
The second block of the pipeline consists of the \ac{MSF} framework \cite{lynen13robust}:
an \ac{EKF} able to process delayed measurements, both relative and absolute, from a theoretically unlimited number of different sensors and sensor types with on-line self-calibration.
Its software implementation is open-source and publicly available \cite{msf_github}.

\ac{MSF} fuses the odometry states, output by \ac{ROVIO} at $\sim 30 \unit{Hz}$,  with the on-board \ac{IMU}, as shown in \autoref{fig:state_estimation_diagram}.
This module improves the estimate of the current state from \ac{ROVIO} by incorporating the inertial information available from the flight control unit.
The final \ac{MSF} output state is a high-frequency odometry message  at $\sim 100 \unit{Hz}$ expressing the transformation, and angular and linear velocities of the \ac{MAV} \ac{IMU} frame, \textit{mav-imu}, with respect to the \textit{enu} frame,
providing a ``global'' state estimate.
This odometry information is then conveyed to the \ac{MAV} position controller.

\subsection{A Drift-Free and Global State Estimation Pipeline}
There were two main motivations behind the cascade configuration (\autoref{fig:state_estimation_diagram}):
(i) obtaining a global estimate of the state, and
(ii) exploiting the duality between the local accuracy of \ac{VIO} and the driftless property of \ac{RTK}.

A global state estimate describes the current status of the \ac{MAV} expressed in a fixed frame.
We use the \textit{enu} frame as a common reference, as the \acp{MAV} fly in the same, rather small, area simultaneously.
This permits sharing odometry information between multiple agents, since they are expressed with respect to a common reference.
This exchange is a core requirement in Challenge 3 and the Grand Challenge, so that each agent can avoid and maintain a minimum safety distance given the position and traveling directions of the others, as detailed in \autoref{sec:control} and \autoref{sec:ch3_multi_agent}.
Moreover, this method enables the \ac{MAV} in Challenge 1 to make additional prior assumptions for platform tracking, as explained in \autoref{subsection_platform_tracking}.

The integration of \ac{ROVIO} and \ac{MSF} with \ac{RTK} \ac{GPS} ensures the robustness of our state estimation pipeline against known issues of the two individual systems.
A pure \ac{VIO} state estimator cascading \ac{ROVIO} and \ac{MSF}, without any external position estimation, would suffer from increasing drift in the local position, but still be accurate within short time frames while providing a high-rate output.
On the other hand, \ac{RTK} \ac{GPS} measurements are sporadic ($\sim 10 \unit{Hz}$), and the position fix can be lost, but the estimated geodetic coordinates have precise $1$ to $5 \unit{cm}$ horizontal position accuracy, $8$ to $15 \unit{cm}$ vertical position accuracy, and do not drift over time \cite{piksi_accuracy}.
This combined solution achieves driftless tracking of the current state, mainly due to \ac{RTK} \ac{GPS}, while providing a robust solution against \ac{RTK} fix loss, since \ac{VIO} alone computes the current status until an external \ac{RTK} position is available again.
As a result, during the entire \ac{MBZIRC}, our \acp{MAV} operated without any localization issues.

\subsection{RTK GPS Integration}
\ac{RTK} \ac{GPS} was integrated on the \acp{MAV} five months before the \ac{MBZIRC}, and evolved during our field trials.
The precision of this system, however, is accompanied by several weaknesses.
Firstly, the \ac{GPS} antenna must be placed as far away as possible from any device and/or cable using USB 3.0 technology.
USB 3.0 devices and cables may interfere with wireless devices operating in the $2.4 \unit{GHz}$ ISM band \cite{intel_paper_usb3_interference}.
Many tests showed a significant drop in the signal-to-noise ratio of \ac{GPS} L1 $1.575 \unit{GHz}$ carrier. A \ac{SDR} receiver was used to identify noise sources.
Simple solutions to this problem were either to remove any USB 3.0 device, to increase the distance of the antenna from any component using USB 3.0, or add additional shielding between antenna and USB 3.0 devices and cables.
Secondly, the throughput of corrections sent from \ac{RTK} base station to the \acp{MAV} must be kept as high as possible.
We addressed this by establishing redundant communication from the base station and the \ac{MAV}:
corrections are sent both over a $5 \unit{GHz}$ WiFi network as \ac{UDP} subnet broadcast as well as over a $2.4 \unit{GHz}$ radio link.
In this way, if either link experienced connectivity issues, the other could still deliver the desired correction messages.
Finally, the time required to gain a \ac{RTK} fix is highly dependent on the number and signal strength of common satellites between the base station and the \acp{MAV}.
Our experience showed that an average number of over eight common satellites, with a signal strength higher than $40 \unit{dB-Hz}$, yields an \ac{RTK} fix in $\sim 10 \unit{min}$\footnote{With the new Piksi Multi GNSS Module, \ac{RTK} fixes are obtained in $3 \unit{min}$ on average.}.

\ac{ASL} released the \ac{ROS} driver used during \ac{MBZIRC}, which is available on-line \cite{piksi_github}.
This on-line repository contains \ac{ROS} drivers for Piksi V2.3 hardware version and for Piksi Multi. Moreover it includes a collection of utilities, such as a \ac{ROS} package that allows fusing \ac{RTK} measurements into \ac{ROVIO}.
The main advantages of our \ac{ROS} driver are: WGS84 coordinates are converted into ENU and output directly from the driver; \ac{RTK} corrections can be sent both over a radio link and Wifi. This creates a redundant link for streaming corrections, which improves the robustness of the system.

\section{Reactive and Adaptive Trajectory Tracking Control}
\label{sec:control}
In order for the \acp{MAV} to perform complex tasks,
such as landing on fast-moving platforms, precisely picking up small objects, and transporting loads,
robust, high-bandwidth trajectory tracking control is crucial.
Moreover, in the challenges,
up to four \acp{MAV}, a \ac{UGV}, and several static obstacles are required to share a common workspace,
which demands an additional safety layer for dynamic collision avoidance.

Our proposed trajectory tracking solution is based on a standard cascaded control scheme where a slower outer trajectory tracking control loop generates attitude and thrust references for a faster inner attitude control loop \cite{achtelik2011onboard}.
The \ac{AscTec} autopilot provides an adaptive and reliable attitude controller.
For trajectory control, we use a \ac{NMPC} developed at our lab \cite{control_github,kamel2016linear,kamel2017nonlinear}.

\begin{figure}[!h]
  \centering
  \includegraphics[width=0.75\linewidth]{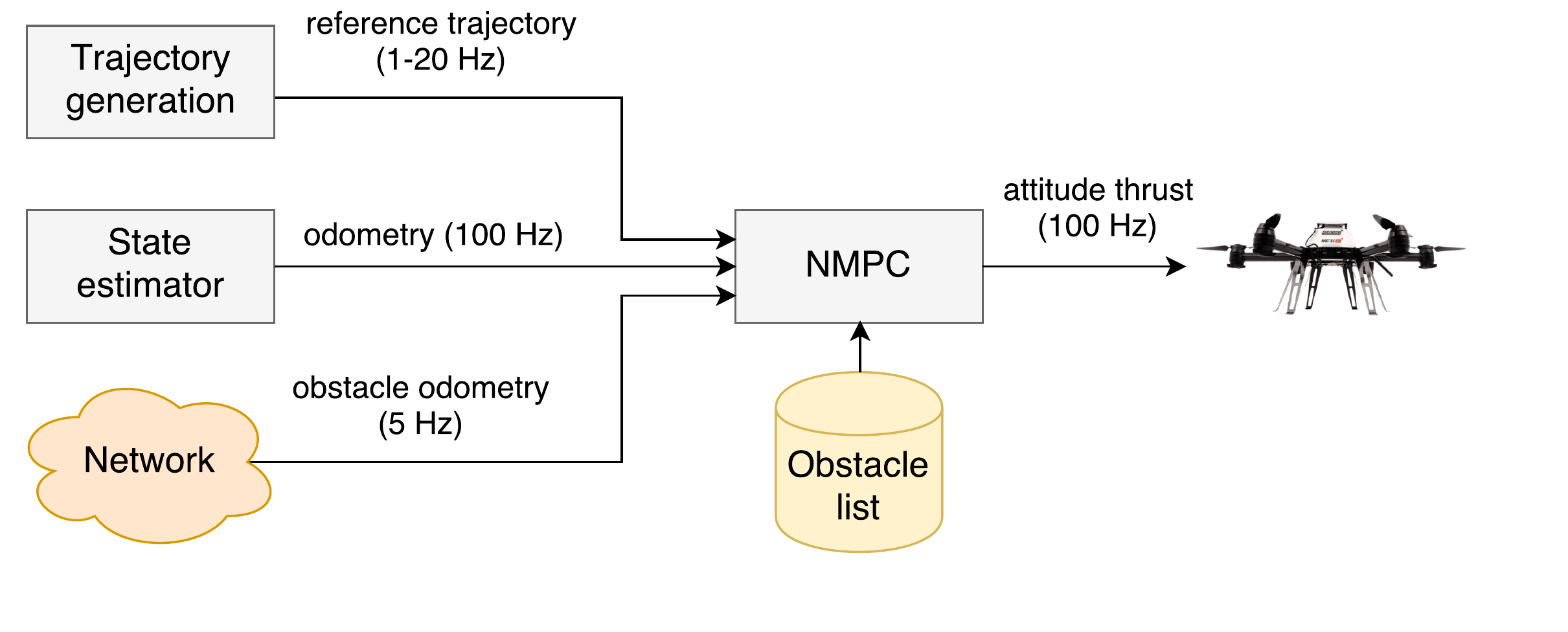}
  \caption{The trajectory control flow used in both challenges. The controller allows trajectory tracking as well as reactive collision avoidance. If an obstacle is listed in the loaded obstacle list and its state is broadcasted over the network, the \ac{NMPC} guarantees a minimum distance to the object.}
  \label{fig:control_pipeline}
\end{figure}

The trajectory tracking controller provides three functionalities:
(i) it tracks the reference trajectories generated by the challenge submodules, (ii) it compensates for changes in mass and wind with an \ac{EKF} disturbance observer, and (iii) it uses the agents' global odometry information for reactive collision avoidance.
For the latter, the controller includes obstacles into its trajectory tracking optimization that guarantees safety distances between the agents \cite{kamel2017nonlinear}.
The signal flow is depicted in \autoref{fig:control_pipeline}.

The obstacle avoidance feature and our global state estimation together serve as a safety layer in Challenge 3 and the Grand Challenge, where we program the \acp{MAV} to avoid all known obstacles and agents.
This paradigm ensures maintaining a minimum distance to an obstacle, provided its odometry and name are transmitted to the controller.
For this purpose, each \ac{MAV} broadcasts its global odometry over the wireless network,
and a ground station relays the drop box position.
We throttle the messages to $5 \unit{Hz}$ to reduce network traffic.
Essentially, this is the only inter-robot communication on the network.

\begin{figure}
\centering
\includegraphics{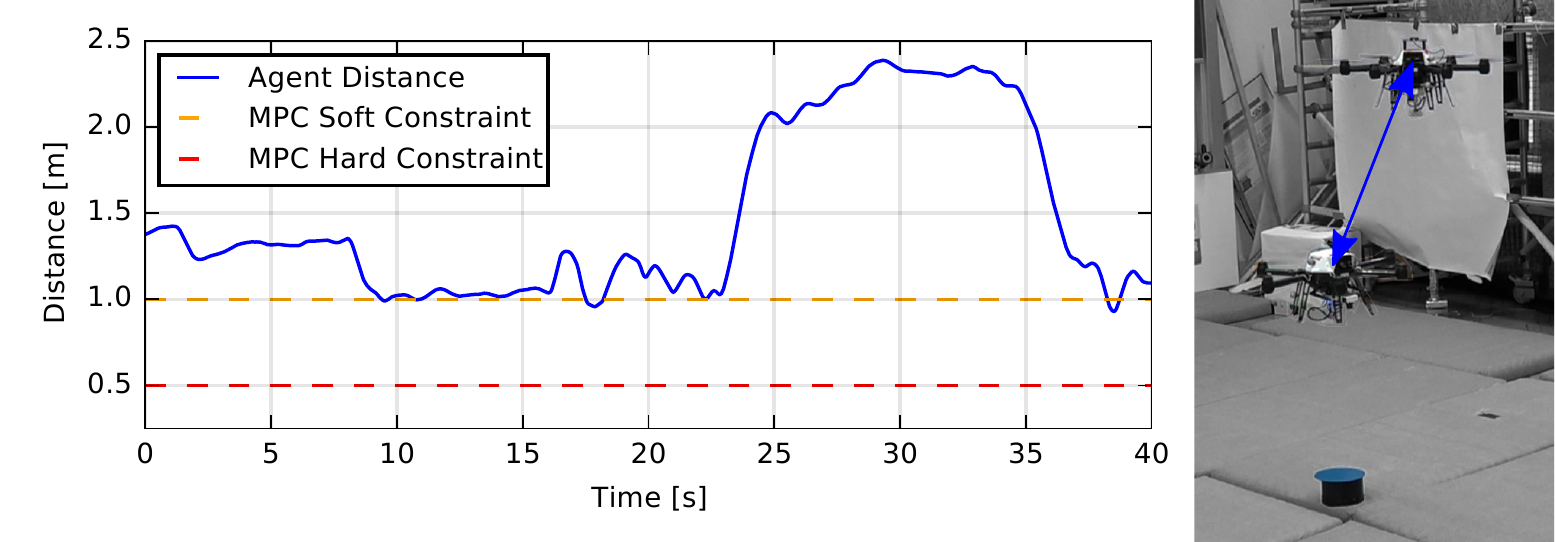}
\caption{Two \acp{MAV} attempting to pick up the same object in a motion capture environment.
Sharing global position over the wireless network allows the \ac{NMPC} to prevent collisions without explicitly communicating the \ac{MAV} intentions.}
\label{fig:collision}
\end{figure}

\autoref{fig:collision} shows an example scenario where two \acp{MAV} attempt to pick up the same object.
The \ac{MAV} further away from the desired object perceives the other vehicle as an obstacle by receiving its odometry over wireless.
Even though the \acp{MAV} do not broadcast their intentions, the controller can maintain a minimum distance between them.

\section{Challenge 1: Landing on a Moving Platform}
\label{sec:ch1}
\begin{figure}
  \centering
  \includegraphics[width=\textwidth]{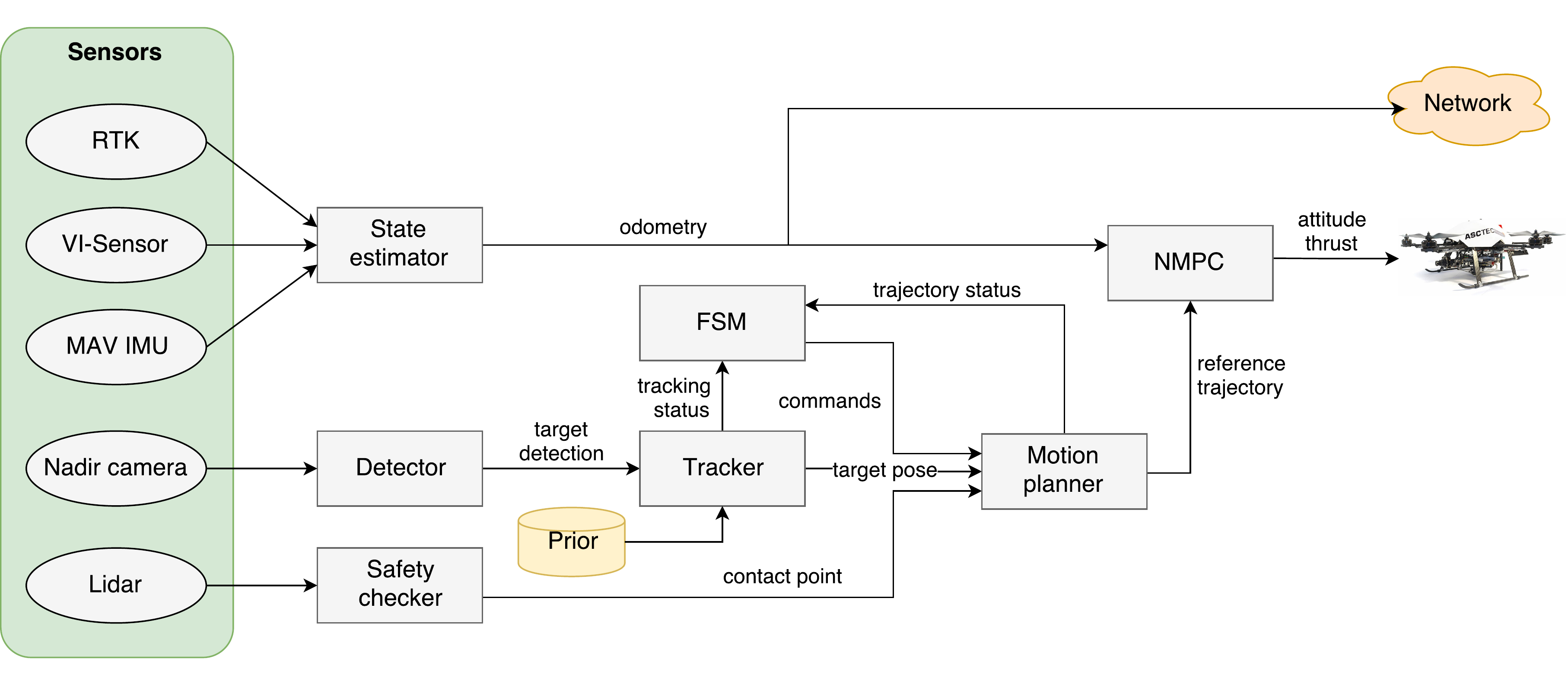}
  \caption{Block overview of the modules used for Challenge 1.}
  \label{fig:ch1_block_overview}
\end{figure}

Challenge 1 requires an \ac{MAV} to land on a moving platform. The landing platform, also referred to in the following as the ``target'', moves inside the arena, on an eight-shaped path with a constant velocity, and is characterized by a specific mark composed of a square containing a circle and a cross (\autoref{fig:ch1_close_partial}).
Challenge 1 can be decomposed into seven major tasks (\autoref{fig:ch1_block_overview}):
\begin{enumerate*}[(i)]
\item \ac{MAV} pose estimation (``State estimator'' block),\label{ch1_pose_estimation}
\item \ac{MAV} control (``NMPC'' block),\label{ch1_control}
\item landing platform detection (``Detector'' block),\label{ch1_detect}
\item landing platform tracking (``Tracker'' block),\label{ch1_track}
\item planning the landing maneuver from the current \ac{MAV} position to above the landing platform (``Motion planner'' block),\label{ch1_plan}
\item executing a final safety check before switching off the propellers (``Safety checker'' block), and \label{ch1_safety_check}
\item steering and coordinating all the previous modules together with a \ac{FSM} (``FSM'' block).
\end{enumerate*}

As \ref{ch1_pose_estimation} and \ref{ch1_control} were addressed in Sections \ref{sec:state} and \ref{sec:control}, only the remaining tasks are discussed in the following.

Based on the challenge specifications, we made the following assumptions to design our approach:
\begin{itemize}
\item the center of the arena and the path are known/measurable using \ac{RTK} \ac{GPS},
\item the landing platform is horizontal and of known size,
\item the height of landing platform is known up to a few centimeters,
\item the width of platform markings is known, and
\item the approximate speed of the platform is known.
\end{itemize}

\subsection{Platform Detection}
\begin{figure}[h]
\centering

\begin{subfigure}[t]{.48\textwidth}
\centering
\includegraphics[width=1.0\linewidth]{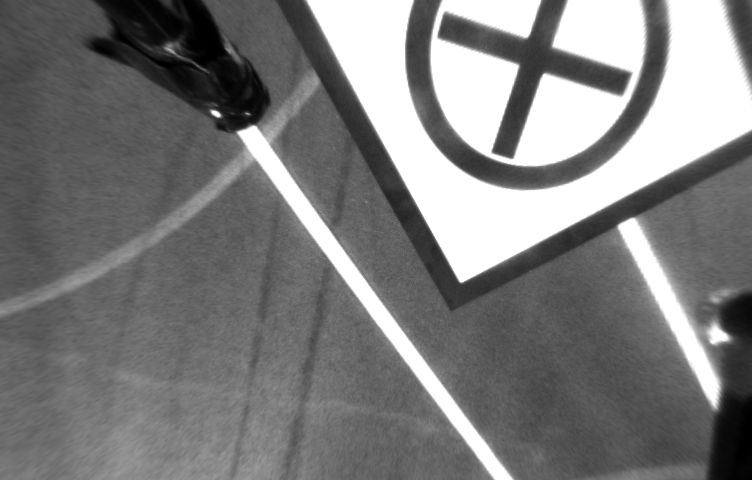}
\caption{During final approach: Good visibility of center markings and partial visibility of outline markings.}
\label{fig:ch1_close_partial}
\end{subfigure}\hfill
\centering
\begin{subfigure}[t]{.48\textwidth}
\centering
\includegraphics[width=1.0\linewidth]{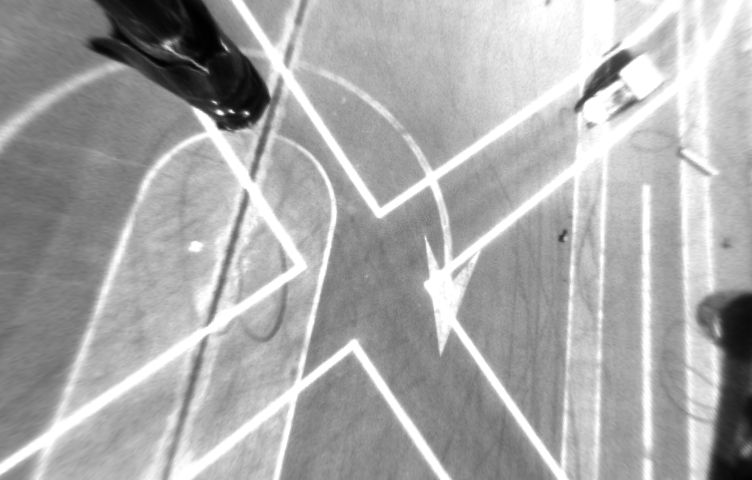}
\caption{During wait/search from high altitude: No center markings visible, outline barely visible.}
\label{fig:ch1_no_outline}
\end{subfigure}\hfill

\caption{Rectified images from the nadir fish-eye camera for detection.
Due to lighting, distance and view point variations, the appearance of the landing target may change significantly.}
\label{fig:ch1_viewpoint}
\end{figure}

A Point Grey Chameleon USB 2.0 camera ($752 \unit{px} \times 480 \unit{px}$) with a fisheye lens in nadir configuration is used for platform detection.
Given the known scale, height above ground, and planar orientation of the platform and its markings, the complete relative pose of the platform is obtainable using monocular vision.
As the landing platform moves at up to $15 \unit{km / h}$, a high frame rate for the detector is desirable.
Even at $30 \unit{fps}$ and $100 \%$ recall and visibility, the platform moves up to $14 \unit{cm}$ between detections.
As a result,
our design uses two independent detectors, invariant to scale, rotation, and perspective, in parallel.
A quadrilateral detector identifies the outline of the platform when far away or when its markings are barely visible (as in \autoref{fig:ch1_no_outline}), and a cross detector relies on the center markings of the platform and performs well in close-range situations (as in \autoref{fig:ch1_close_partial}).

\subsubsection{Quadrilateral Detector}
As the lens distortion is rectified, a square landing platform appears as a general quadrilateral under perspective projection.
Thus, a simple quadrilateral detector can pinpoint the landing platform regardless of scale, rotation, and perspective.
We used the quadrilateral detector proposed in the AprilTags algorithm \cite{olson2011tags}.
It combines line segments that end respectively start sufficiently close to each other into sequences of four.
To increase throughput of the quadrilateral detector, the line segment detection of the official AprilTags implementation is replaced by the EDLines algorithm as proposed in \cite{edlines}.
As the quadrilateral detector does not consider any of the markings within the platform, it is very robust against difficult lighting conditions while detecting many false positives.  Thus, a rigorous outlier rejection procedure is necessary.

\subsubsection{Cross Detector}
\begin{figure}[t]
\centering
\includegraphics[width=0.5\linewidth]{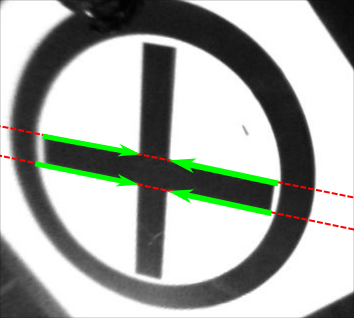}
\caption{Selected line segments (green) and their corresponding line (red). The orientation of the line segments is indicated by the arrow tip.}
\label{fig:cross_detector}
\end{figure}

Because the quadrilateral detector is designed for far-range detection and does not work in partial visibility conditions, we use a secondary detector, called the  ``cross detector'', based on platform markings for close-range maneuvers.
The cross detector uses the same detected line segments as output by the EDLines algorithm \cite{edlines} as input and processes them as follows:
\begin{enumerate}
	\item For each found line segment the corresponding line in polar form is calculated (\autoref{fig:cross_detector}).

    \item Clusters of two line segments whose corresponding lines are sufficiently similar (difference of angle and offset small) are formed and their averaged corresponding line is stored.

    \item Clusters whose averaged corresponding lines are sufficiently parallel are combined, thus yielding a set of four line segments in a 2 by 2 parallel configuration (\autoref{fig:cross_detector}).

    \item The orientation of line segments is determined such that two line segments on the same corresponding line point toward each other, thus allowing the center point of the detected structure to be computed.
\end{enumerate}

\hfill

Similarly to the quadrilateral detector, the cross detector is scale- and rotation-invariant.
However, the assumption of parallel corresponding lines (Step 3) does not hold under perspective transformation.
This can be compensated by using larger angular tolerances when comparing lines and by the fact that the detector is only needed during close-range maneuvers when the camera is relatively close to above the center of the platform.

The cross detector also generates many false-positive matches which are removed in the subsequent outlier rejection stage.

\subsubsection{Outlier Rejection}
As the landing platform is assumed to be aligned horizontally at a known height, the observed 2D image coordinates can be directly reprojected into 3D arena coordinates.

\begin{figure}[h!]
  \centering
  \includegraphics{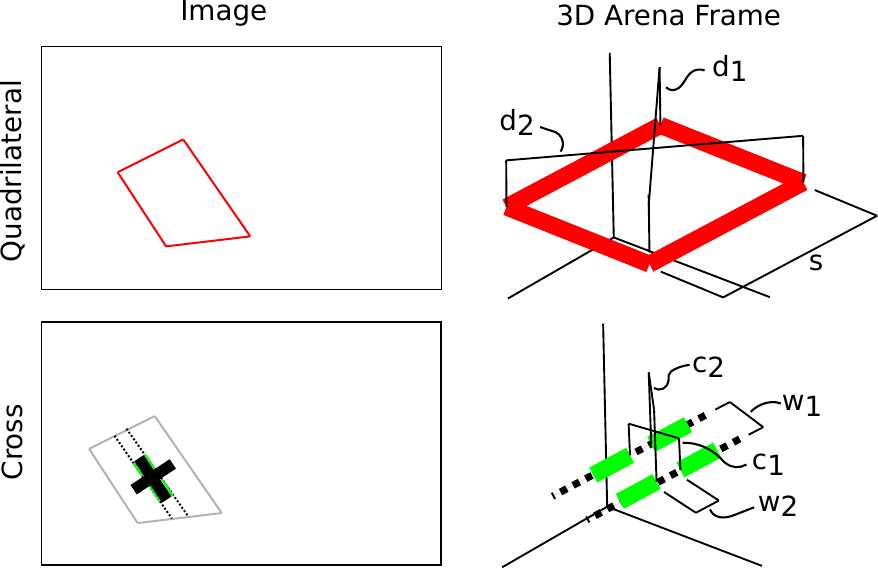}
  \caption{Calculation of measurements for outlier rejection: Image coordinates are reprojected onto a fixed $z$ plane (given by the z-height of the platform) and several simple measurements ($d_{1},d_{2},s,c_{1},c_{2},w_{1},w_{2}$) are taken.
  Note that reprojections of false positive detections that are not planar in the real world lead to highly distorted objects on the fixed $z$ plane, and thus have no chance passing the outlier rejection.}
  \label{fig:ch1_outlier}
\end{figure}

For the output of the quadrilateral detector, this enables calculating true relative scale and ratio based on the four corner points.
Similarly, the observed line width of the cross detector output can be determined.
Together with the prior probability of observing a platform in a certain location, these metrics are used to compute an overall probability of having observed the true landing platform.
\autoref{fig:ch1_outlier} visualizes a simple example for both detectors.

For each detected quadrilateral, we compute the ratio of sides $r$ and relative scale $s$ from the diagonal lengths, $d_{1}$ and $d_{2}$:
\begin{align}
r=
\begin{cases}
\frac{d_{1}}{d_{2}} & \text{, if $d_{2} > 0$,} \\
0           & \text{, if $d_{2} = 0$,}
\end{cases} &&
c = \frac{d_{1}+d_{2}}{2 \cdot t } \texttt{,}
\end{align}
where $t = 1.5 \cdot \sqrt{2} \unit{m}$ is the nominal size of the platform diagonal according to the challenge specifications.

Based on ratio $r$, scale $c$, and the calculated $x$-$y$-position, the ratio likelihood $l_{r} = L(r,1,\sigma_{r})$, scale likelihood $l_{c} = L(c,1,\sigma_{c})$, and position likelihood $l_{p} = \mathcal{L}_{\mathrm{track}}(x,y)$ are calculated.
$L(y,\mu,\sigma)$ is modeled as a Gaussian likelihood function and $\mathcal{L}_{\mathrm{track}}$ is a joint likelihood of the along and across-track distribution according to the path described in \autoref{ch1_path}. Note that $\mathcal{L}_{\mathrm{track}}$ is a numerical approximation, as the two distributions are not truly independent.
Depending on the tuning of $\sigma$, the response of the likelihood functions is sharp, and thus a threshold on the approximate joint likelihood $l_{r} \cdot l_{c} \cdot l_{p}$ yields good outlier rejection.

Similarly, the measurements $c_{1},c_{2},w_{1},w_{2}$ are taken for each response of the cross detector and their individual likelihood is calculated and combined with a position likelihood according to the path.
Here the nominal value (mean) for $w_{1}$ and $w_{2}$ is $15$\,cm, and for $c_{1}$ and $c_{2}$ it is $15 \cdot \sqrt{2}$\,cm.

Adjusting the covariances of each likelihood function and the threshold value allows the outlier rejection procedure to be configured to a desired degree of strictness.

\subsection{Platform Tracking}
\label{subsection_platform_tracking}
After successfully detecting the landing target, the \ac{MAV} executes fast and high-tilt maneuvers in order to accelerate towards it.
During these maneuvers, the target may exit the \ac{FoV} for a significant period of time, leading to sparse detections.
Thus, a key requirement for the tracker is an ability to deduce the most probable target location in the absence of measurements.

In order to maximize the use of a priori knowledge about the possible target on-track location, along-track movement and speed, we chose a non-linear particle filter.
Particle filters are widely used in robotics for non-linear filtering \cite{thrun2002particle}, as they approximate the true a posteriori of an arbitrary complex process and measurement model. In contrast to the different variants of Kalman filters, particle filters can track multiple hypotheses (multi-modal distributions) and are not constrained by linearization or Gaussianity assumptions, but generally are more costly to compute.

The following elements are needed for our particle filter implementation based on the Bayesian Filtering Library \cite{bfl-url}:
\begin{itemize}
	\item \textit{State}. Each particle represents a sampling $(x_{A}, y_{A}, \theta)$, where $x_{A}$ and $y_{A}$ are the location in \textit{arena} frame $A$, and $\theta$ corresponds to the movement direction along the track.
  As $z_{A}$ and velocity are assumed, they are not part of the state space.
    \item \textit{A priori distribution}. See \autoref{sec:ch1_prior_dist}.
    \item \textit{Process model}. Generic model (\autoref{sec:process_model}) of a vehicle moving with steering and velocity noise along a path (\autoref{ch1_path}).
    \item \textit{Measurement model}. A simple Gaussian measurement model is employed. Particles are weighted according a 2D Gaussian distribution centered about a detected location.
    \item \textit{Re-sampling}. Sampling importance re-sampling is used.
\end{itemize}

The resulting a posteriori distribution can then be used for planning and decision-making, e.g., aborting a landing approach if the distribution has not converged sufficiently or tracking multiple hypotheses until the next detection.

\subsubsection{Platform Path Formulation}
\label{ch1_path}
Piecewise cubic splines are used to define the path along which the target is allowed to move within certain bounds.
The main advantage of such a formulation is its generality, as any path can be accurately and precisely parametrized.
However, measuring distance along or finding a closest point on a cubic spline is non-trivial, given that no closed form solution exists.
Instead, we leverage iterative algorithms to pre-compute and cache the resulting data, thus providing predictable runtime and high refresh rates.
Since our on-board computer is equipped with sufficient memory, we store the following mappings:
\begin{itemize}
	\item Parametric form with range $[0, \ldots, 1]$ to length along the track, and vice-versa.
	\item $x_{A}$, $y_{A}$ position in the \textit{arena} frame $A$ to parametric form using nearest neighbor search.
\end{itemize}

\begin{figure}[h!]
  \centering
  \includegraphics{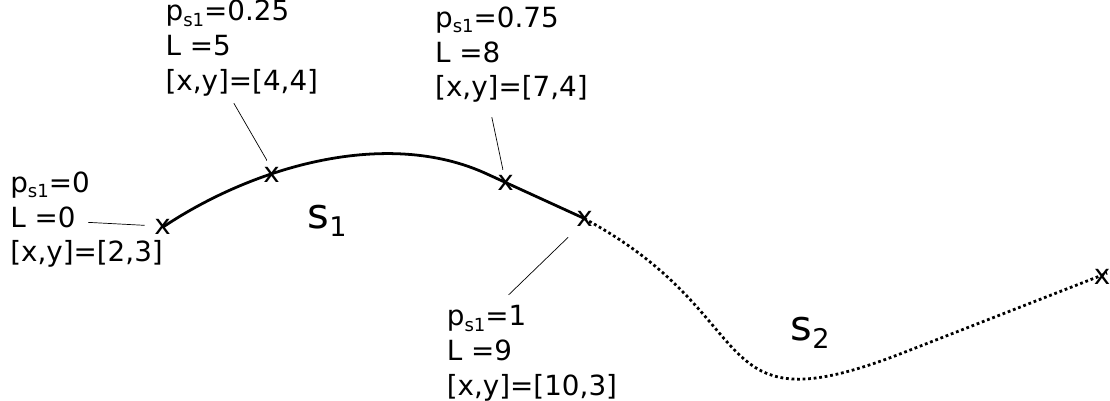}
  \caption{Simplified view of piece-wise cubic spline formulation. $p_{s1}$ is the parameter along the first cubic spline piece, $l$ the length along the track, and $xy$ indicates the coordinates in \textit{arena} frame.
  The $l$ and $xy$ values are cached in KD-trees for efficient lookup of the corresponding parameter $p$.}
  \label{fig:ch1_path}
\end{figure}

A major benefit of this approach is that it requires no iterative algorithms during runtime. \autoref{fig:ch1_path} provides an illustrative example.
To facilitate efficient lookup and nearest-neighbor search, the samplings are stored in indexed KD-trees \cite{blanco2014nanoflann}.

\subsubsection{Prior Distribution}
\label{sec:ch1_prior_dist}
\begin{figure}[h!]
\centering
\begin{subfigure}[t]{.48\textwidth}
\centering
\includegraphics[width=1.0\linewidth]{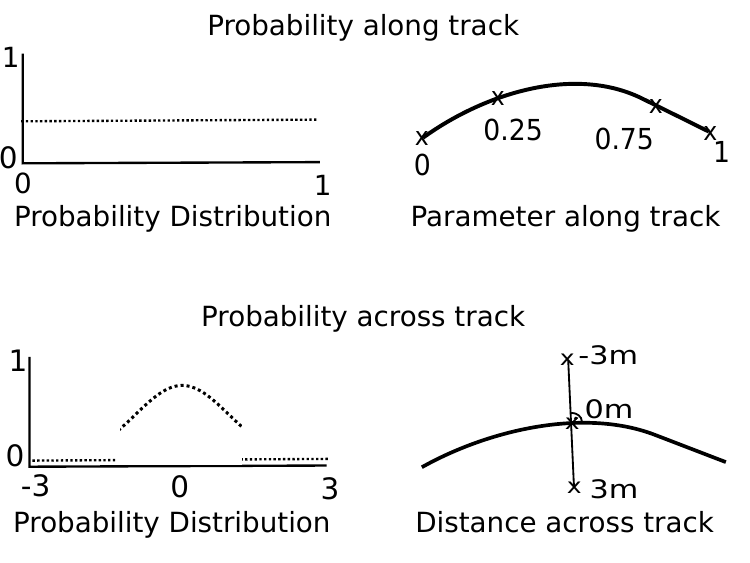}
\caption{Independent prior distributions: Uniform along the track (along parameter $p$ of the splines), truncated normal distribution across the track (perpendicular distance in meters).}
\label{fig:ch1_prior_schema}
\end{subfigure}\hfill
\centering
\begin{subfigure}[t]{.48\textwidth}
\centering
\includegraphics[width=1.0\linewidth]{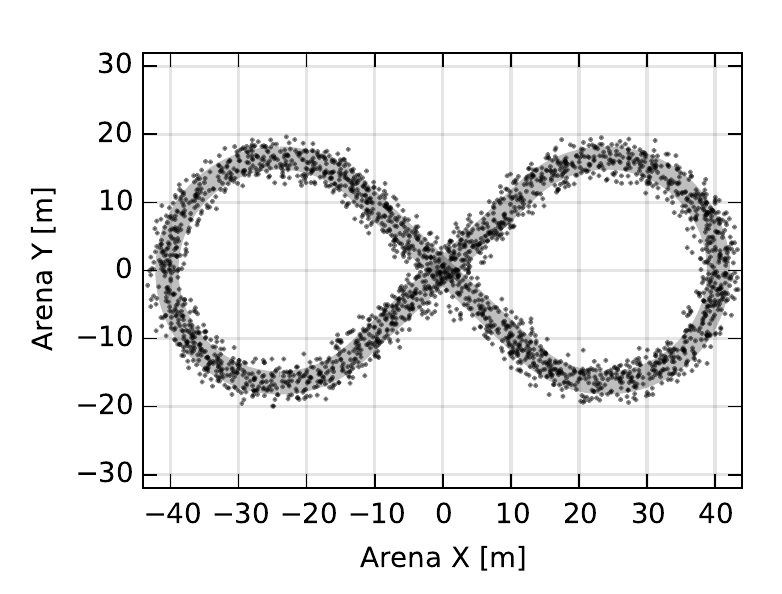}
\caption{Sampled prior distribution: Each dot corresponds to one of $2500$ samples on the $3 \unit{m}$ wide platform path (gray).
Note that the cut-off of the truncated Gaussian distribution is chosen to leave some slack in case of slight positioning errors of the track.}
\label{fig:ch1_prior_img}
\end{subfigure}\hfill
\caption{Schematic and sampled prior distributions.}
\label{fig:ch1_prior}
\end{figure}

The location of each particle is initially sampled from two independent distributions, where
(i) the location along the track is modeled as a uniform distribution along the full length of the track, i.e., the target can be anywhere, and
(ii) the location across the track is modeled as a truncated Gaussian distribution, i.e, we believe that the platform tends to be more towards the center of the path, and cannot be outside the road limits.
\autoref{fig:ch1_prior} shows a schematic and sampled image of the prior.
Note that this is a simplification, as the two distributions are not truly independent, e.g., they overlap along curves.

\subsubsection{Process Model}
\label{sec:process_model}
\begin{figure}[h!]
  \centering
  \includegraphics{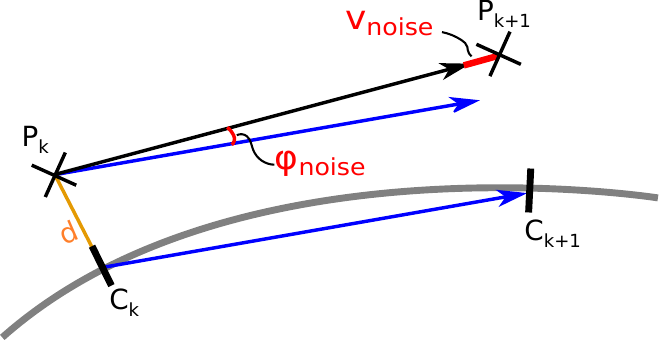}
  \caption{Illustration of the process model.
  $P_{k}$ and $P_{k+1}$ are the position of an individual particle at time $k$ or $k+1$ respectively.
  $C_{k}$ is the closest point to $P_{k}$ on the cubic spline path center.
  $C_{k+1}$ corresponds to the position after moving along the path at speed $s$ for $\Delta t$ seconds.}
  \label{fig:ch1_motion}
\end{figure}

We chose a generic process model that is independent of the vehicle type, e.g., Ackermann, unicycle or differential drive.
Another motivation behind selecting this model was its execution time, as the following process is applied to $\sim 2500$ individual particles at $\sim 50 \unit{Hz}$.
It is based on the assumption that there is an ideal displacement vector of movement along the track between two sufficiently small timesteps, as exemplified by the vector between $C_{k}$ and $C_{k+1}$ in \autoref{fig:ch1_motion}.
In order to obtain this vector, the current position of a particle $P_{k}$ is mapped onto the closest location on-track $C_{k}$ and then displaced along the track according to ideal speed and timestep size, thus obtaining $C_{k+1}$.
Note that these operations are very efficient based on the cached samples discussed in subsection \ref{ch1_path}.

The ideal displacement vector is then added to the current true position $P_{k}$ and disturbed by steering noise $\varphi_{\mathrm{noise}}$ and speed noise $v_\mathrm{noise}$, both of which are sampled from zero-mean Gaussian distributions.

\begin{figure}[h]
\centering
\begin{subfigure}[t]{.48\textwidth}
\centering
\includegraphics[width=1.0\linewidth]{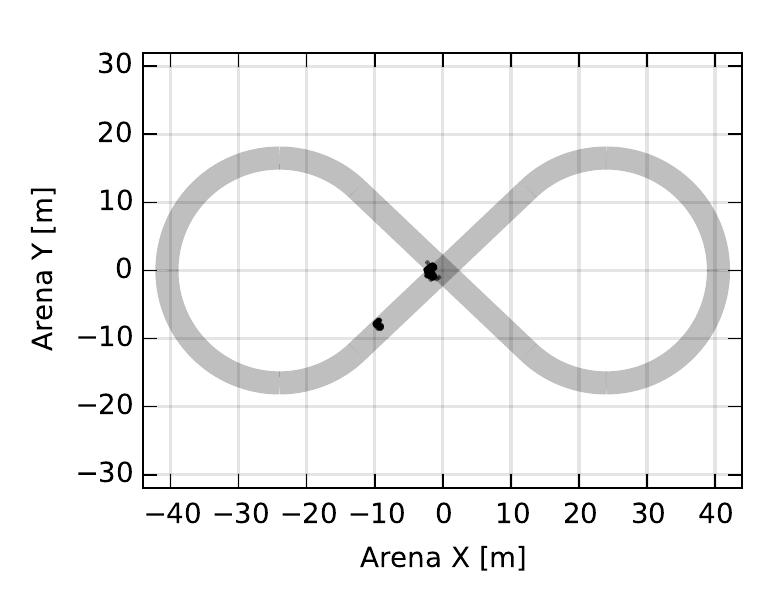}
\caption{State after exactly one detection and a few cycles without detection: Direction not determined yet, yielding two separate possible hypotheses that move in opposite directions.}
\label{fig:ch1_split_particles}
\end{subfigure}\hfill
\centering
\begin{subfigure}[t]{.48\textwidth}
\centering
\includegraphics[width=1.0\linewidth]{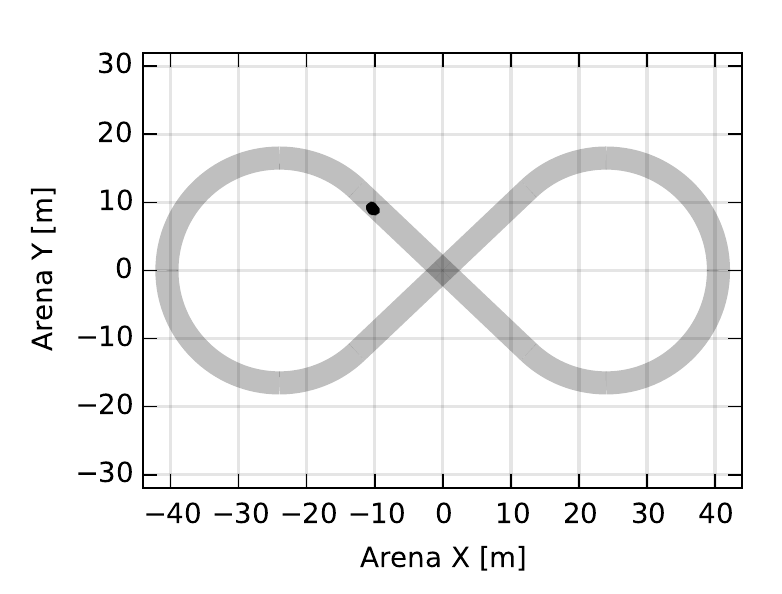}
\caption{Converged tracker: Single, consistent hypothesis.}
\label{fig:ch1_conv_particles}
\end{subfigure}\hfill
\caption{Visualization of particle filter states.}
\label{fig:ch1_particles}
\end{figure}

\subsubsection{Convergence Criteria}

The target location distribution calculated by the particle filter is used for autonomous decision-making, such as starting or aborting a landing approach.
Each timestep, the tracker determines whether the state has converged by simply checking if more than a certain threshold of the probability mass lies inside a circle with a given radius (subsequently  called ``convergence radius''), centered on the current weighted mean (\autoref{fig:ch1_convergence}).
This criterion can be determined in linear time and has proven to be sufficiently precise for our purposes with a probability mass threshold of 0.75 with a convergence radius of $1 \unit{m}$.
\autoref{fig:ch1_particles} shows a typical split-state of the particle filter as it occurs after a single detection.
The two particle groups are propagated independently along the track, and converge on one hypothesis as soon as a second measurement is available.

\begin{figure}[h]
\centering
\begin{subfigure}[t]{.48\textwidth}
\centering
\includegraphics[width=1\linewidth]{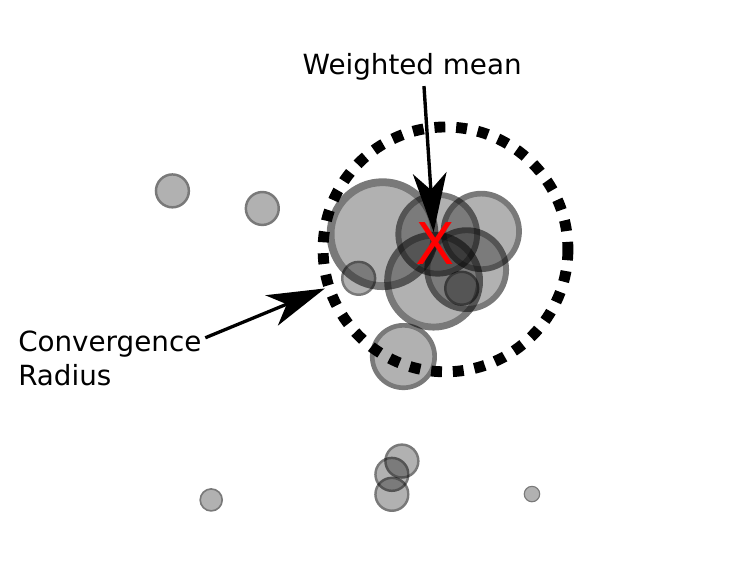}
\caption{Example of a converged state. A large majority of probability mass (summed weight) lies inside the convergence radius.}
\label{fig:ch1_convergence}
\end{subfigure}\hfill
\centering
\begin{subfigure}[t]{.48\textwidth}
\centering
\includegraphics[width=1\linewidth]{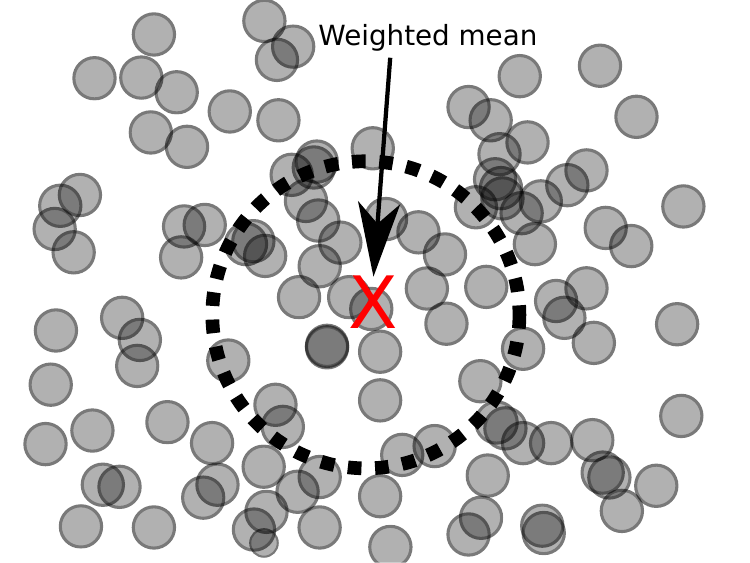}
\caption{Example of a diverged state.}
\label{fig:ch1_no_convergence}
\end{subfigure}\hfill
\caption{Calculation of the convergence criteria. Each particle is visualized using a gray circle whose radius represents the current weight of the particle.}
\label{fig:ch1_convergence}
\end{figure}

As the process model increases the particle scattering if no measurements are present, the filter automatically detects divergence after a period with no valid detections.

To calculate the future position of the target, a subsampling of the current particles is propagated forward in discrete timesteps using the process model only.
This gives a spatio-temporal 4D trajectory of the predicted future state of the target over a given time horizon, which then can be used for planning.

\subsection{Motion Planning}
The motion planning task can be divided based on 3 different modes of operation:
\begin{enumerate}
	\item \textit{Searching}. The current state of the tracker is ignored and a fixed location or path that maximizes chances of observing the target is followed.
  Here, the \ac{MAV} hovers $10 \unit{m}$  above the center of the eight-shaped path.
    \item \textit{Following}. The \ac{MAV} should follow the target closely, e.g., $2 \unit{m}$ above its center, with the same velocity.

    \item \textit{Final approach}. The \ac{MAV} should approach the target so that a landing is possible. The constraints are as follows: relative position and velocity in $x$ and $y$ directions is zero, relative velocity in $z$ direction is chosen so that the airframe can absorb the impact shock (here: $< 0.75 \unit {m / s}$).
    The first order derivatives of relative $x$ and $y$ speeds are also fixed to zero.
\end{enumerate}

As mentioned, the tracker outputs a sampled spatio-temporal 4D trajectory of the predicted target position.
Using this data, the \ac{MAV} trajectory is calculated such that its position for each sample of the predicted target position and time coincides with the constraints as quickly as possible while satisfying flight envelope restrictions.

\begin{figure}[h]
\centering
\begin{subfigure}[t]{.48\textwidth}
\centering
\includegraphics[width=1\linewidth]{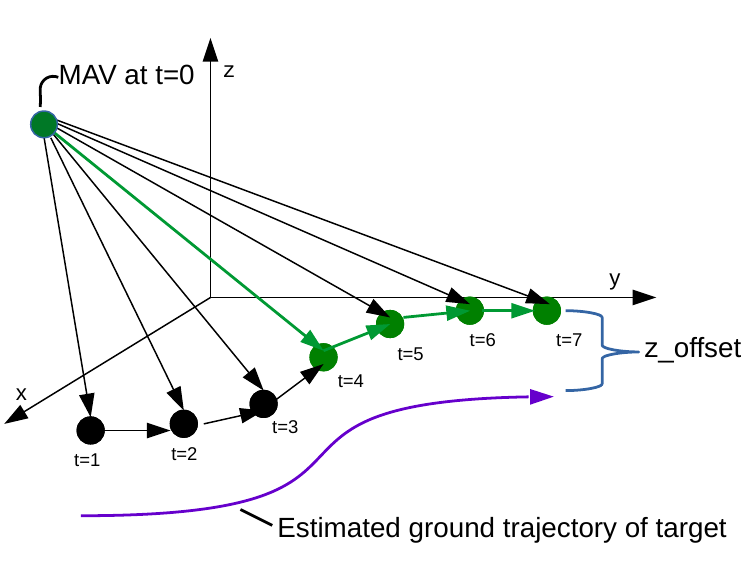}
\end{subfigure}\hfill
\centering
\begin{subfigure}[t]{.48\textwidth}
\centering
\includegraphics[width=1\linewidth]{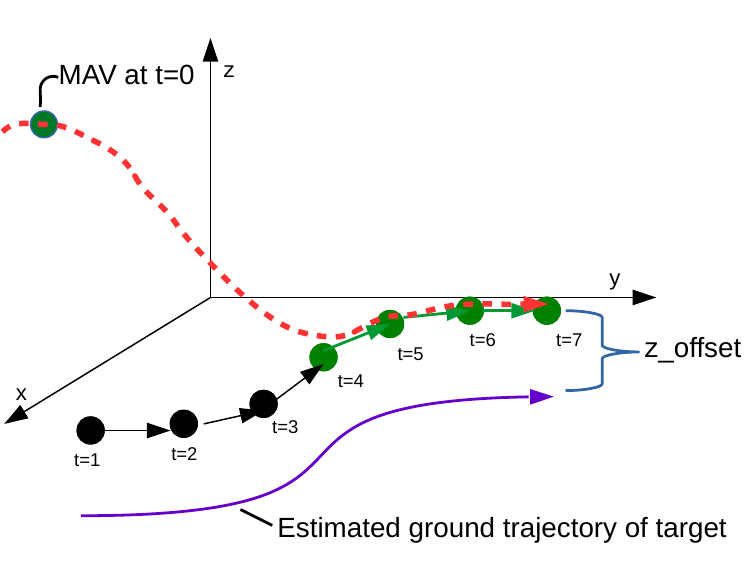}
\end{subfigure}\hfill
\caption{Directed graph based on current location and trajectory estimation of the target, with indicated shortest (cost-wise) path (green edges/vertices).
Each edge represents a possible motion primitive.
The red dashed line indicates the resulting smooth trajectory through the selected vertices.}
\label{fig:ch1_graph}
\end{figure}

We optimized the pipeline such that replanning at a frequency of $50 \unit{Hz}$ is possible, as the trajectory is updated after each step of the particle filter. This ensures that trajectory planning is always based on the most recent available information.
The generated trajectory consists of smoothly joined motion primitives up to acceleration, as proposed by \cite{mueller2015computationally}.
In order to select a valid set of primitives, a graph is generated and the cheapest path is selected using Dijkstra's algorithm. The set of graph vertices consists of the current \ac{MAV} position and time (marked as start edge), possible interception points, and multiple possible end positions according to the predicted target trajectory and the constraints (\autoref{fig:ch1_graph}).
Each graph edge represents a possible motion primitive between two vertices.
Vertices can have defined position, time and/or velocity.
The start state is fully defined as all properties are known, whereas intermediate points might only have their position specified, and possible end-points have fixed time, position, and velocity according to the prediction.

Dijkstra's algorithm is then used to find the shortest path between the start and possible end-states.
Intermediate states are updated with a full state (position, velocity and time) as calculated by the motion primitive whenever their Dijkstra distance is updated.
Motion primitives between adjacent vertices are generated as follows:
\begin{itemize}
	\item If the start and end times of two vertices are set, then the trajectory is chosen such that it minimizes jerk.
    \item If the end time is not set, then the path is chosen such that it selects the fastest trajectory within the flight envelope.
\end{itemize}

The cost function used to weigh edges is simply the duration of the motion primitive representing the edge times a multiplier.
The multiplier is $1$ for all edges that are between predicted platform locations, and $l^{3}$ for all edges connecting the start state with possible intersection points, where $l$ is the distance between points.
Effectively, this results in an automatic selection of the intersection point, with a heavy bias to intersect as soon as possible.

\subsection{Lidar Landing Safety Check}
\label{sec:ch1_lidar}
A successful Challenge 1 trial requires the \ac{MAV} to come to rest in the landing zone with the platform intact and its propellers producing no thrust.
Starting the final descent and then switching off the propellers on time was mission-critical for the overall result.
These commands must be issued only when the \ac{MAV} is considered to be above the landing platform, with a height small enough to allow its magnetic legs to attach to the metal part of the target.
The range output of a \ac{Lidar} sensor is employed to check the distance of the \ac{MAV} from the landing platform.
Even though the \ac{MAV} already has all necessary information to land available at this stage, i.e., its global position and an estimate of the current platform location, an extra safety check is performed using the \ac{Lidar} output:
before triggering the final part of the landing procedure, it is necessary that the \ac{Lidar} detects the real platform below the \ac{MAV}.

This final safety check is executed using a stand-alone \ac{ROS} node, which receives the \ac{MAV} odometry (from the state estimation module), the estimated position of the landing platform (from the tracking module) and raw distance measurements from the \ac{Lidar}.
Raw measurements are provided in the form of $_L\vec{d}_{L,C_p} = (0, 0, z)\T$, where $z$ is the distance measurement from the sensor lens to a contact point $C_p$ of the laser beam.
They are expressed in a \textit{lidar} frame $L$, whose $z$ axis points out from the sensor lens.
Raw measurements are converted in \textit{arena} frame $A$ distance measurements ${_A}\vec{d}_{A,C_p}$, by using the \ac{MAV} odometry ${_A}\mat{T}_{A,B}$ and the qualitative displacement from \textit{lidar} frame $L$ to \textit{mav-imu} frame $I$ ${_B}\mat{T}_{B,L}$:
\begin{align}
	{_A}\vec{d}_{A,C_p} = {_A}\mat{T}_{A,B} \, {_B}\mat{T}_{B,L} \, {_L}\vec{d}_{L,C_p}\texttt{,}
    \label{equation_lidar_contact}
\end{align}
where $_L\vec{d}_{L,C_p}$ indicates the raw measurement in \textit{lidar} frame $L$.
Vector $_A\vec{d}_{A,C_p}$ can be seen as a ``contact point'' of the \ac{Lidar} beam, expressed in global coordinates.

The aforementioned \ac{ROS} node signals that the \ac{MAV} is actually above the platform if two conditions are met:
\begin{enumerate*}[(i)]
\item the global \ac{MAV} position is above the estimated global platform position\label{lidar_condition_mav}, regardless of its altitude, and
\item the \ac{Lidar} contact point intercepts the estimated position of the platform\label{lidar_condition_lidar}.
\end{enumerate*}
A Mahalanobis distance-based approach is employed to verify these conditions:
\begin{align}
	d_{M}(x) = \sqrt{(\vec{x}-\vec{\mu})\T \mat{S}^{-1} (\vec{x}-\vec{\mu})}\texttt{.}
\end{align}
This is applied with different vector and matrix dimensions to check the conditions above:
\begin{enumerate}[(i)]
\item $\vec{x} = (x_B, y_B)\T$, $\vec{\mu} = (x_T, y_T)\T$, and $\mat{S}=\mathrm{diag}(\sigma_{x_T}^2, \sigma_{y_T}^2)$,
where $\vec{x}$ and $\vec{\mu}$ denote 2D global positions of the \ac{MAV} and the estimated platform center, respectively. The diagonal matrix $\mat{S}$ contains the covariance of the estimated platform position, provided by the tracker module.
\item $\vec{x} = {_A}\vec{d}_{A,C_p}$, $\vec{\mu} = (x_T, y_T, z_T)\T$, and $\mat{S}=\mathrm{diag}(\sigma_{x_T}^2, \sigma_{y_T}^2, \sigma_{z_T}^2)$,
where $\vec{x}$ and $\vec{\mu}$ denote the 3D \ac{Lidar} beam contact point and the estimated platform center, respectively. The diagonal matrix $\mat{S}$ contains the covariance of the estimated 3D platform position, which are provided by the tracker module.
\end{enumerate}

\begin{figure}
\centering

\begin{subfigure}[t]{.48\textwidth}
\centering
\includegraphics[width=1.0\linewidth]{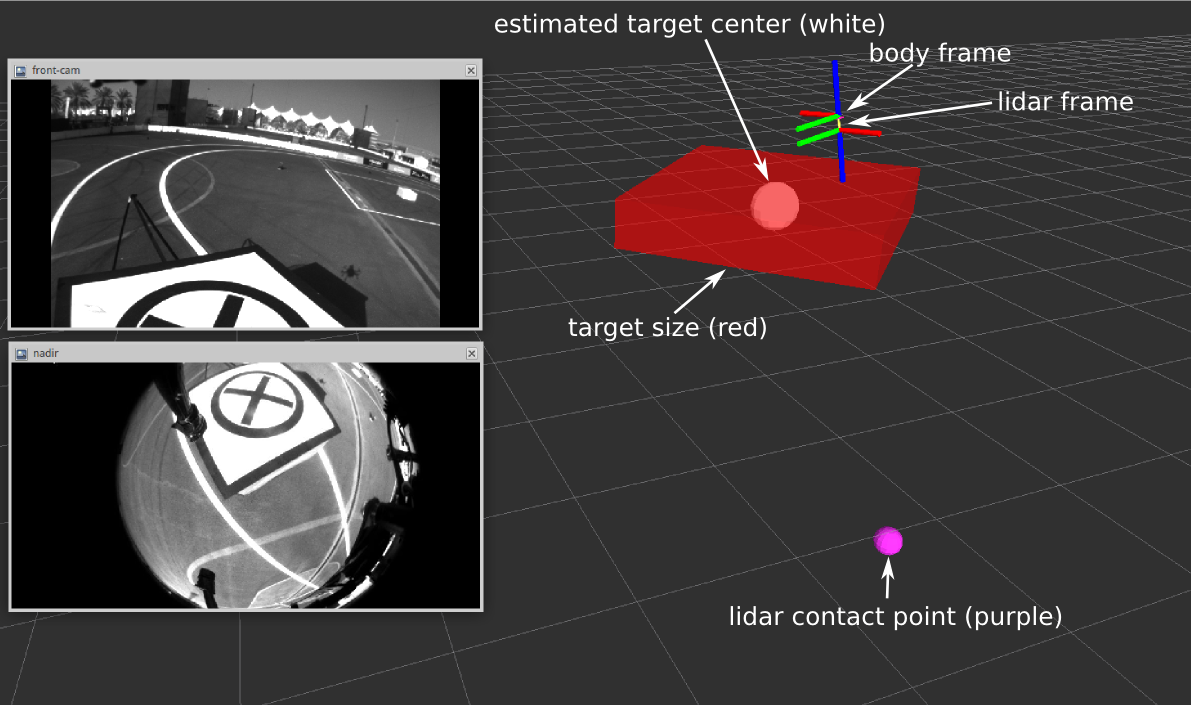}
\caption{First landing attempt: \ac{MAV} body frame was above the estimated location of the landing platform but, since the contact point of the \ac{Lidar} beam was on the ground and not on the supposed platform position, the final maneuver was aborted.}
\label{fig:ch1_lidar_abort}
\end{subfigure}\hfill
\centering
\begin{subfigure}[t]{.48\textwidth}
\centering
\includegraphics[width=1.0\linewidth]{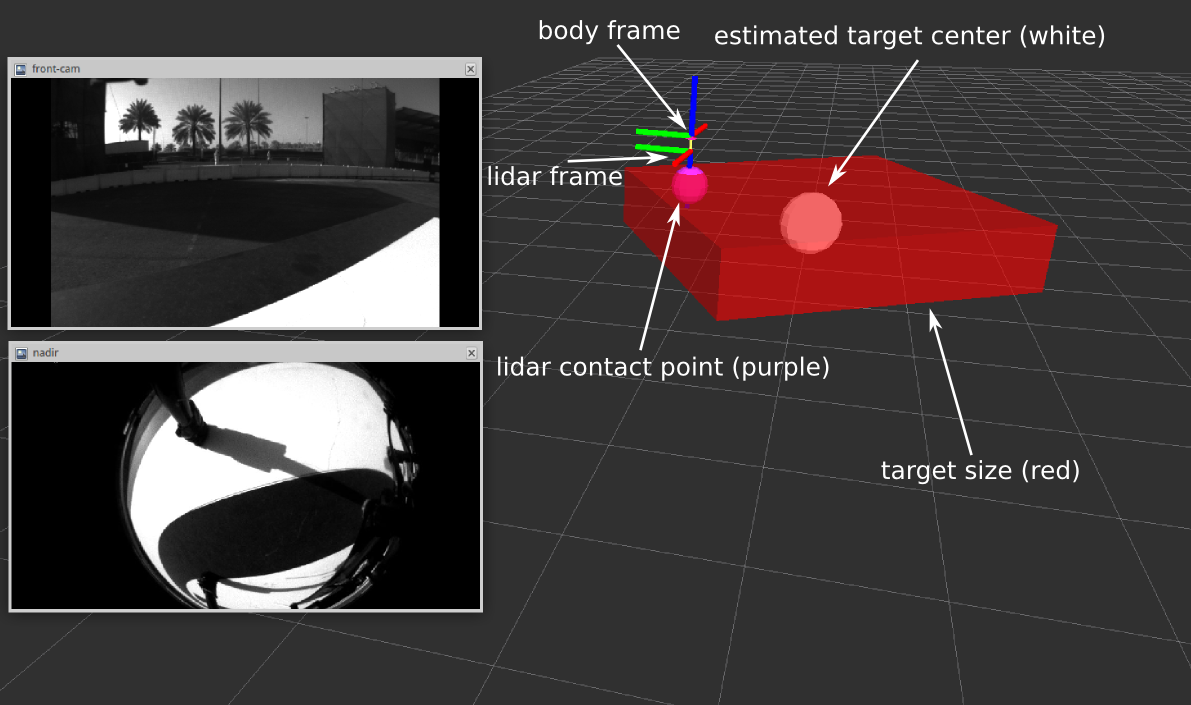}
\caption{Third landing attempt: \ac{MAV} body frame was above the estimated location of the landing platform and, since the contact point of the \ac{Lidar} beam was on the supposed platform position, the final maneuver was considered completed and the motors were switched off.}
\label{fig:ch1_lidar_land}
\end{subfigure}\hfill

\caption{Visualization of \ac{Lidar} data recorded during the first Grand Challenge trial.
Both images show the positions of the \ac{MAV} (body frame) and of the \ac{Lidar}, the images of the two cameras (front image from \ac{VI-Sensor} and bottom image from mono camera), the estimated platform center provided by the tracker module (white ball), the available space to land if the target were on the estimated position (red parallelepiped), and the contact point of the \ac{Lidar} beam.}
\label{fig:ch1_lidar_test}
\end{figure}

The two conditions have different thresholds, $\delta_{i} = 1.0$, and $\delta_{ii} = 1.5$, that can be set and adjusted dynamically.
Each condition is satisfied if its Mahalanobis distance is below the correspondence threshold.
Even though these two conditions are tightly related, since ${_A}\vec{d}_{A,C_p}$ is computed using ${_A}\mat{T}_{A,B}$ in \autoref{equation_lidar_contact}, they are handled separately.
This is because, by setting different thresholds, a higher importance is given to condition \ref{lidar_condition_lidar} than to \ref{lidar_condition_mav}.
This strategy allowed for a less error-prone final approach of the landing maneuver.
To demonstrate this,
\autoref{fig:ch1_lidar_test} shows two different landing attempts executed during the first Grand Challenge trial.
In the first two attempts, the final maneuver was aborted before completion, since only the first condition was met.
In the third attempt, the final approach was triggered as both conditions were met.

In \autoref{fig:ch1_lidar_abort}, it can be seen in the two camera images that the landing target was actually ahead the \ac{MAV} during the final approach.
Even though the first condition was satisfied, such that the 2D position of the \ac{MAV} was above the estimated platform position, the \ac{Lidar} contact point was on the ground.
This clearly indicates that the real platform was in a different location, and that probably the output of the tracker reached a weak convergence state.
Relying only on the estimated target position would have led to an incorrect descent maneuver; likely resulting in flying against the \ac{UGV} carrying the target.
On the other hand, \autoref{fig:ch1_lidar_land} shows the system state when both conditions were met and the landing approach was considered complete.

\subsection{Finite State Machine}
\begin{figure}
  \centering
  \includegraphics[width=0.75\textwidth]{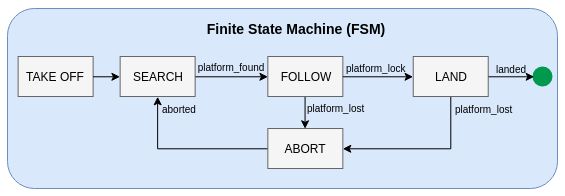}
  \caption{Task-level architecture for Challenge 1.}
  \label{fig:ch1_finite_state_machine}
\end{figure}

A high-level \ac{FSM} was implemented using the SMACH \ac{ROS} package \cite{smach_web} to integrate the Challenge 1 submodules.
The main states in the \ac{FSM} are displayed in \autoref{fig:ch1_finite_state_machine} and briefly explained below.
\begin{description}[font=\normalfont]
\item[SEARCH:] The \ac{MAV} is commanded to fly above the center of the arena, and the tracker module begins searching for the target platform.
When the motion direction of platform is successfully estimated and the target considered locked, the state switches to ``FOLLOW''.
\item[FOLLOW:] The \ac{MAV} is commanded to closely follow the target as long as either the platform is detected at a high rate, e.g., more than $7 \unit{Hz}$, or the tracker module decides the lock on the platform is lost.
In the former case, the \ac{MAV} is considered to be close enough to the platform, as the high detection rate is indicating, and the state is switched to ``LAND''.
In the latter case, the platform track is considered lost, so ``ABORT'' mode is triggered.
\item[LAND:] The \ac{MAV} first continues following the platform while gradually decreasing its altitude.
Once it is below a predefined decision height, the output of the \ac{Lidar} is used to determine whether the estimated position of the landing platform overlaps with the actual target position.
If that is the case, a fast descent is commanded and the motors are switched off as soon as a non-decreasing motion on the $z$-axis is detected, i.e., the \ac{MAV} lands on the platform to conclude Challenge 1.
If any of the previous safety checks fail, the state switches to ``ABORT''.
\item[ABORT:] The \ac{MAV} is first commanded to increase its altitude to a safety value in order to avoid any possibly dangerous situations.
Then, the \ac{FSM} transitions to ``SEARCH'' to begin seeking the platform again.
\end{description}

\section{Challenge 3: Search, Pick Up, and Relocate Objects}
\label{sec:ch3}
\begin{figure}
  \centering
  \subcaptionbox{An overview of the arena with 6 moving, 10 small static and 3 large static objects.\label{fig:ch3_foto_arena}}{\includegraphics{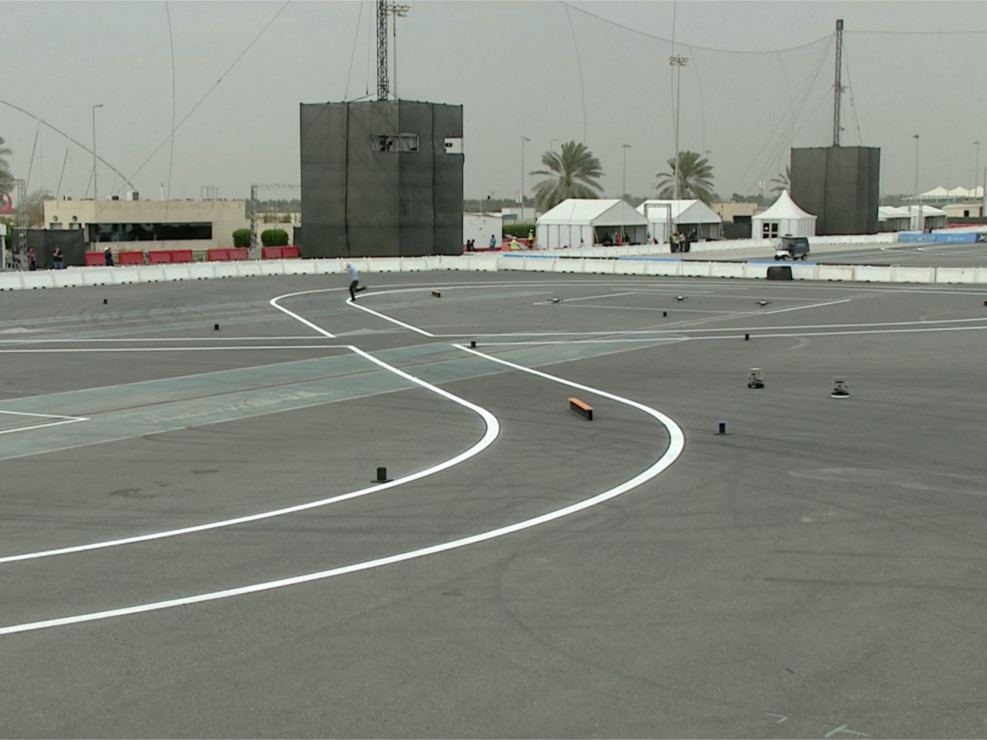}}\hfill%
  \subcaptionbox{An \ac{MAV} dropping a red disc in the drop box container.\label{fig:ch3_drop}}{\includegraphics{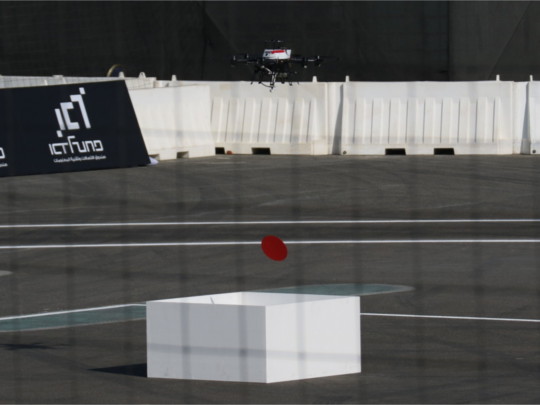}}%

  \caption{In Challenge 3 up to three \acp{MAV} need to collect as many objects as possible in the arena and deliver them to the drop box or dropping zone.}
\label{fig:ch3_arena}
\end{figure}

Challenge 3 requires a team of up to three \acp{MAV} equipped with grippers to search, find, pick, and relocate a set of static and moving objects in a $60 \unit{m} \times 90 \unit{m}$ planar arena (\autoref{fig:ch3_arena}).
The arena has 6 moving and 10 stationary small objects as well as 3 stationary large objects.
The moving objects move at random velocities under $5 \unit{km/h}$.
Small objects have a cylindrical shape with a diameter of $200 \unit{mm}$ and a maximum weight of $500 \unit{g}$.
Large objects have a rectangular shape with dimensions $150 \unit{mm} \times 2000 \unit{mm}$, a maximum weight of $2 \unit{kg}$, and require collabrative transportation.
Note that the presented system ignores large objects
since our collaboration approach was not integrated at the time of the challenge \cite{tagliabue2017robust,tagliabue2017}.\footnote{Also note that no team attempted to perform the collaborative transportation task during the challenge.}
The color of an object determines its type and score.
Teams gain points for each successful delivery to the drop container or dropping zone.

The challenge evokes several integration and research questions.
Firstly, our system must cater for the limited preparation and challenge execution time at the Abu Dhabi venue.
Every team had two $20 \unit{min}$ testing slots, two $20 \unit{min}$ challenge slots, and two $20 \unit{min}$ Grand Challenge slots, with flying otherwise prohibited.
Furthermore, each slot only had $\sim 30 \unit{min}$ preparation time to setup the network, \acp{MAV}, \ac{FSM}, and tune daylight-dependent detector parameters.
These time constraints require a well-prepared system and a set of tools for deployment and debugging.
Thus, a simple and clean \emph{system architecture} is a key component.

A second challenge is the \emph{multi-agent system}.
In order to deploy three \acp{MAV} simultaneously, we developed solutions for workspace allocation, exploration planning, and collision avoidance.

The third and most important task in this challenge is \emph{aerial gripping}.
Being able to reliably pick up $200 \unit{mm}$ discs in windy outdoor environments is challenging for detection, tracking, visual servoing, and physical interaction.
Given that only delivered objects yield points,
a major motivation was to design a reliable aerial gripping system.

Driven by these challenges,
we developed an autonomous multi-agent system for collecting moving and static small objects with unknown locations.
In the following we describe our solutions to the three key problems above.
The work behind this infrastructure is a combined effort of \cite{Gawel16,bahnemann2017decentralized}.

\subsection{System Architecture}
During development, it was recognized that a major difficulty for our system is the simultaneous deployment of more than one \ac{MAV}.
We also recognized network communication to be a known bottleneck in competitions and general multi-robot applications.
With these considerations, we opted for a decentralized system in which each agent can fulfill its task independently while sharing minimal information.

\begin{figure}
  \centering
  \includegraphics[width=\textwidth]{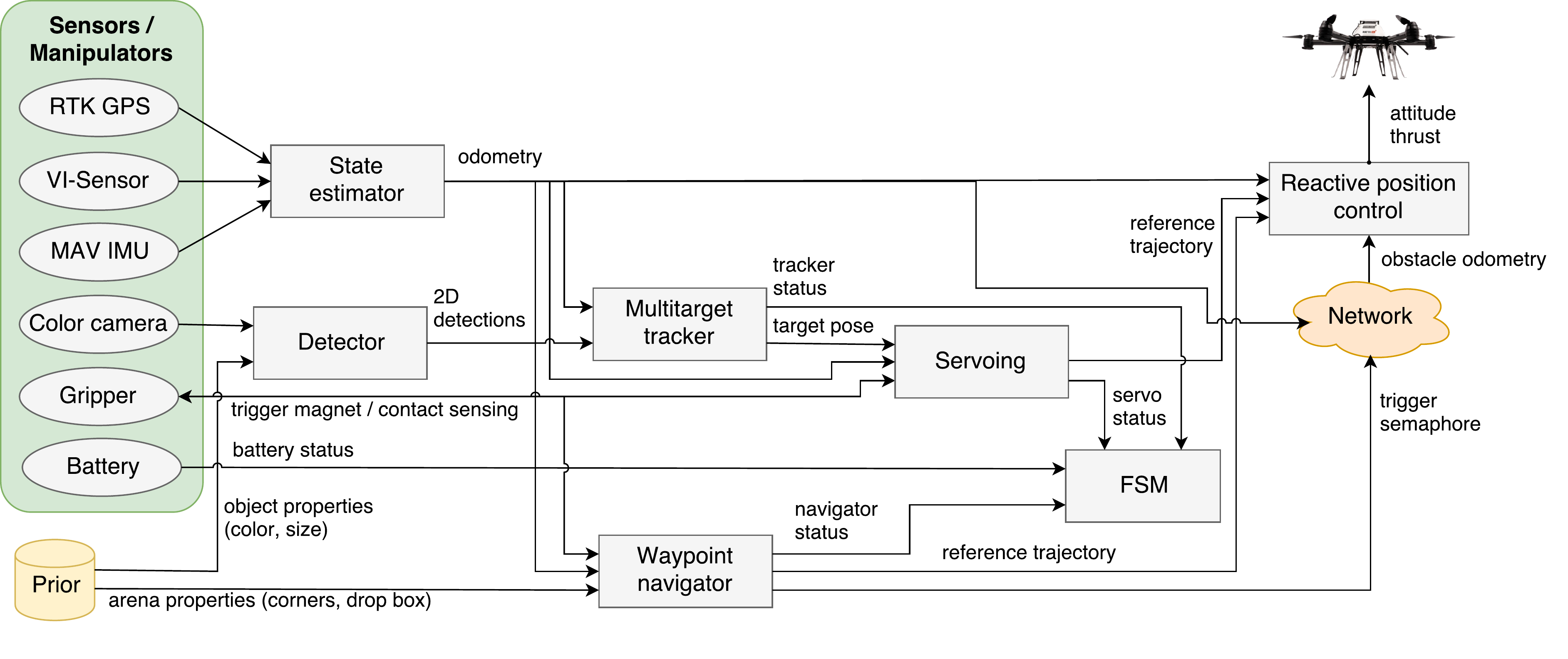}
  \caption{System diagram for Challenge 3.
  In addition to the mandatory state estimation and control modules, the system includes a waypoint navigator to explore the arena and fly to pre-programmed waypoints, and an elaborate object detection, tracking, and servoing pipeline.}
  \label{fig:ch3_system}
\end{figure}

Challenge 3 can be decomposed into four major tasks:
\begin{enumerate*}[(i)]
  \item state estimation,
  \item control,
  \item waypoint navigation, and
  \item aerial gripping.
\end{enumerate*}
A block diagram of this system for a single \ac{MAV} is shown in \autoref{fig:ch3_system}.
The key components for autonomous flight consist of the \emph{State Estimator} and \emph{Reactive Position Control} presented in \autoref{sec:state} and \autoref{sec:control}.
The \emph{Detector}, \emph{Multitarget Tracker}, and \emph{Servoing} blocks form the aerial gripping pipeline.
The \emph{Waypoint Navigator} handles simple navigation tasks, such as take off, exploration or object drop off.
Each \ac{MAV} communicates its odometry and controls a drop box semaphore over the \emph{Network} and receives prior information about the task.
The \emph{Prior} data consist of offside and onside parameters.
The object sizes, arena corners, and workspace allocation were taken from the challenge descriptions.
The object color thresholding, camera white balancing, the drop box position, and the number of \acp{MAV} to engage was defined during the on-stage preparation time.

\begin{figure}
  \centering
  \includegraphics[width=0.8\textwidth]{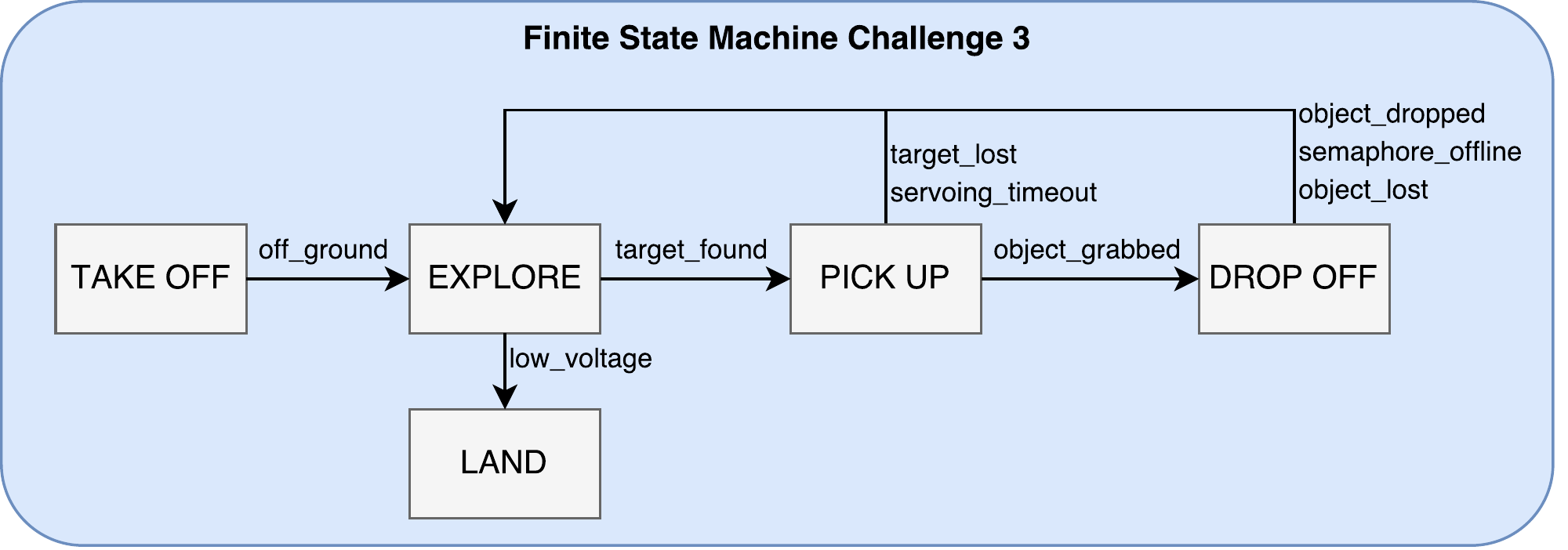}
  \caption{The \ac{FSM} architecture for Challenge 3.
  Each \ac{MAV} alternates between exploration and greedy small object pickup before landing with low battery voltage.}
  \label{fig:ch3_fsm}
\end{figure}

All modules are organized and scheduled in a high-level SMACH \emph{\ac{FSM}} \cite{smach_web}.
Its full decentralized workflow is depicted in \autoref{fig:ch3_fsm}.
After take off, each \ac{MAV} alternates between exploring a predefined area and greedily picking up and delivering the closest object.

\begin{description}[font=\normalfont]
\item[TAKE OFF:] a consistency checker verifies the state estimation concerning drift, \ac{RTK} \ac{GPS} fix, etc., and takes the \ac{MAV} to a predefined exploration altitude. After taking off, the \ac{MAV} state switches to ``EXPLORE''.
\item[EXPLORE:] the \ac{MAV} follows a predefined zig-zag exploration path at a constant altitude using the waypoint navigator.
While searching, the object tracker uses the full resolution camera image to find targets from heights between $4 \unit{m}$ and $7 \unit{m}$.
If one or more valid targets are detected, the closest target is locked by the tracker and the \ac{MAV} state switches to ``PICK UP''.
A valid target is one that is classified as small object, lies within the assigned arena bounds and outside the drop box, and has not had too many pickup attempts.

The mission is terminated if the battery has low voltage. The \ac{MAV} state switches to ``LAND'', which takes it to the starting position. As challenge resets with battery replacement were unlimited, this state was never entered during the challenge.
\item[PICK UP:] the object tracking and servoing algorithms run concurrently to pick up an object.
The detection processes the camera images at a quarter of their resolution to provide high rate feedback to the servoing algorithm.
If the gripper's Hall sensors do not report a successful pickup, the target was either lost by the tracker or the servoing timed out.
The \ac{MAV} state reverts to ``EXPLORE''. Otherwise, the servoing was successful and the state switches to ``DROP OFF''.
\item[DROP OFF:] the \ac{MAV} uses the waypoint navigator to travel in a straight line at the exploration altitude to a waiting point predefined based on the hard-coded drop box position.
Next, the drop container semaphore is queried until it can be locked. Once the drop box is available, the \ac{MAV} navigates above the predefined drop box position, reduces its altitude to a dropping height, and releases the object.
Then, it frees the drop box and semaphore and returns to ``EXPLORE''. A drop off action can fail if the object is lost during transport or the semaphore server cannot be reached.
It is worth mentioning that, in our preliminary \ac{FSM} design, the \ac{MAV} used a drop box detection algorithm based on the nadir camera.
However, hard coding the drop box position proved to be sufficient for minimal computational load.
\end{description}

\subsection{Multi-Agent Coverage Planning and Waypoint Navigation}
\label{sec:ch3_multi_agent}
The main requirements for multi-agent allocation in this system are, in order of decreasing priority: collision avoidance, robustness to agent failure, robustness to network errors, full coverage, simplicity, and time optimality.

\begin{figure}
\centering
\begin{subfigure}{0.49\textwidth}
  \centering
  \includegraphics{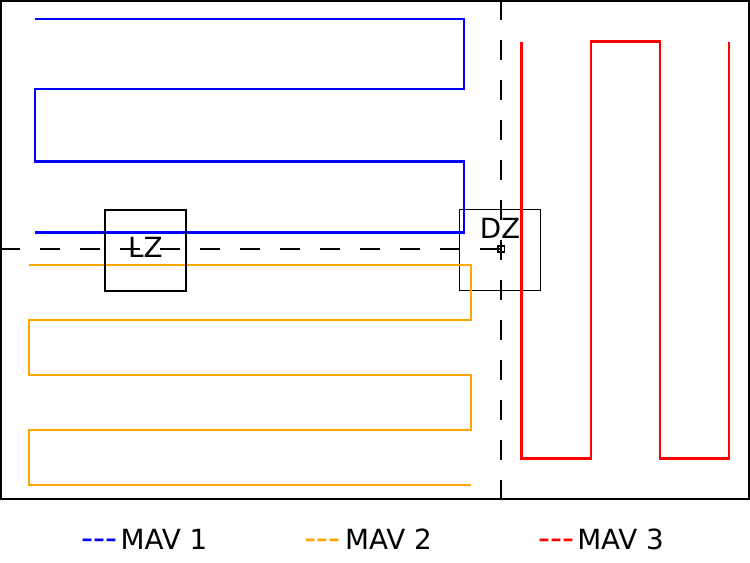}
  \caption{Three \acp{MAV}.}
  \label{fig:ch3_exploration_3}
\end{subfigure}\hfill
\begin{subfigure}{0.49\textwidth}
  \centering
  \includegraphics{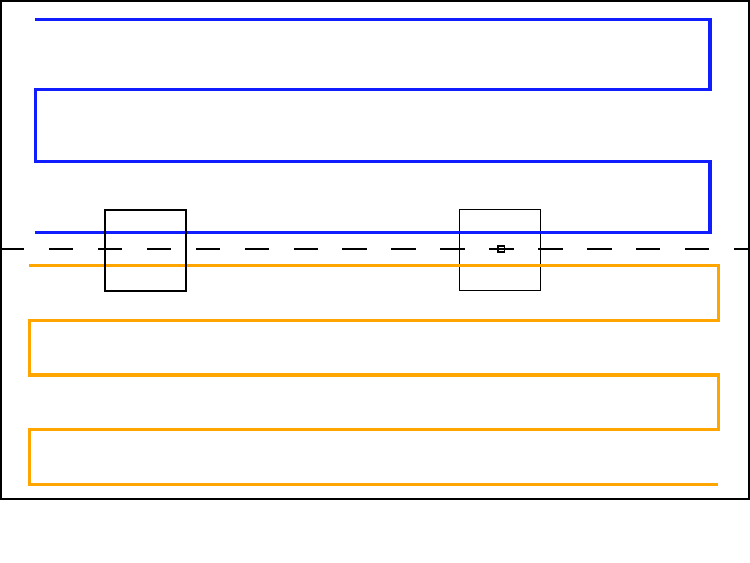}
  \caption{Two \acp{MAV}.}
  \label{fig:ch3_exploration_2}
\end{subfigure}\hfill
\begin{subfigure}{0.49\textwidth}
  \centering
  \includegraphics{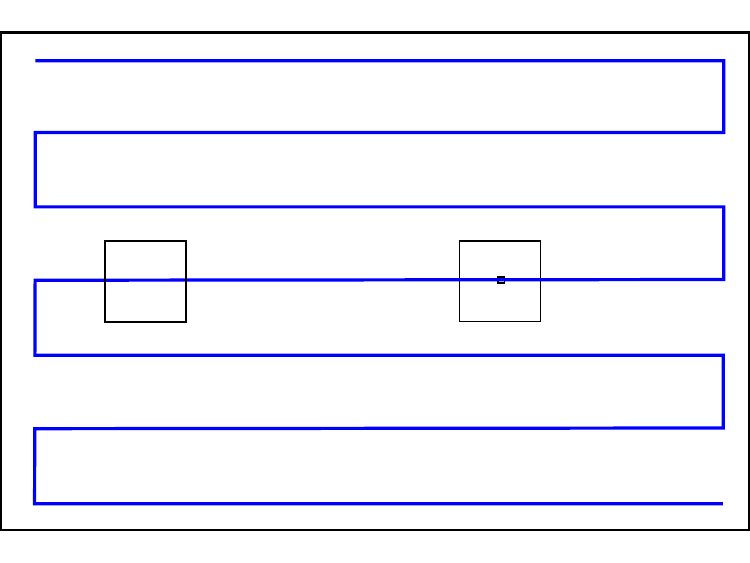}
  \caption{One \ac{MAV}.}
  \label{fig:ch3_exploration_2}
\end{subfigure}\hfill
\begin{subfigure}{0.49\textwidth}
  \centering
  \includegraphics{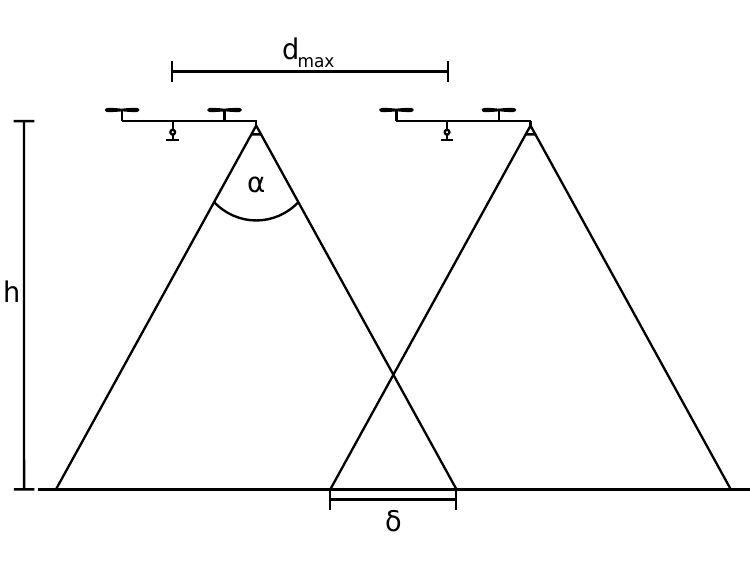}
  \caption{The sweep distance.}
  \label{fig:ch3_exploration_sweep_distance}
\end{subfigure}\hfill
\centering
\caption{For object search, the arena is divided into convex regions based on the arena corners, dropbox in the \acf{DZ}, \acf{LZ} and number of \acp{MAV}.
Each \ac{MAV} explores its region with a zig-zag path.
The maximum sweep distance $d_{\mathrm{max}}$ is a function of altitude $h$, camera \ac{FoV} $\alpha$, and overlap $\delta$.}
\label{fig:ch3_exploration}
\end{figure}

Besides using reactive collision avoidance (\autoref{sec:control}), we separate the arena into one to three regions depending on the number of \acp{MAV} as illustrated in \autoref{fig:ch3_exploration}.
Each \ac{MAV} explores and picks up objects only within its assigned region.
Additionally, we predefine a different constant navigation altitude for each \ac{MAV} to avoid interference during start and landing.
While this setup renders the trajectories collision free by construction, corner cases, such as a moving object transitioning between regions or narrow exploration paths, can be addressed by the reactive control scheme.

Our system architecture was tailored for robustness to agent failures.
The proposed decentralized system and arena splitting allows deployment of a different number of \acp{MAV} in each challenge trial and even between runs.
Automated scripts divide the arena and calculate exploration patterns.

Our design minimizes the communication requirements
to achieve robustness to network failures.
Inter-\ac{MAV} communication, i.e., odometry and drop box semaphore, is not mandatory.
This way, even in failure cases, an \ac{MAV} can continue operating its state machine within its arena region.
All algorithms run on-board the \acp{MAV}.
Only the drop box semaphore may not be locked anymore, but objects will still be dropped within the dropping zone.
Furthermore, our system does not require human supervision except for initialization and safety piloting, which does not rely on wireless connectivity.

For \ac{CPP} we implement a geometric sweep planning algorithm to ensure object detections.
The algorithm automatically calculates a lawnmower path for a given convex region as shown in \autoref{fig:ch3_exploration_sweep_distance}.
The maximum distance $d_{\mathrm{max}}$ between two line sweeps is calculated based on the camera's lateral \ac{FoV} $\alpha$, the \ac{MAV} altitude $h$, and a user defined view overlap $\delta \in [0\ldots1]$:
\begin{align}
  d_{\mathrm{max}} = (1 - \delta) \cdot 2 \cdot h \cdot \tan{\frac{\alpha}{2}}\texttt{.}
\end{align}
This distance is rounded down to create the minimum number of equally spaced sweeps covering the full polygon.
During exploration, the waypoint navigator keeps track of the current waypoint in case exploration is continued after an interruption.

In order to process simple navigation tasks, such as take off, exploration, delivery or landing,
we have a waypoint navigator module that commands the \ac{MAV} to requested poses, provides feedback on arrival, and flies predefined maneuvers, such as dropping an object in the drop box.
The navigator generates velocity ramp trajectories between current and goal poses.
This motion primitive is a straight line connection and thus collision free by construction, as well as easy to interpret and tune.

The main design principles behind our navigation framework are
simplicity, coverage completeness, and ease of restarts,
which were preferable over time optimality provided by alternative methods,
such as informative path planning, optimal decision-making, or shared workspaces.
Even when exploring the full arena with one \ac{MAV} at $5 \unit{m}$ altitude with $2 \unit{m / s}$ maximum velocity and $4 \unit{m / s^2}$ acceleration, the total coverage time is less than $5 \unit{min}$.
Furthermore, our practical experience has shown that the main difficulty in this challenge is aerial gripping, rather than agent allocation.
Our \acp{MAV} typically flew over all objects in the assigned arena,
but struggled to detect or grip them accurately.

\subsection{Object Detection, Tracking, and Servoing}
Our aerial gripping pipeline is based on visual servoing,
a well-established technique where information extracted from images is used to control robot motion.
In general, visual servoing is independent of the underlying state estimation,
allowing us to correctly position an \ac{MAV} relative to a target object without external position information.
In particular, we use a pose-based visual servoing algorithm.
In this approach, the object pose is first estimated from the image stream.
Then, the \ac{MAV} is commanded to move towards the object to perform grasping.

\begin{figure}
  \centering
  \subcaptionbox{A distorted input image.\label{fig:ch3_color}}{\includegraphics{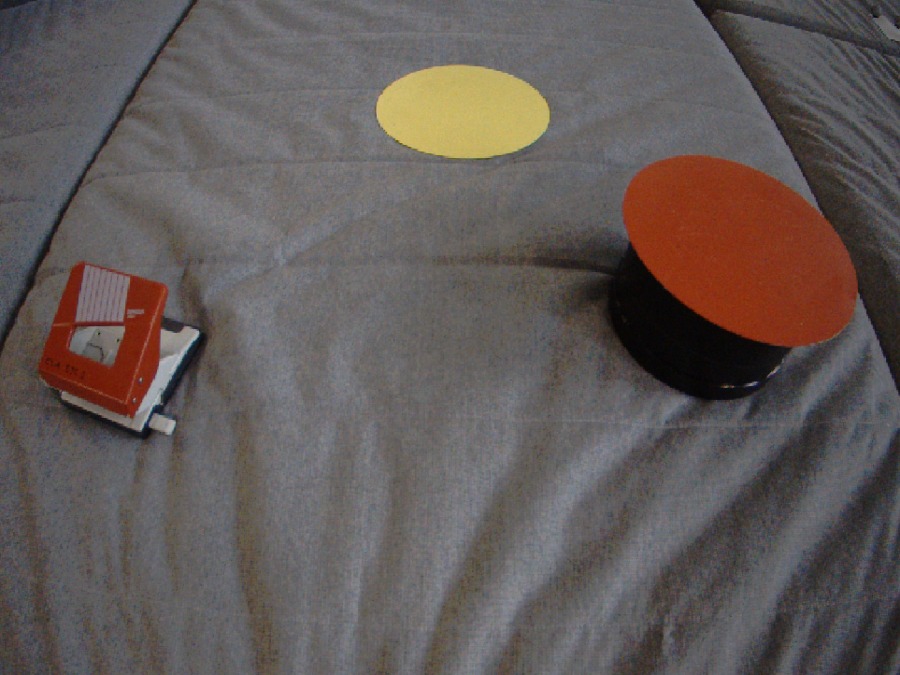}}\hfill%
  \subcaptionbox{The binary image for the red color.\label{fig:ch3_binary}}{\includegraphics{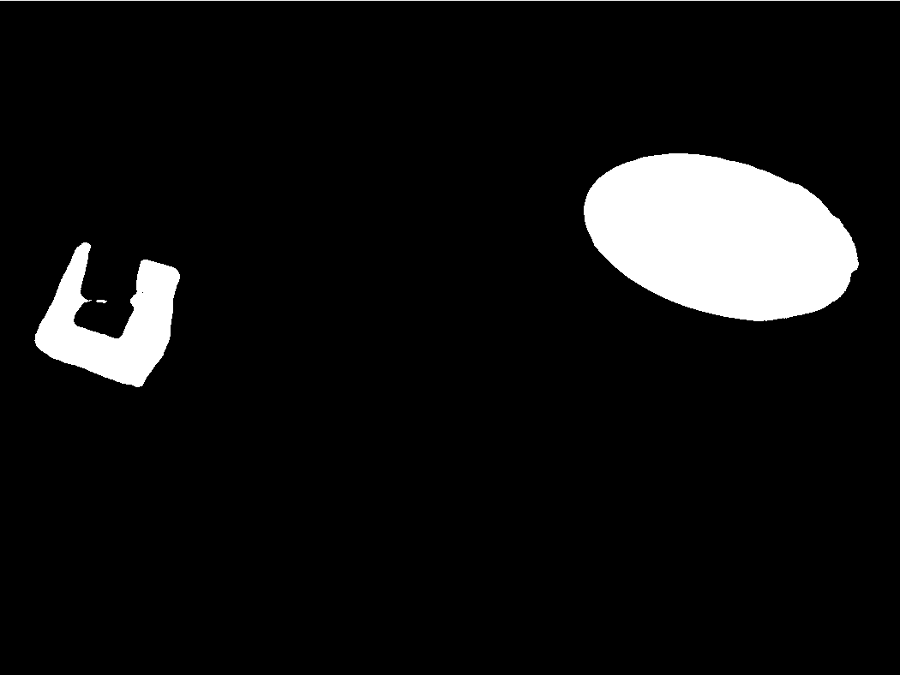}}%

  \subcaptionbox{Detections with classifications.\label{fig:ch3_classified}}{\includegraphics{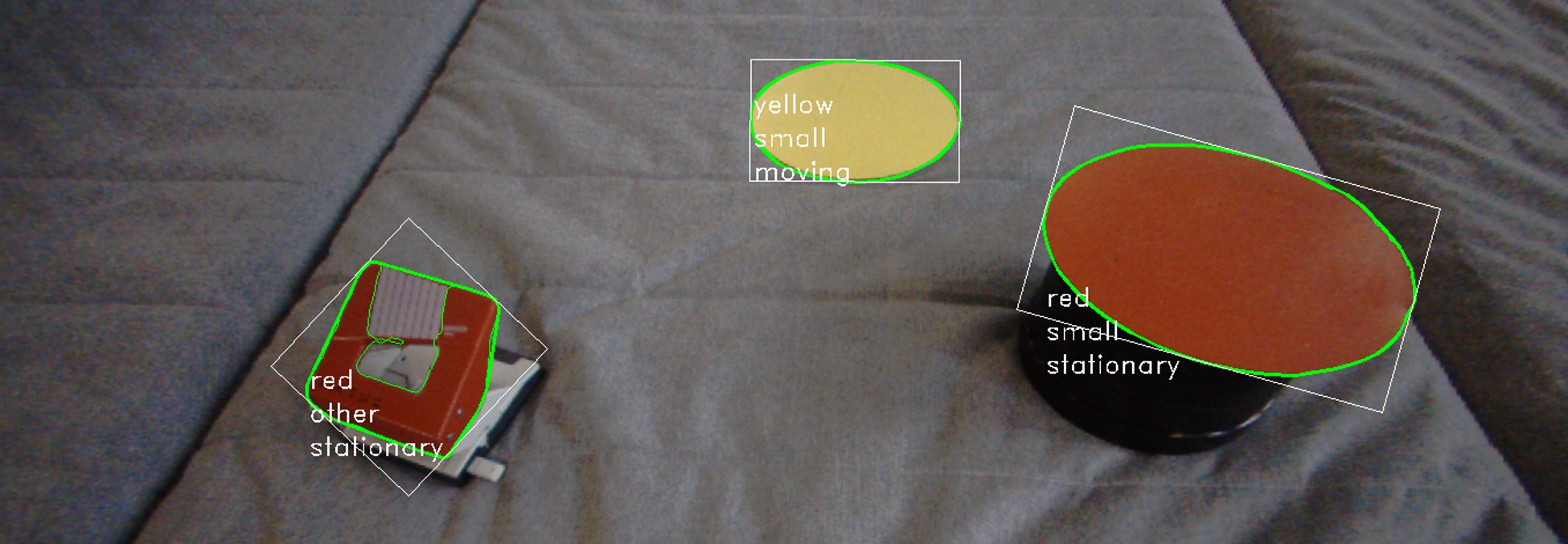}}%

\caption{The object detection pipeline. We use blob-detection and hand-crafted shape classification.}
\label{fig:ch3_detection_pipeline}
\end{figure}

The challenge rules specify the colors and shapes of the object to be found.
However, the floor material and exact color code/gloss type of the objects were unknown until the challenge start.
Thus, we decided to develop a blob-detector with hand-crafted shape classification that uses the known object specifications and can be tuned with human interpretation.
\autoref{fig:ch3_detection_pipeline} depicts an example of our image processing methods.

In order to detect colored objects, time-stamped RGB images are fed into our object detector.
The detector undistorts the images and converts its pixel values from RGB to CIE L*a*b* color space.
For each specified object color, we apply thresholds on all three image channels to get the single channel binary image. After smoothing out the binary images using morphological operators, the detections are returned as the thresholded image regions.

For each detection, we compute geometrical shape features such as length and width in pixel values, convexity, solidity, ellipse variance, and eccentricity from the contour points. We use these features to classify the shapes into small circular objects, large objects, and outliers.
A small set of intuitively tunable features and manually set thresholds serve to perform the classification.

On the first day of preparation, we set the shape parameters and tuned the coarse color thresholding parameters. Before each trial, we adjusted the camera white balance and refined the color thresholds.

\begin{figure}
  \centering

  \subcaptionbox{An object detection with image points.\label{fig:ch3_detection}}{\includegraphics{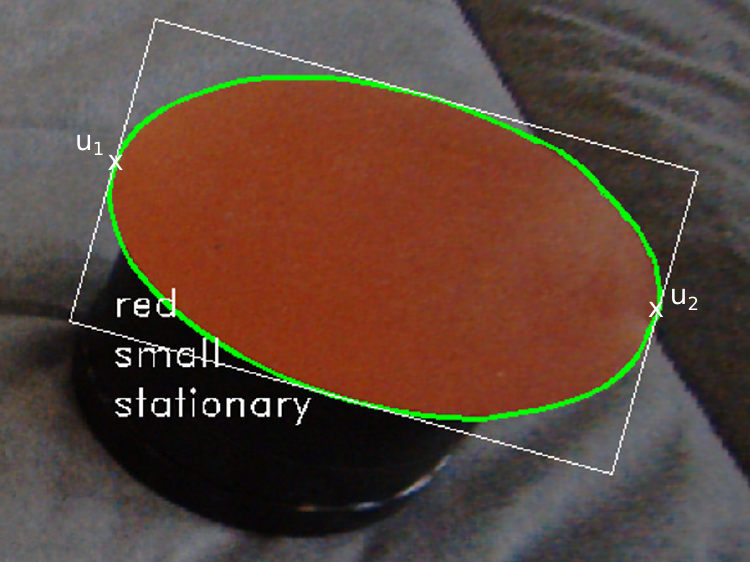}}\hfill%
  \subcaptionbox{The camera/object configuration.\label{fig:ch3_projection}}{\includegraphics{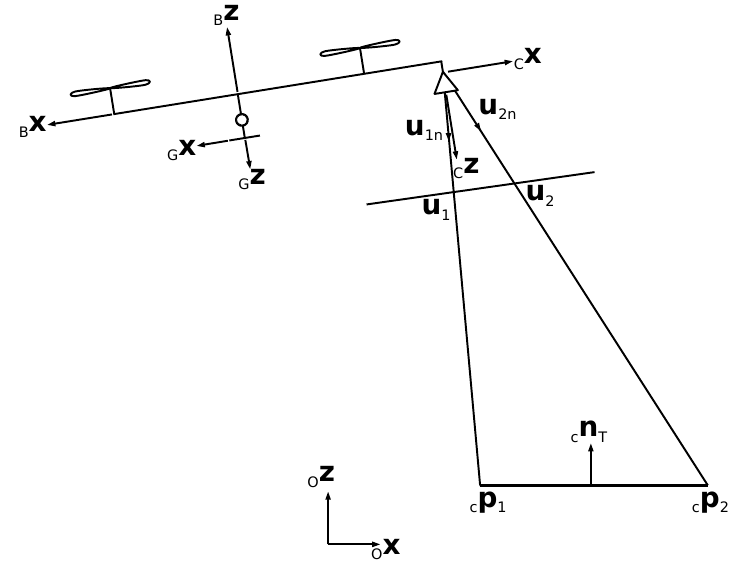}}%

  \caption{The problem of calculating the 3D positions $_C\vec{p}_1$ and $_C\vec{p}_2$ from a single monocular object detection $\vec{u}_1$ and $\vec{u}_2$.
  Assuming that the object lies on flat ground and that its physical size is known, the inverse projection problem can be solved.}
\label{fig:ch3_inv_projection}
\end{figure}

Given the 2D object detections, our aim is to find the 3D object pose for tracking.
\autoref{fig:ch3_inv_projection} displays the problem of calculating the position of two points $_C\vec{p}_1$ and $_C\vec{p}_2$ on an object with respect to the \textit{camera} coordinate frame $C$.
Assuming that the objects lie on a plane perpendicular to the gravity aligned odometry frame $z$-axis ${{_O}\vec{z}}$,
and that the metric distance $m$ between the two points is known, we formulate the constraints:
\begin{align}
 ||{_C\vec{p}_1} - {_C\vec{p}_2} || &= m, \label{eq:distance} \\
 {_C\vec{n}_T}\T \, \left( {_C\vec{p}_1} - {_C\vec{p}_2} \right) &= 0, \label{eq:perpendicular}
\end{align}
where ${_C\vec{n}_T} = \mat{R}_{CO}\,{_O\vec{z}}$ is the object normal expressed in the \textit{camera} coordinate frame and $\mat{R}_{CO}$ is the rotation matrix from the odometry coordinate frame $O$ to the \textit{camera} coordinate frame $C$.

The relation between the mapped points $\vec{u}_1$ and $\vec{u}_2$ in the image and the corresponding points ${_C\vec{p}_1}$ and ${_C\vec{p}_2}$ is expressed with two scaling factors ${\lambda_1}$ and ${\lambda_2}$ as:
\begin{align}
 {_C\vec{p}_1} &= {\lambda_1} \, \vec{u}_{1n}, \label{eq:lambda_1} \\
 {_C\vec{p}_2} &= {\lambda_2} \, \vec{u}_{2n}, \label{eq:lambda_2}
\end{align}
where $\vec{u}_{1n}$ and $\vec{u}_{2n}$ lie on the normalized image plane pointing from the \emph{focal point} to the points ${_C\vec{p}_1}$ and ${_C\vec{p}_2}$ computed through the perspective projection of the camera.
For a pinhole model, the perspective projection from image coordinates $\begin{pmatrix}u_{x}&u_{y}\end{pmatrix}\T$ to image vector $\begin{pmatrix}u_{nx}&u_{ny}&1\end{pmatrix}\T$ is:
\begin{align}
  u_{nx} &= \frac{1}{f_x} ( u_{x} - p_x ), &
  u_{ny} &= \frac{1}{f_y} ( u_{y} - p_y ),
\end{align}
where $p_x$, $p_y$ is the principal point and $f_x$, $f_y$ is the focal length obtained from an intrinsic calibration procedure \cite{furgale2013unified,kalibr_github}.

\begin{figure}
  \centering
  \includegraphics{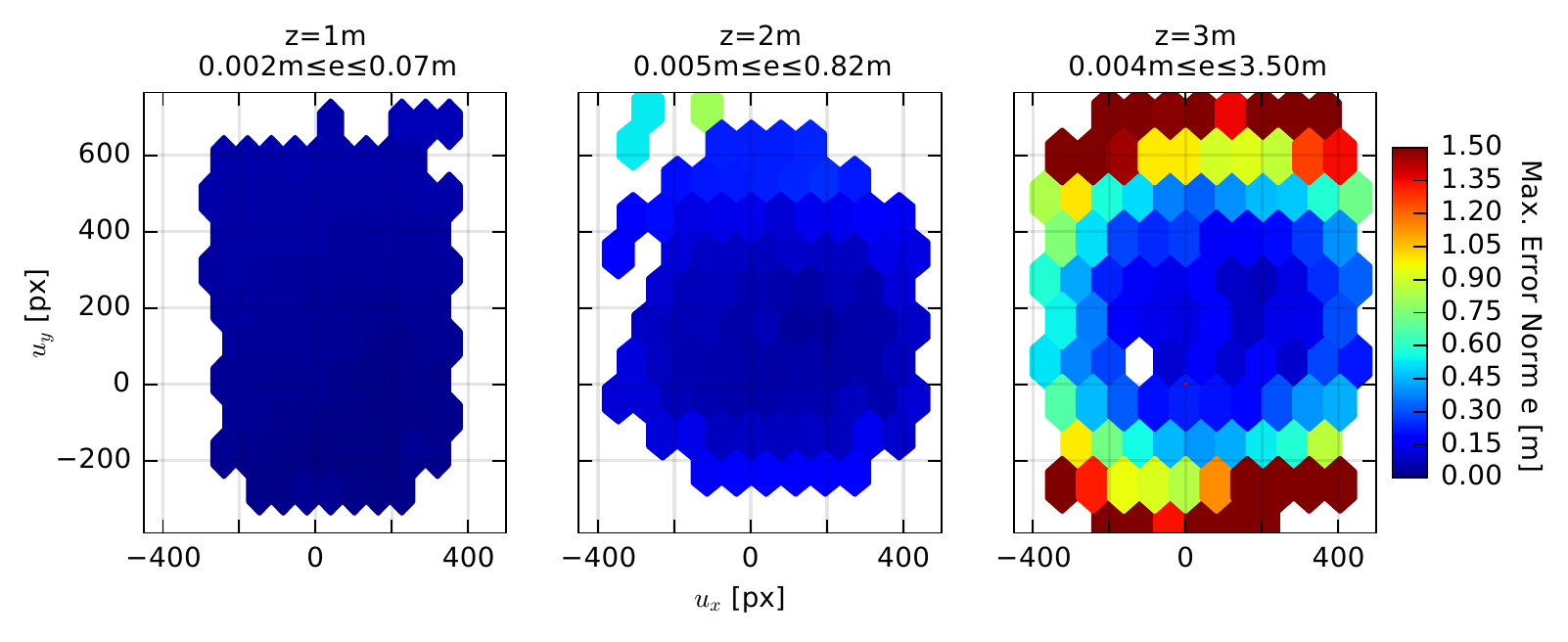}
  \caption{A Vicon motion capture validation of the detection error on the image plane.
  Best accuracy is obtained for centrally positioned, close objects, while
  detections at the image boundary and high altitudes degrade.}
  \label{fig:ch3_detection_error}
\end{figure}

Inserting \autoref{eq:lambda_1} and \autoref{eq:lambda_2} into \autoref{eq:distance} and \autoref{eq:perpendicular} and solving for ${\lambda_1}$ and ${\lambda_2}$  yields the scaling factors which allow the inverse projection from 2D to 3D:
\begin{align}
  \lambda_1 &= m\,\frac{\lvert{_C\vec{n}_T\T} \, {\vec{u}_{2n}}\rvert}{\kappa}, &
  \lambda_2 &= m\,\frac{\lvert{_C\vec{n}_T\T} \, {\vec{u}_{1n}}\rvert}{\kappa}, &
  \kappa &=
  \lVert\left({_C\vec{n}_T\T} \, {\vec{u}_{2n}}\right) \, {\vec{u}_{1n}} -
  \left({_C\vec{n}_T\T} \, {\vec{u}_{1n}}\right) \, {\vec{u}_{2n}}\rVert .
  \label{eq:ch3_inverse_reprojection_scale}
\end{align}
We consider the mean of $_C\vec{p}_1$ and ${_C\vec{p}_2}$ as the object center point in 3D.
\autoref{fig:ch3_detection_error} shows a ground truth validation of the detection error on the image plane for different camera poses.
Due to image boundary effects, image smoothing, and resolution changes, the detections at the image boundary tend to be less accurate than those near the center.
We compensate for this in the tracking and servoing pipeline.

The object's 3D position (and velocity) is tracked in a multi-target hybrid \ac{KF}.
A hybrid \ac{KF} was chosen for task since a constant sampling time for the arrival of the detections cannot be guaranteed.
The tracker first removes all outlier detections based on their classification, shape-color inconsistency, and flight altitude-size inconsistency.
It then computes the inverse projection of the detections from 2D to 3D in camera coordinate frame using \autoref{eq:ch3_inverse_reprojection_scale}.
Using the \ac{MAV}'s pose estimate and the extrinsic calibration of the camera to the \ac{MAV} \ac{IMU} \cite{furgale2013unified,kalibr_github}, it transforms the object position from the camera coordinate frame to the \textit{odom} coordinate frame.

For each observed object, a \ac{KF} is initialized to track its position and velocity.
The assignment of detections to already initialized \acp{KF} is performed in an optimal way using the Hungarian algorithm \cite{kuhn1955hungarian} with Euclidean distance between the estimated and measured position as the cost.
The filter uses a constant 2D velocity motion model for moving objects, and a constant position model for static ones.
The differential equation governing the constant position model can be written as:
\begin{align}
\dot{\vec{x}}_s(t) =
\begin{pmatrix}
\dot{p}_x(t) \\
\dot{p}_y(t) \\
\dot{p}_z(t)
\end{pmatrix} = \vec{v}_s(t) \texttt{,}
\end{align}
where $\vec{x}_s(t)$ is the position of the object with components $p_x(t)$, $p_y(t)$ and $p_z(t)$ and $\vec{v}_s(t)$ is a zero-mean Gaussian random vector with independent components.
This model was chosen since it provides additional robustness to position estimation.

Considering the prior knowledge that the objects can only move on a plane parallel to the ground, we chose a 2D constant velocity model for moving objects where the estimated velocity is constrained to the $xy$-plane.
The differential equation governing this motion can be written as:
\begin{align}
\dot{\vec{x}}_m(t) =
\begin{pmatrix}
  \dot{p}_x(t) \\
  \dot{p}_y(t) \\
  \dot{p}_z(t) \\
  \dot{v}_x(t) \\
  \dot{v}_y(t)
\end{pmatrix}
=
\begin{pmatrix}
  0 & 0 & 0 & 1 & 0 \\
  0 & 0 & 0 & 0 & 1 \\
  0 & 0 & 0 & 0 & 0 \\
  0 & 0 & 0 & 0 & 0 \\
  0 & 0 & 0 & 0 & 0
\end{pmatrix}
\vec{x}_m(t) + \vec{v}_m(t)\texttt{,}
\end{align}
where $\vec{x}_m(t)$ is the state vector with components $p_x(t)$, $p_y(t)$ and $p_z(t)$ as the position and $v_x$ and $v_y$ as the velocity.
The vector $\vec{v}_m(t)$ was again a zero-mean Gaussian random vector with independent components.

\begin{figure}
  \centering
  \includegraphics{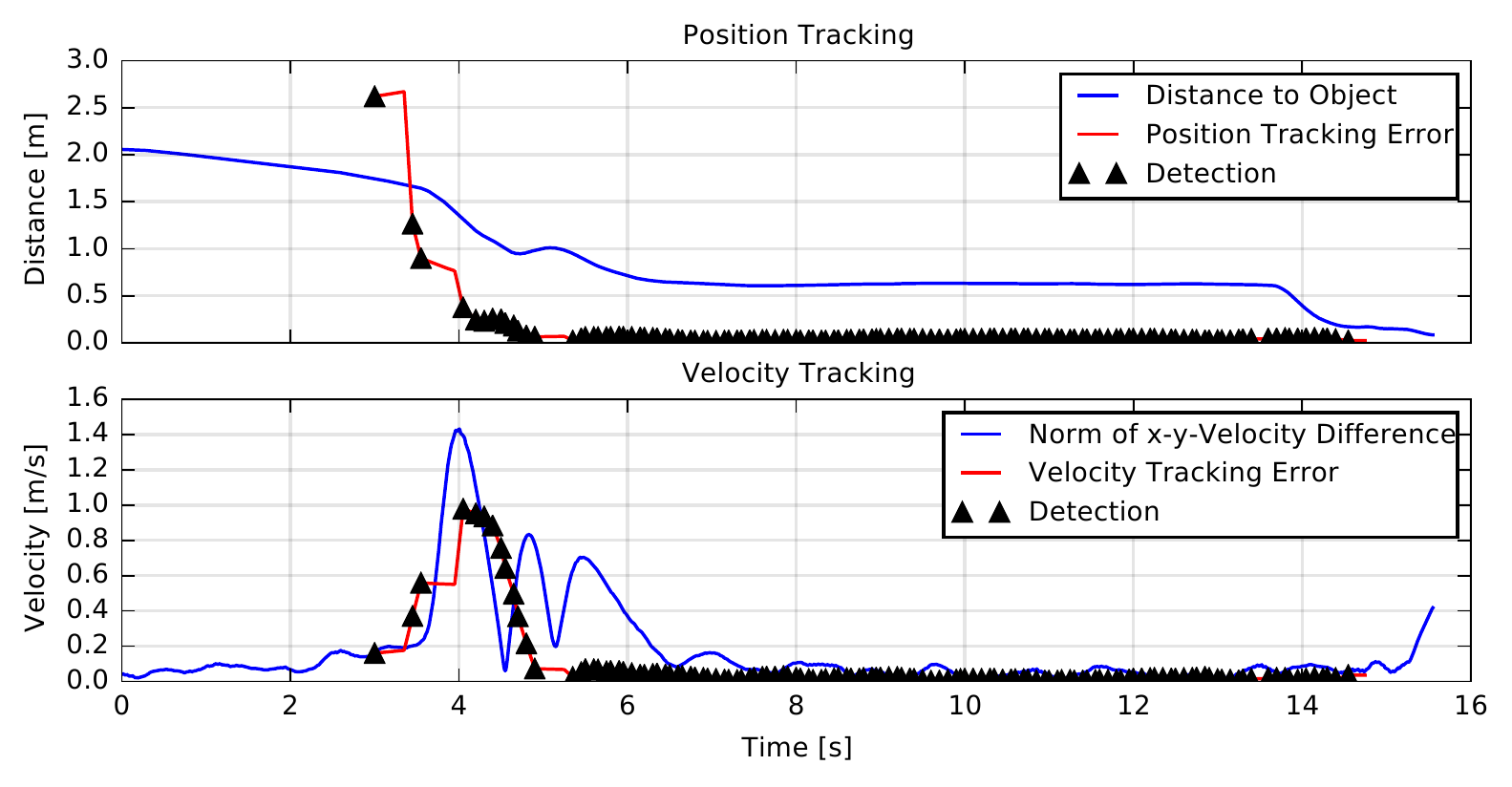}
  \caption{A Vicon motion capture validation of tracking and picking up a moving object at $1$\,km$/$h.
  Large initial tracking errors quickly converge when centering above the object.}
  \label{fig:ch3_servoing_result}
\end{figure}

\autoref{fig:ch3_servoing_result} exemplifies a ground truthed tracking evaluation.
Since an object is usually first detected on the image boundary, it is strongly biased.
We weight measurements strongly in the filter, such that the initial tracking error converges quickly once more accurate central detections occur.

\begin{figure}
  \centering
  \subcaptionbox{Entering cone.\label{fig:ch3_servoing_1}}{\includegraphics[width=0.25\textwidth]{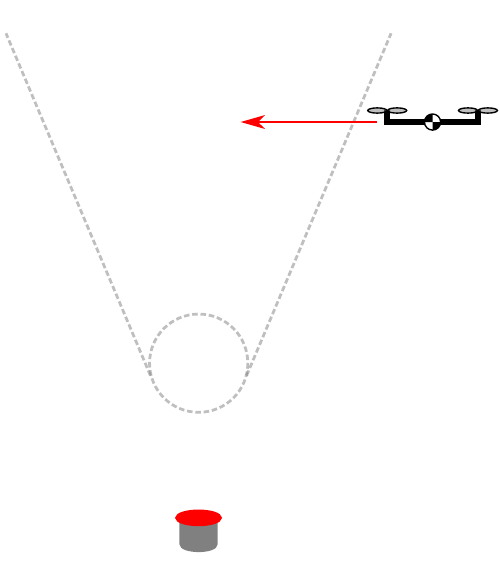}}\hfill%
  \subcaptionbox{Cone constrained descent.\label{fig:ch3_servoing_2}}{\includegraphics[width=0.25\textwidth]{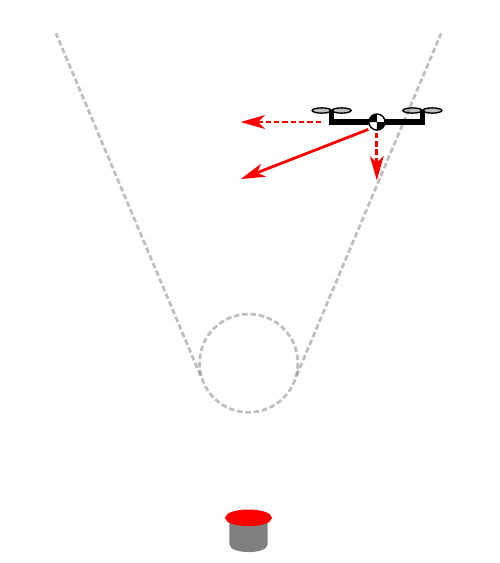}}\hfill%
  \subcaptionbox{Unconstrained descent.\label{fig:ch3_servoing_3}}{\includegraphics[width=0.25\textwidth]{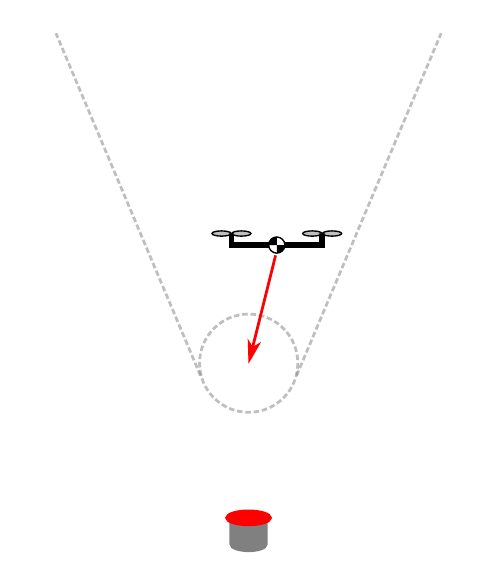}}%

  \subcaptionbox{Magnet activation.\label{fig:ch3_servoing_4}}{\includegraphics[width=0.25\textwidth]{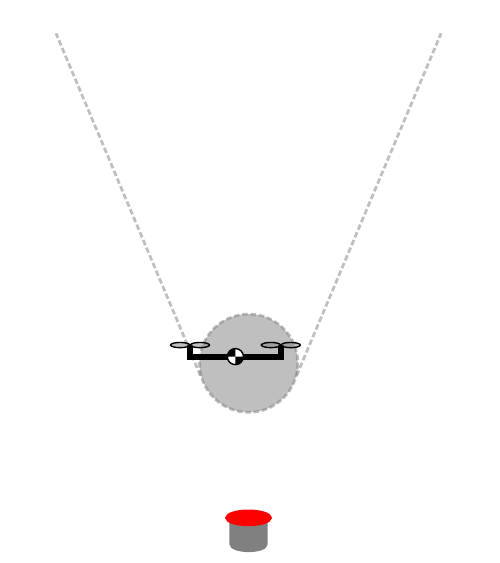}}\hfill%
  \subcaptionbox{Blind descent.\label{fig:ch3_servoing_5}}{\includegraphics[width=0.25\textwidth]{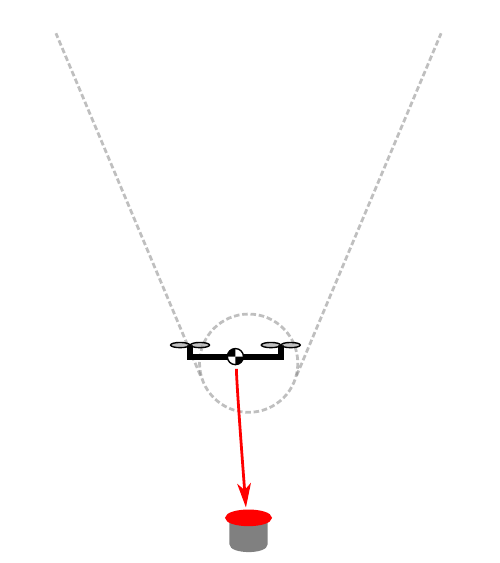}}\hfill%
  \subcaptionbox{Pick up.\label{fig:ch3_servoing_6}}{\includegraphics[width=0.25\textwidth]{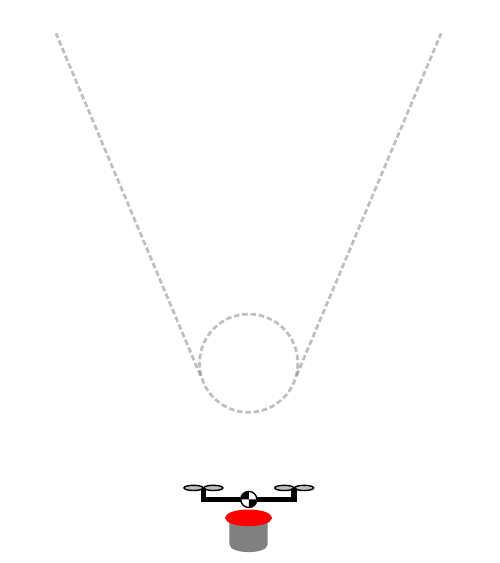}}%

\caption{The six stages of visual servoing. The cone limited descent rate ensures continuous object detections.}
\label{fig:ch3_servoing}
\end{figure}

As shown in \autoref{fig:ch3_detection_error} and \autoref{fig:ch3_servoing_result}, centering the target in the image reduces the tracking error and increases the probability of a successful pickup.
\autoref{fig:ch3_servoing} summarizes our servoing approach.
The algorithm directly sends tracked object positions and velocities to the controller relative to the gripper frame $G$.
Given its prediction horizon, the \ac{NMPC} automatically plans a feasible trajectory towards the target.
In order to maintain the object within \ac{FoV} and to limit descending motions, $z$-position is constrained such that the \ac{MAV} remains in a cone above the object.
When the \ac{MAV} is centered in a ball above the object, it activates the magnet and approaches the surface using the current track as the target position.
The set point height is manually tuned such that the vehicle touches the target but does not crash into it.
If the gripper does not sense target contact upon descent and the servoing times out, or the tracker loses the target during descent, the \ac{MAV} reverts to exploration.

\subsection{Gripper}
\label{sec:ch3_gripper}
\begin{figure}
  \centering
  \subcaptionbox{The module.\label{fig:ch3_gripper_module}}{\includegraphics{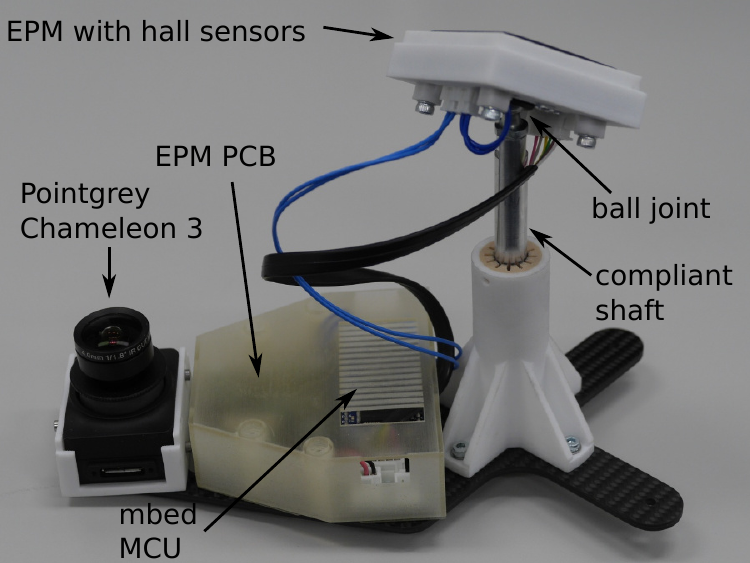}}\hfill%
  \subcaptionbox{The gripper retraction on impact.\label{fig:ch3_gripper_compliance}}{\includegraphics{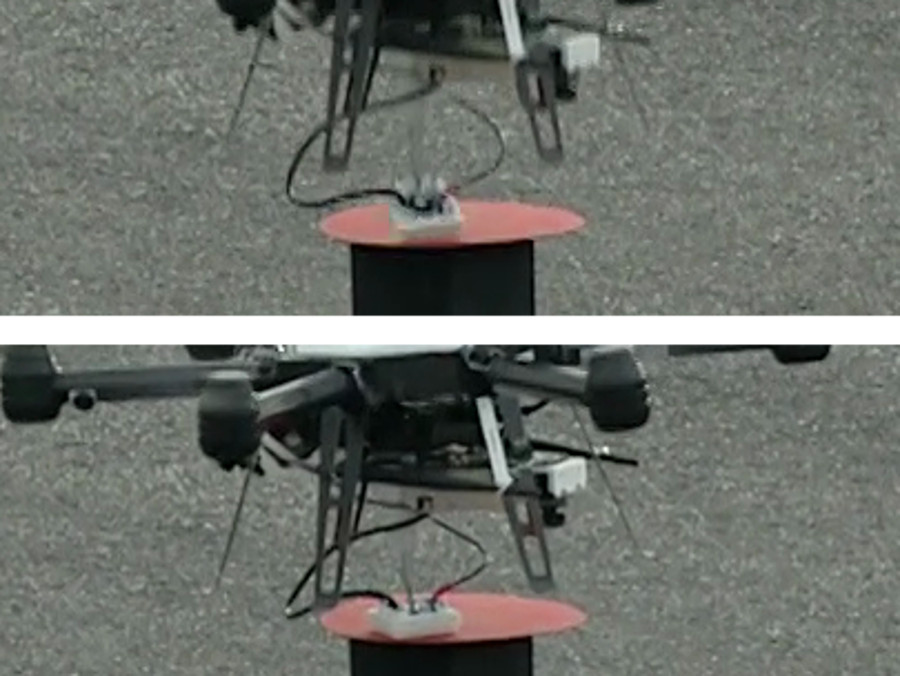}}%

\caption{The gripper-camera combination used to detect and pick-up objects.
The gripper is compliant and has Hall sensors to perceive contact.}
\label{fig:ch3_gripper}
\end{figure}

The aerial gripping of the ferromagnetic objects is complemented by an energy-saving, compliant \ac{EPM} gripper design.
Unlike regular electromagnets, \acp{EPM} only draw electric current while transitioning between the states.
The gripper-camera combination is depicted in \autoref{fig:ch3_gripper_module}.
The gripper is designed to be lightweight, durable, simple, and energy efficient.
Its core module is a NicaDrone \ac{EPM} with a typical maximum holding force of $150\unit{N}$ on plain ferrous surfaces.
The \ac{EPM} is mounted compliantly on a ball joint on a passively retractable shaft (\autoref{fig:ch3_gripper_compliance}).
The gripper has four Hall sensors placed around the magnet to indicate contact with ferrous objects.
The change in magnetic flux density indicates contact with a ferrous object.
The total weight of the setup is $250 \unit{g}$.

\section{Preparation and Development}
\label{sec:preparation}
Our preparation before the competition was an iterative process involving simulations, followed by indoor and outdoor field tests.
New features were first extensively tested in simulation with different initial conditions. This step served to validate our methods in controlled, predictable conditions.
Then, indoor tests were conducted in a small and controlled environment using the \ac{MAV} platforms to establish physical interfaces.
Finally, upon completing the previous two steps, one or more outdoor tests were executed.
The data collected in these tests were then analyzed in post-processing to identify undesirable behaviors or bugs arising in real-world scenarios.
This procedure was continuously repeated until the competition day.

\subsection{Gazebo}
\begin{figure}
  \centering
  \subcaptionbox{Gazebo-based simulation.\label{fig:gazebo_sim}}{\includegraphics{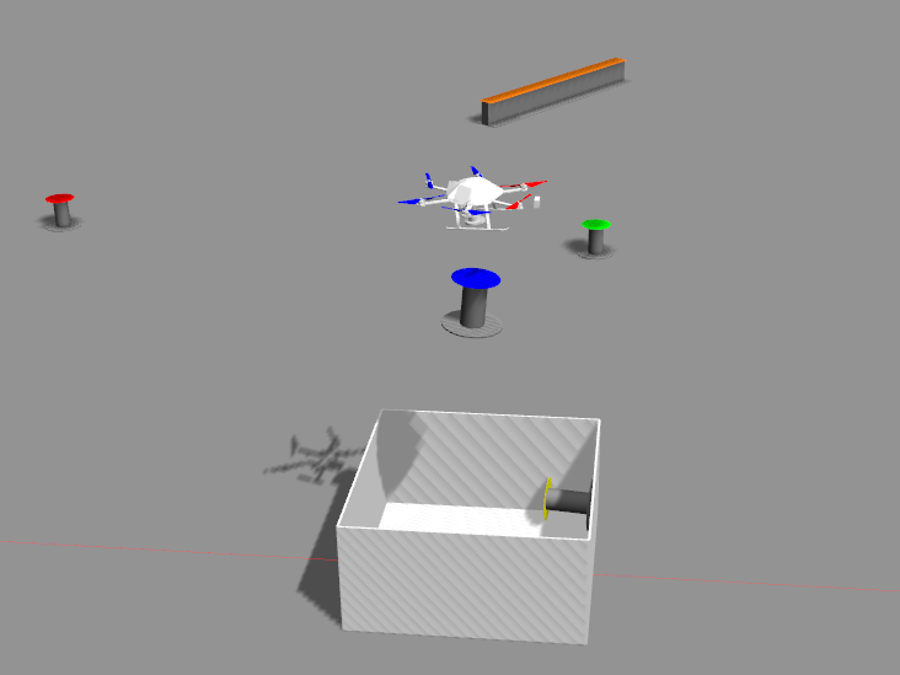}}\hfill%
  \subcaptionbox{Field test setup.\label{fig:field_test_setup}}{\includegraphics{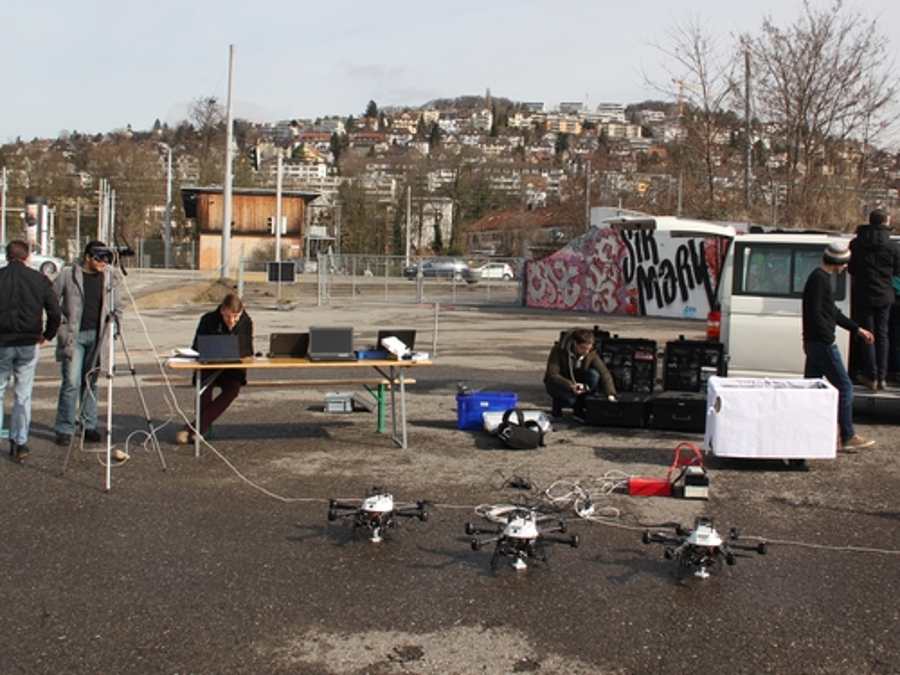}}%

\caption{Preparation before the competition. \ac{SIL} simulations and field testing.}
\label{fig:preparation}
\end{figure}

Gazebo was used as the simulation environment to test our algorithms before the flight tests \cite{gazebo_web}. An example screenshot is shown in \autoref{fig:gazebo_sim}.
We opted for this tool as it is already integrated within \ac{ROS}, providing an easily accessible interface.
Moreover, \ac{ASL} previously developed a Gazebo-based simulator for \acp{MAV}, known as RotorS \cite{Furrer2016}.
It provides multirotor \ac{MAV} models, such as the \ac{AscTec} Firefly, as well as mountable simulated sensors, such as generic odometry sensors and cameras.
As the components were designed with the similar specifications as their physical counterparts, we were able to test each module of our pipeline in simulation before transferring features to the real platforms.
Furthermore \ac{SIL} simulations decouple software-related issues from problems arising in real-world experiments,
thus allowing for controlled debugging during development.

\subsection{Testing and Data Collection}
\label{sec:testing_and_data_collection}
In the months preceding the final competition, high priority was given to testing each newly developed feature outdoors.
A specific, available location shown in \autoref{fig:field_test_setup} served as our testing area.
Through the precise accuracy of our \ac{RTK} receiver, it was possible to define and mark every relevant point in the testing area to replicate the setup of the actual arena in Abu Dhabi.
For instance, we drew the eight-shaped path and placed the take off and drop off zones.

To simplify flight trials and validate the functionalities of our \acp{MAV} in a physical environment, several extra tools were developed.
Node manager \cite{node_manager_web} is a \ac{GUI} used to manage the running and configured \ac{ROS} nodes on different hosts, such as a ground control stations and the \acp{MAV}.
Moreover, a Python \ac{GUI} was implemented to provide visual feedback to an operator checking the \ac{MAV} states.
This interface displays relevant information, such as, the status of \ac{RTK} fix, navigation waypoints or position and velocity in arena frame.

On-field data was collected using rosbags:
a set of tools for recording from and playing back \ac{ROS} topics.
During the tests, raw sensor data, such as \ac{IMU} messages, camera images, \ac{Lidar} measurements, and \ac{RTK} positions, as well as the estimated global state, were recorded on-board the \acp{MAV}.
This information could then be analyzed and post-processed off-board after the field test.

\section{Competition Results}
\label{sec:results}
We successfully deployed our \ac{MAV} system in the \ac{MBZIRC} 2017 in Abu Dhabi.
Competing against $24$ other teams for a total prize money of almost USD $2\unit{M}$, we placed second in the individual Challenge 3 and second in the Grand Challenge.
\autoref{results-summary} summarizes the results achieved during the competition.
The following sections analyze each challenge separately and summarize the lessons learned from developing a competition-based system.\footnote{See also the video: \url{https://youtu.be/DXYFAkjHeho}.}

\begin{table*}
\centering
\ra{1.3}
\begin{tabular}{@{}lllrcrrrrrr@{}}
  \toprule
            & \multicolumn{3}{c}{\textbf{Challenge 1}}                        & \phantom{abc} &  \multicolumn{6}{c}{\textbf{Challenge 3}}                          \\
   Trial    & Landed          & Intact          & Time                        & & Red   & Green        & Blue         & Yellow        & Orange  & Points           \\ \midrule
   ICT $1$  & No              & No              & -                           & & $0$   & $\mathbf{1}$ & $0$          & $\mathbf{1}$  & $0$     & $\mathbf{6}$     \\
   ICT $2$  & No              & No              & -                           & & $1$   & $1$          & $0$          & $0$           & $0$     & $3$              \\\midrule
   GCT $1$  & Yes             & Yes              & $124\unit{s}$              & & $2$   & $2$          & $0$          & $0$           & $0$     & $6$              \\
   GCT $2$  & Yes             & Yes              & $\mathbf{54}\unitbold{s}$  & & $0$   & $\mathbf{1}$ & $\mathbf{1}$ & $\mathbf{1}$  & $0$     & $\mathbf{10}$    \\
  \bottomrule
\end{tabular}
\caption{Summary of the results obtained in Challenge 1 and Challenge 3 in the \acfp{ICT} during the first two days and the \acfp{GCT} on the third day.
The columns for Challenge 3 show the objects delivered into the bin and the resulting points.
}
\label{results-summary}
\end{table*}

\subsection{Challenge 1}
\label{sec:ch1_results}
In Challenge 1 we gradually improved during the competition days to an autonomous landing time comparable to the top three teams of the individual challenge trials.
We showed start-to-stop landing within $56 \unit{s}$ and landing at $15 \unit{km / h}$ within approximately $4 \unit{s}$ after the first detection of the landing platform from a hovering state.

\subsubsection{Individual Challenge Trials}
\label{sec:ch1_ict_results}
During the two individual challenge trials, on the \num{16}th and \num{17}th March \num{2017}, the \ac{MAV} could not land on the target.
This was due to different reasons.
Before the first day, the on-board autopilot was replaced with an older, previously untested version due to hardware problems arising during the rehearsals before the first challenge run.
This caused the \ac{MAV} to flip during take off, since some controller parameters were not tuned properly.
Calibration and tuning of the spare autopilot could not be done on-site, as test facilities were not available.

During the second day of Challenge 1, the \ac{MAV} could take off and chase the platform but crashed during a high-velocity flight, as its thrust limits were not yet tuned correctly.
This led to the loss of altitude during fast and steep turn maneuvers and subsequent ground contact.
After analyzing this behavior over the first two days, the on-board parameters were properly configured and we were ready to compete in the two Grand Challenge trials.

\subsubsection{Grand Challenge Trials}
\label{sec:ch1_gct_results}
In both landing trials of the Grand Challenge, the \ac{MAV} landed on the moving platform while it was driving at the maximum speed of $15 \unit{km / h}$ (see \autoref{fig:platform_firefly}).
In the first run, the \ac{MAV} performed three attempts before being able to land. The first two attempts were automatically aborted due to weak tracker convergence and the consequential noncompliance of the Lidar safety check (see \autoref{sec:ch1_lidar}).

\begin{figure}
  \centering
  \subcaptionbox{Detection rate during challenge run.
  A detection may come from any of the two detectors but must pass the outlier rejection.
  When the platform is clearly visible, the detection rate approaches the $50 \unit{Hz}$ output rate of the camera.
  Plotted rate is smoothed by a size 3 Gaussian kernel with $\sigma=1$.
   Upwards facing triangles mark detections using the quadrilateral detector, downwards facing triangles mark detections using the cross detector.\label{fig:ch1_results_a}}{\includegraphics{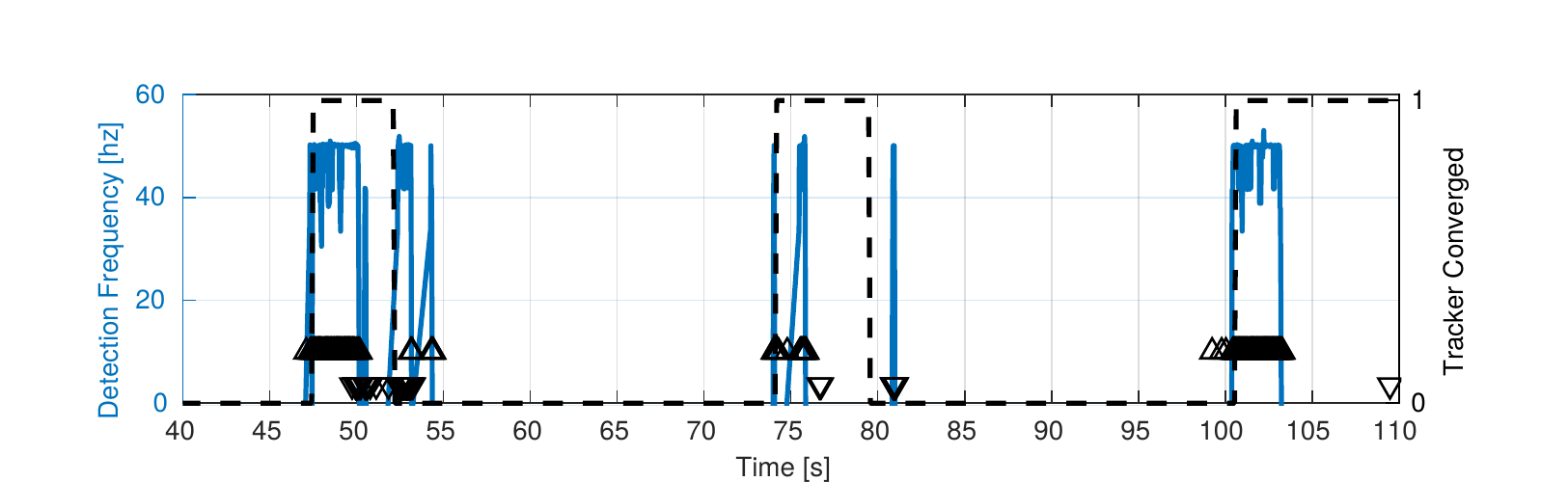}}%

 \subcaptionbox{Position error of the \ac{MAV} and the estimated position of the platform center. The fast reduction of distance between the \ac{MAV} and the estimated target is recognizable in all three attempts.
  \label{fig:ch1_results_b}}{\includegraphics{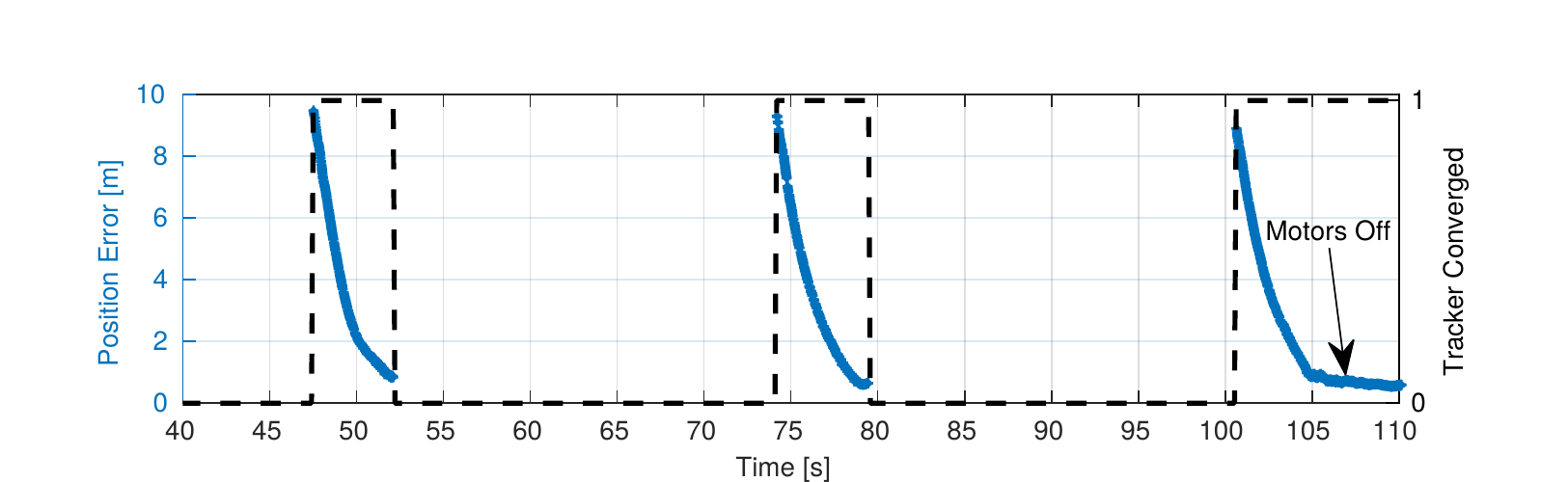}}%

  \subcaptionbox{ Magnitude of the \ac{MAV} velocity vector.
  Periods with almost zero velocity indicate that the \ac{MAV} was hovering above the center of the arena, waiting for the landing platform to enter its \ac{FoV}.
  \label{fig:ch1_results_c}}{\includegraphics{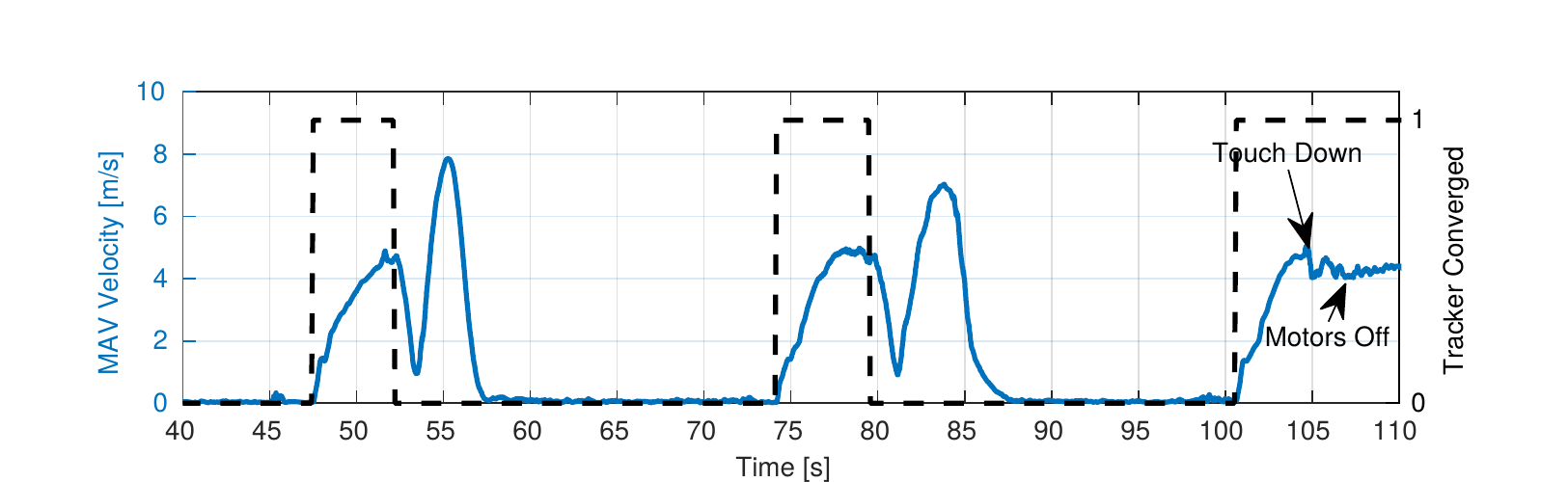}}%
\caption{Challenge 1: On-board data collected during the first Grand Challenge trial.
  Time starts when the starting signal was given by the judges.
  The first $60 \unit{s}$ are cut since the platform was not in the \ac{MAV} \ac{FoV}.
  The first two landing attempts were automatically aborted due to weak tracker convergence and the consequential noncompliance of the Lidar safety check, whereas in the last one the \ac{MAV} landed successfully.
  In each plot, the dashed black line shows when the tracker converged.}
\label{fig:ch1_results}
\end{figure}

\begin{figure}
\centering

\includegraphics{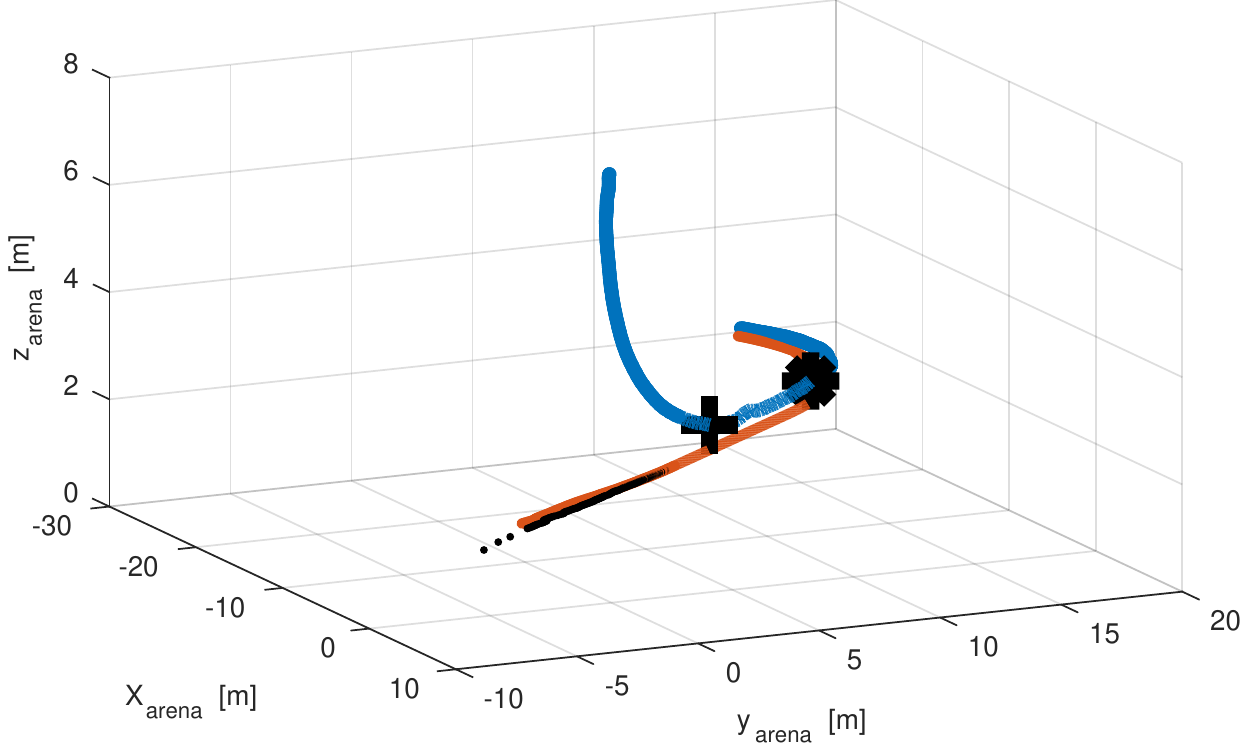}

  \caption{Final landing trajectory visualized in arena coordinates.
  The blue line corresponds to the \ac{MAV} path, and the red line to the estimated platform path.
  The black cross marks the touchdown position, while the black asterisk indicates the position where the motors have been shut down.
  The plot begins at the time of tracker convergence. Each small black dot represents a visual detection of the target by any of the two detectors.
  Touchdown occurs $4.06 \unit{s}$ after convergence and shutdown $6.33 \unit{s}$ after convergence.}
\label{fig:ch1_landing}
\end{figure}

\autoref{fig:ch1_results} shows the data collected on-board the \ac{MAV} during the first trial of the Grand Challenge.
Time starts when the starting signal was given by the judges.
The first $60 \unit{s}$ are cut since the platform was not in the \ac{MAV} \ac{FoV}.
The attempts were executed in the following time intervals: $[64.5, 69.1] \unit{s} $, $[91.2, 96.5]\unit{s}$ and $[117.6, 125.0]\unit{s}$.
\autoref{fig:ch1_results_b} shows that every time the tracker converged, the \ac{MAV} could reduce the position error below $1 \unit{m}$ in about $4 \unit{s}$.
The first two attempts were aborted approximately at $52 \unit{s}$ and at $79 \unit{s}$, due to weak convergence of the tracker module.
The third attempt started at $117.6 \unit{s}$ and after $4 \unit{s}$ the \ac{MAV} touched the landing platform, as can be seen in the spike of the velocity in \autoref{fig:ch1_results_c}, at $121 \unit{s}$.
The motors were not immediately turned off, as explained in \autoref{sec:ch1_lidar}, for safety reasons.
The ``switch-off motors'' command was triggered at $124 \unit{s}$, successfully concluding the Challenge 1 part of the Grand Challenge.
During this first run, a total of $397$ individual platform detections were performed, which are marked in \autoref{fig:ch1_results_a}.
No outliers reached the tracker.
The tracker and both detectors run at the same frequency as the camera output, $\sim 50 \unit{Hz}$.
When the platform was clearly visible, it could be detected in almost every frame, as evidenced in \autoref{fig:ch1_results_a}.
Depending on flight direction, lighting and relative motion, the total number of detections per landing attempt varies significantly.
Interestingly, the first landing attempt had a very high number of detections of both detectors, but was aborted due to a combination of a strict security criterion and a non-ideally tuned speed of the platform.
The cross detector (black marks) could detect the platform after the abortion, thus reseting the filter because the \ac{MAV} performed a fast maneuver to move back to the center of the field, effectively gaining height for a better camera viewpoint.
It would have been possible to directly re-use these measurements to let the filter converge and start chasing the platform again before moving to a safe hovering state.
However, we disabled immediate re-convergence to minimize flight conditions which might result in crash with the platform.
The second and third landing attempts prove the capability of the tracker to accurately predict the platform position - even with sparse (second) and no detection (third) during the final approach, the \ac{MAV} was able to chase respectively land safely on the platform.
\autoref{fig:ch1_landing} shows a 3D plot of the successful landing approach and its steep descent directly after tracker convergence.

In the second trial the \ac{MAV} landed in the first attempt, in $56 \unit{s}$.
Hence, we successfully demonstrated our landing approach and contributed to winning the silver medal in the Grand Challenge.

\subsection{Challenge 3}
\begin{figure}
  \centering
  \subcaptionbox{Score as a combination of objects, time, and autonomy for all four trials.
  We continuously scored in each trial and were able to beat the \acf{ICT} winner team's record with our last \acf{GCT}.\label{fig:ch3_score}}{\includegraphics{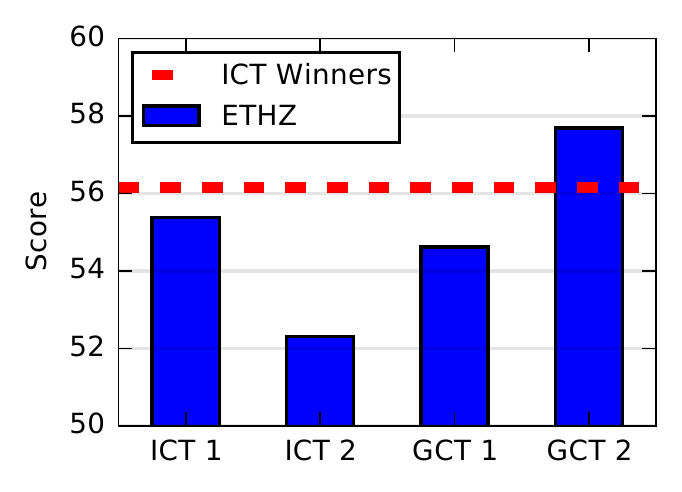}}\hfill%
  \subcaptionbox{Numbers and colors of items delivered to the bin.
  We collected up to four moving or static small objects per run.\label{fig:ch3_num_items}}{\includegraphics{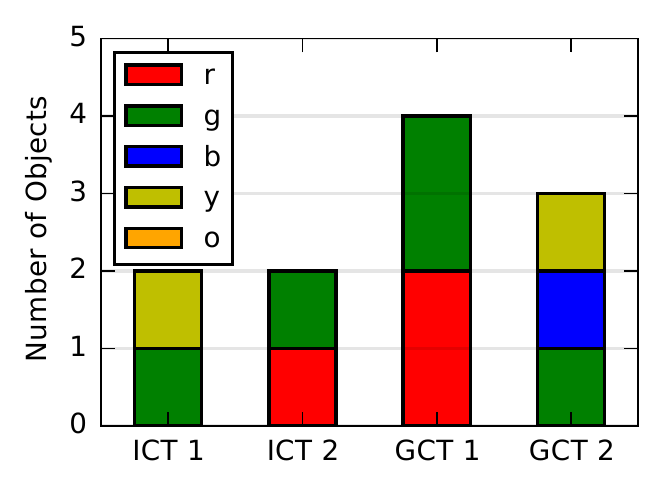}}%

  \subcaptionbox{Aerial gripping statistics of the two \acp{ICT}.
  $13$ out of $14$ servoing attempts succeeded, but due to gripper failures, only $6$ out of $13$ objects were gripped and $4$ of those delivered successfully.\label{fig:ch3_aerial_gripping_stats}}{\includegraphics{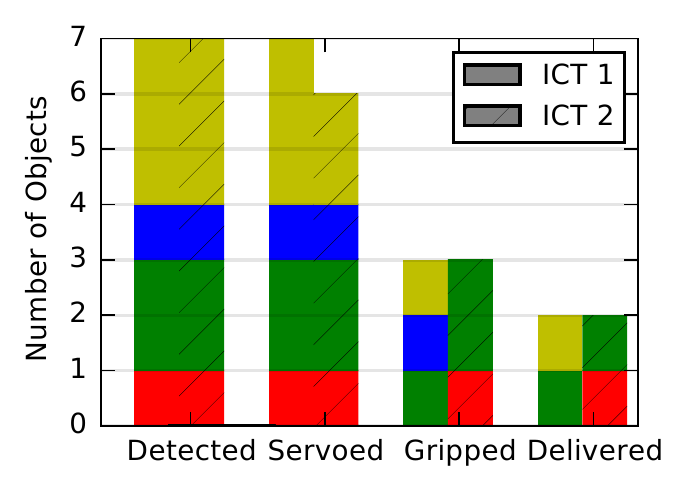}}\hfill%
  \subcaptionbox{The accumulated flight time in the two \acp{ICT}.
  Due to state estimation errors only two \acp{MAV} were engaged simultaneously, reaching a flight time of $19$ out of $120 \unit{min}$.\label{fig:ch3_flight_time}}{\includegraphics{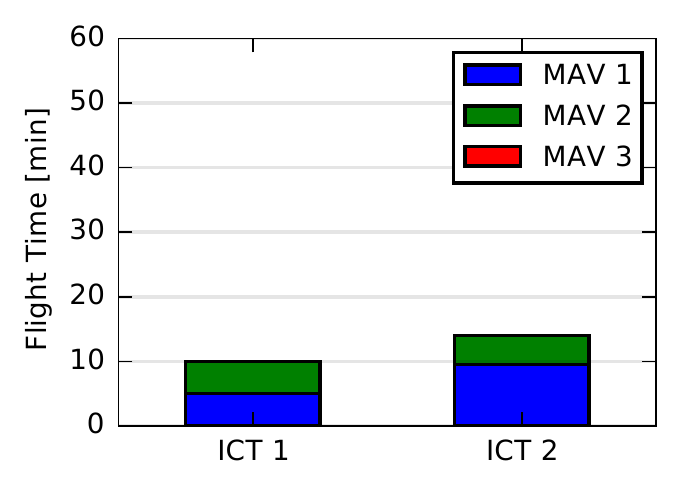}}%

\caption{Statistics for Challenge 3 from the two individual and the two Grand Challenge trials.}
\label{fig:ch3_stats}
\end{figure}

We finished second in Challenge 3 where we engaged up to three \acp{MAV} simultaneously.
Overall, we showed steady performance over all trials and were even able to surpass the individual challenge winning score during the our last Grand Challenge trial (see \autoref{fig:ch3_score}).
As \autoref{results-summary} shows, we collected both moving and static small objects and at least two and at most four objects in each trial.
Our system had a visual servoing success rate of over $90\%$.

\subsubsection{Individual Challenge Trials}
\label{sec:ch3_ict_results}
Two days before the challenge, we had two rehearsal slots of $20 \unit{min}$ each to adjust our system for the arena specifications.
On the first day, we measured the arena boundaries, set up the WiFi, and collected coarse object color thresholds.
On the second day, first flight and servoing tests were performed with a single \ac{MAV}.
During these early runs,
we noticed that the \ac{MAV} missed some object detections and that the magnet could not grip the heavier objects with paint layers thicker than our test objects at home.
We partially improved the gripping by extending the \ac{EPM} activation cycle.
Unfortunately, we were not able to find the detection issues due to missing debug data during the \ac{MBZIRC} itself.

In the first individual trial we then collected a static green object and a moving yellow object securing second place already.
In the second trial we were able to increase the accumulated flight time of our \acp{MAV} but did not improve on the number and score of objects (see \autoref{fig:ch3_num_items} and \autoref{fig:ch3_flight_time}).

\begin{figure}
 \centering
 \begin{subfigure}{0.33\textwidth}
   \centering
   \includegraphics[width=1.0\textwidth]{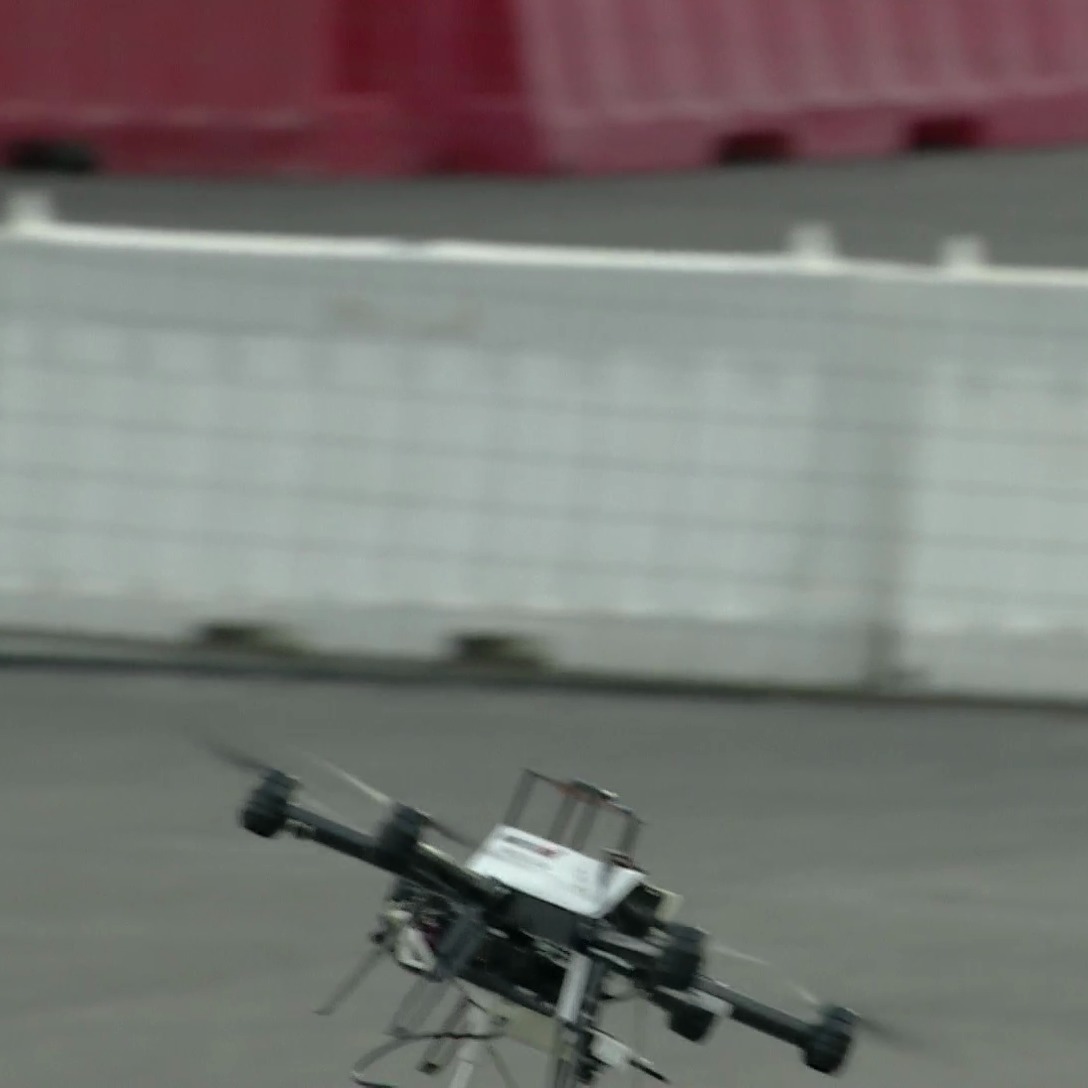}
   \caption{Sudden object detection.}
   \label{fig:ch3_seq_detect}
 \end{subfigure}\hfill
 \begin{subfigure}{0.33\textwidth}
   \centering
   \includegraphics[width=1.0\textwidth]{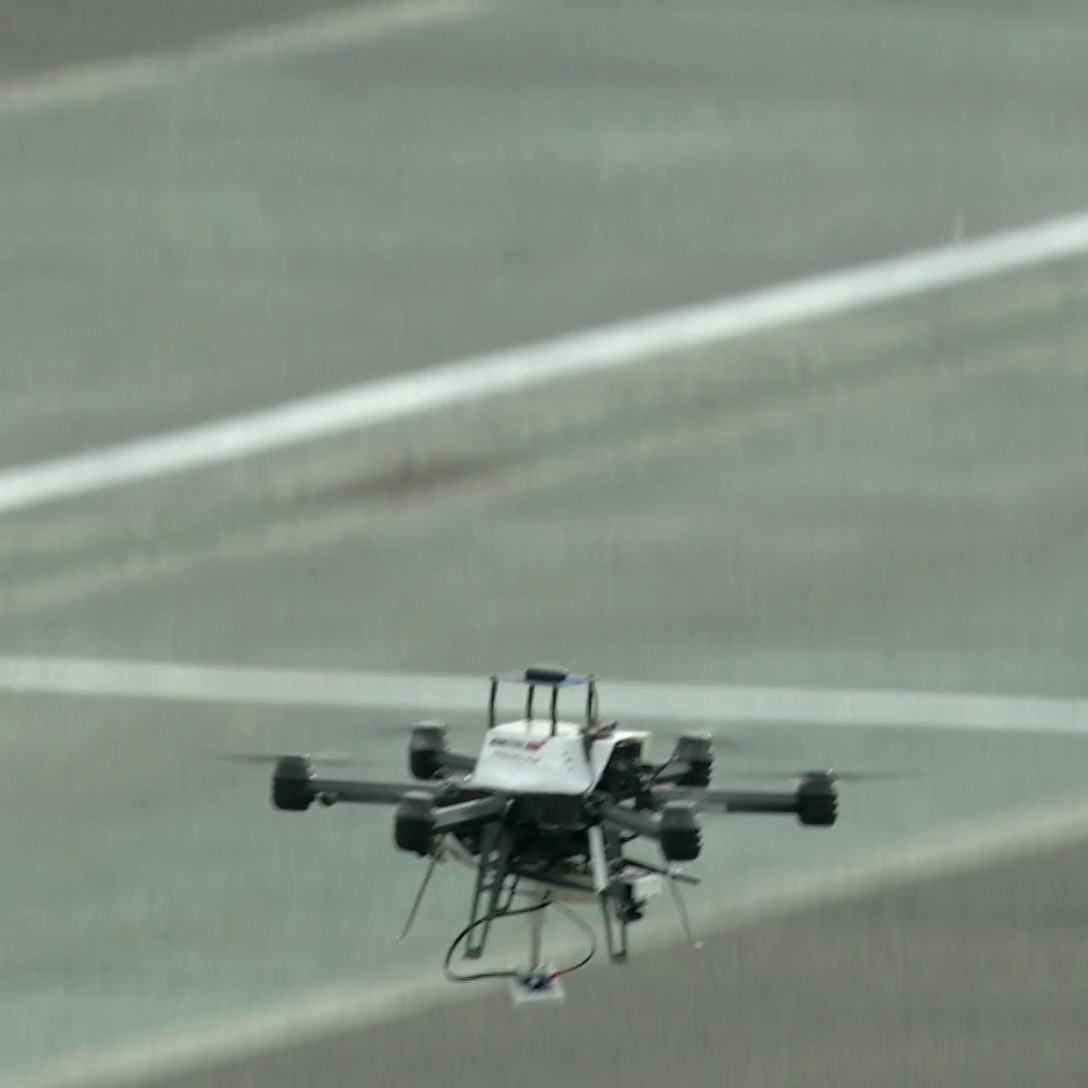}
   \caption{Descent in a cone.}
   \label{fig:ch3_seq_descent}
 \end{subfigure}\hfill
 \begin{subfigure}{0.33\textwidth}
   \centering
   \includegraphics[width=1.0\textwidth]{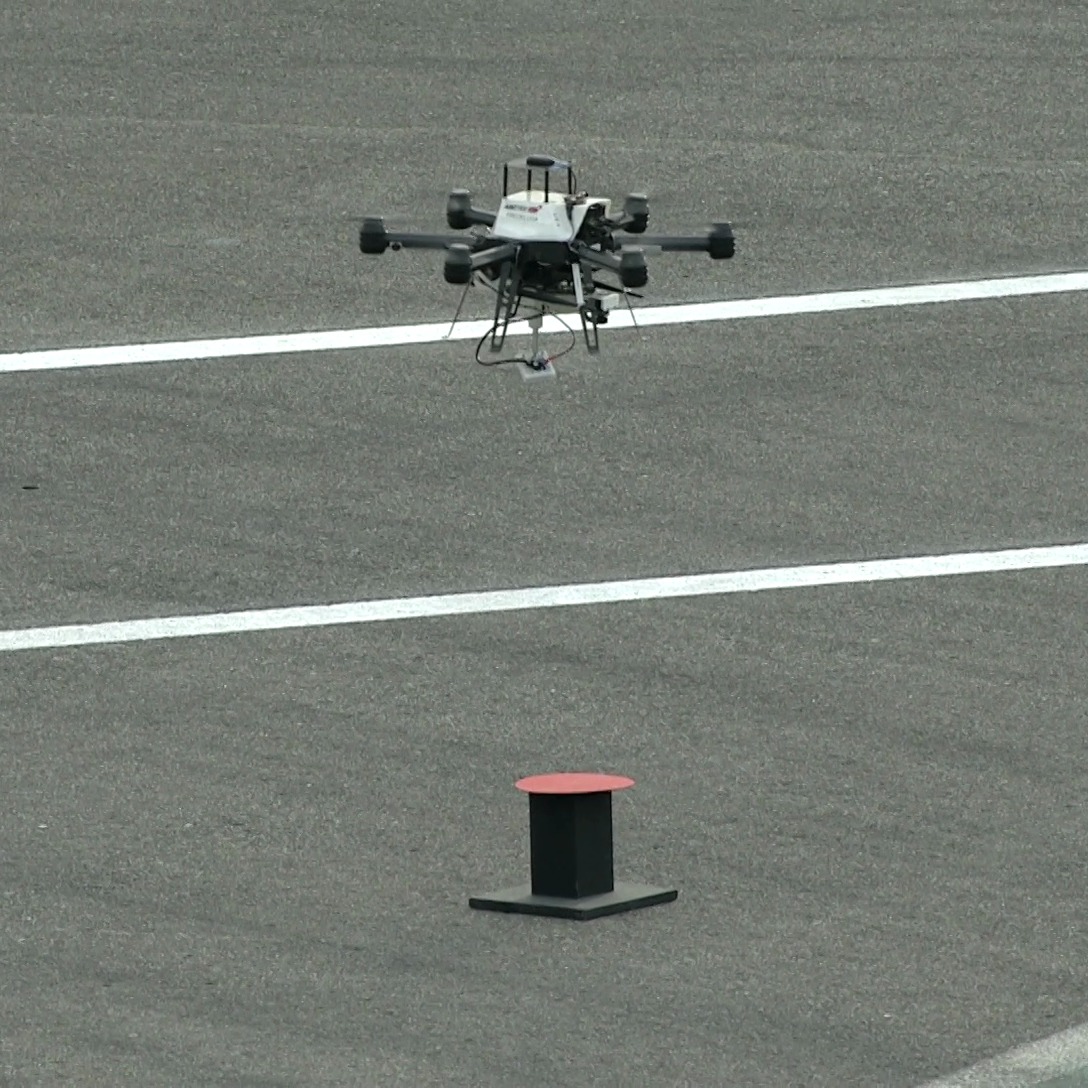}
   \caption{Center above object.}
   \label{fig:ch3_seq_center}
 \end{subfigure}\hfill
 \begin{subfigure}{0.33\textwidth}
   \centering
   \includegraphics[width=1.0\textwidth]{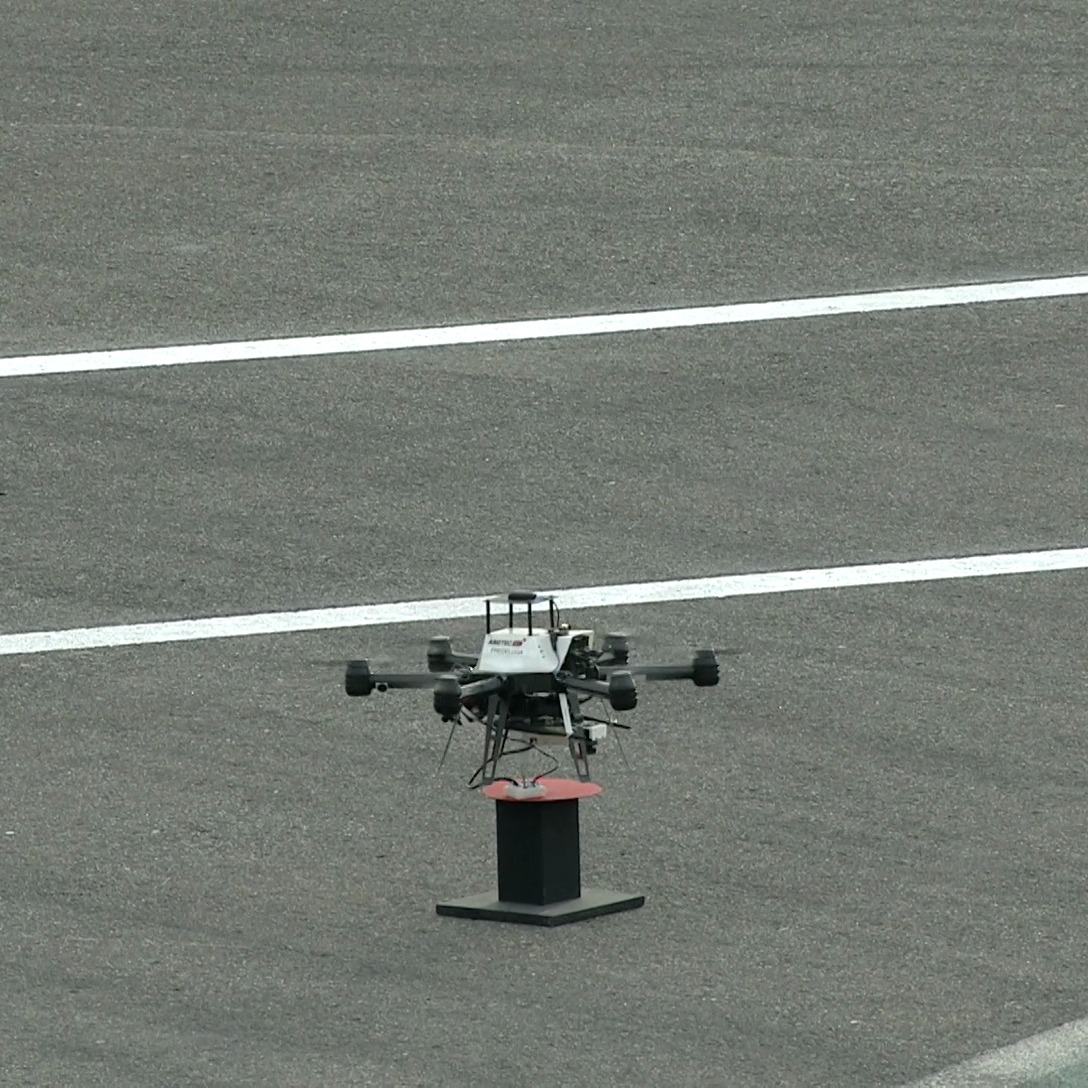}
   \caption{Grab object.}
   \label{fig:ch3_seq_contact}
 \end{subfigure}\hfill
 \begin{subfigure}{0.33\textwidth}
   \centering
   \includegraphics[width=1.0\textwidth]{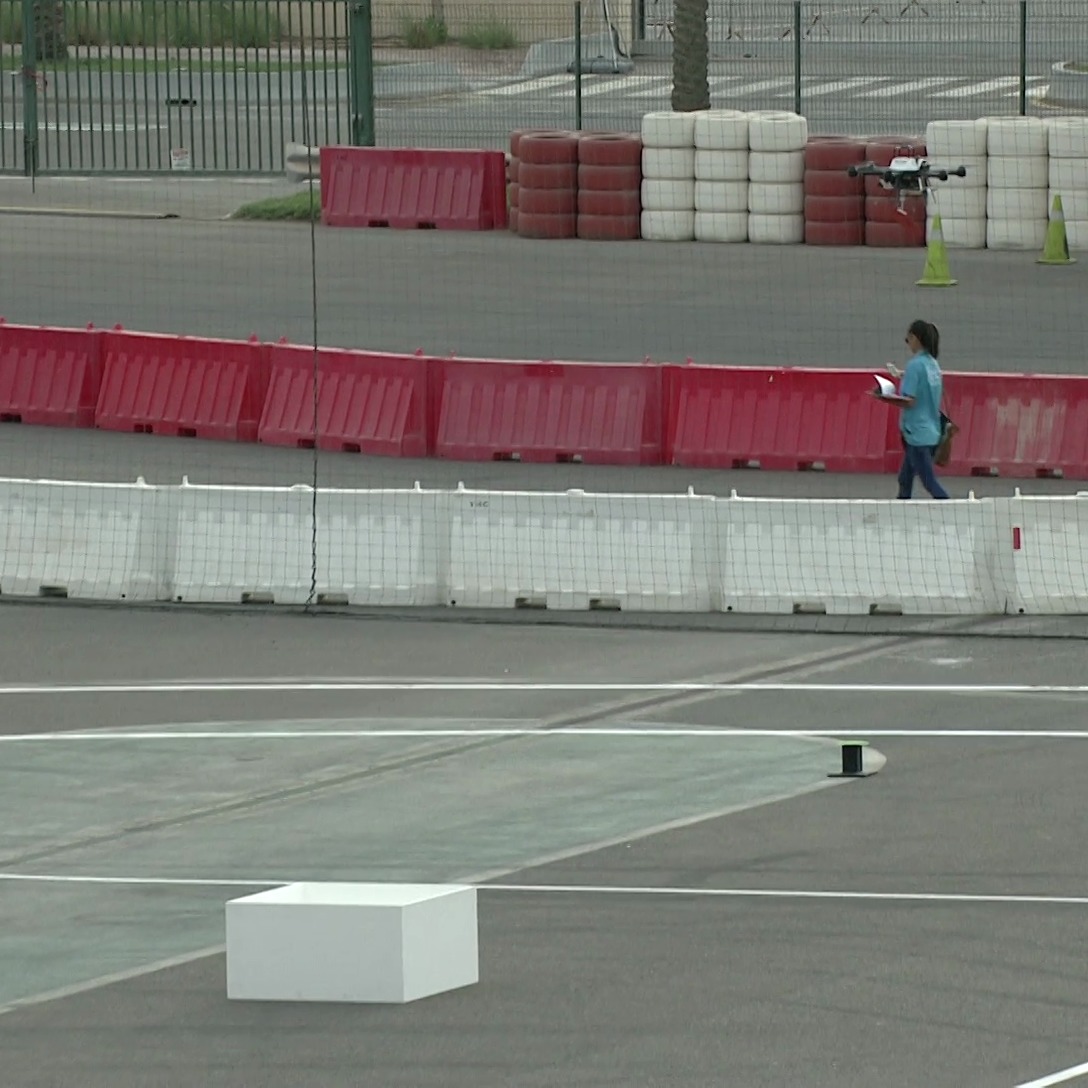}
   \caption{Wait for semaphore response.}
   \label{fig:ch3_seq_contact}
 \end{subfigure}\hfill
 \begin{subfigure}{0.33\textwidth}
   \centering
   \includegraphics[width=1.0\textwidth]{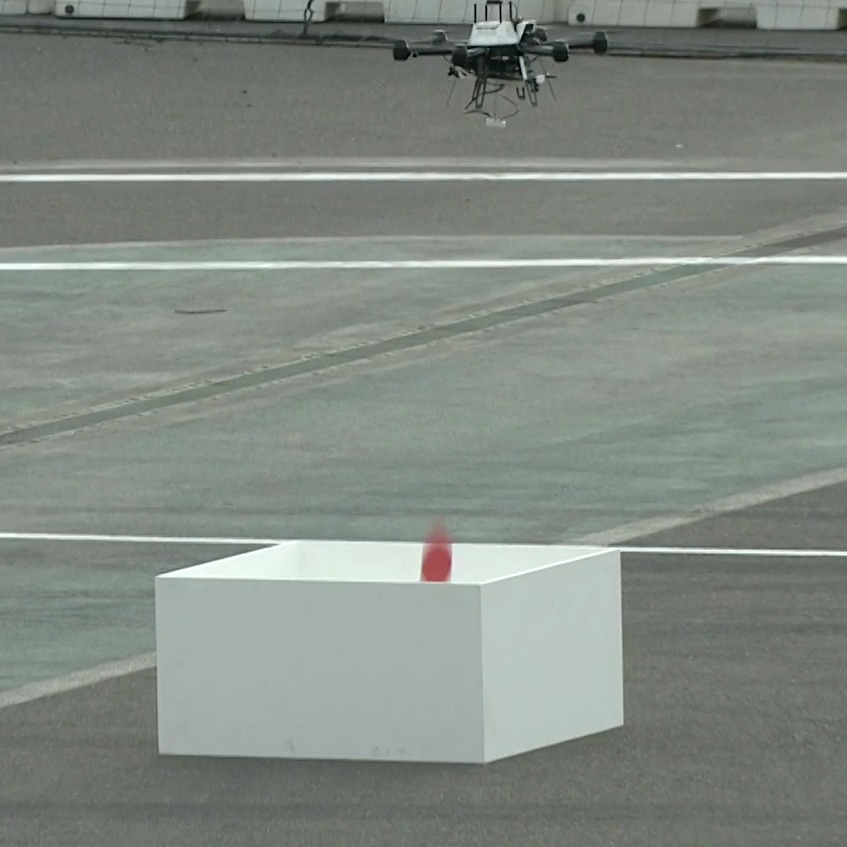}
   \caption{Drop object.}
   \label{fig:ch3_seq_drop}
 \end{subfigure}\hfill
 \centering
 \caption{An example aerial gripping sequence from the second trial of Challenge 3.
 The total time between object detection and drop off is $30 \unit{s}$.
 If an \ac{MAV} detected an object, it usually servoed it accurately.}
 \label{fig:ch3_gripping_sequence}
 \end{figure}

Despite achieving an accumulated flight time of less than $16\%$, our system still outperformed all but one team in the individual challenge.
We believe that this was mainly due to our system's accurate aerial gripping pipeline.
\autoref{fig:ch3_gripping_sequence} shows a complete $30 \unit{s}$ gripping and dropping sequence from the second trial.
When detecting an object in exploration, a waypoint above the estimated position of the object is set as reference position in the \ac{NMPC}.
In this way, the \ac{MAV} quickly reaches the detected object and starts the pickup maneuver.
The centering locks the object track.
Precise state estimation through \ac{VIO} and disturbance observance allows landing exactly on the disc even in windy conditions.

\begin{figure}
 \centering
 \begin{subfigure}{0.19\textwidth}
   \centering
   \includegraphics[width=1.0\textwidth]{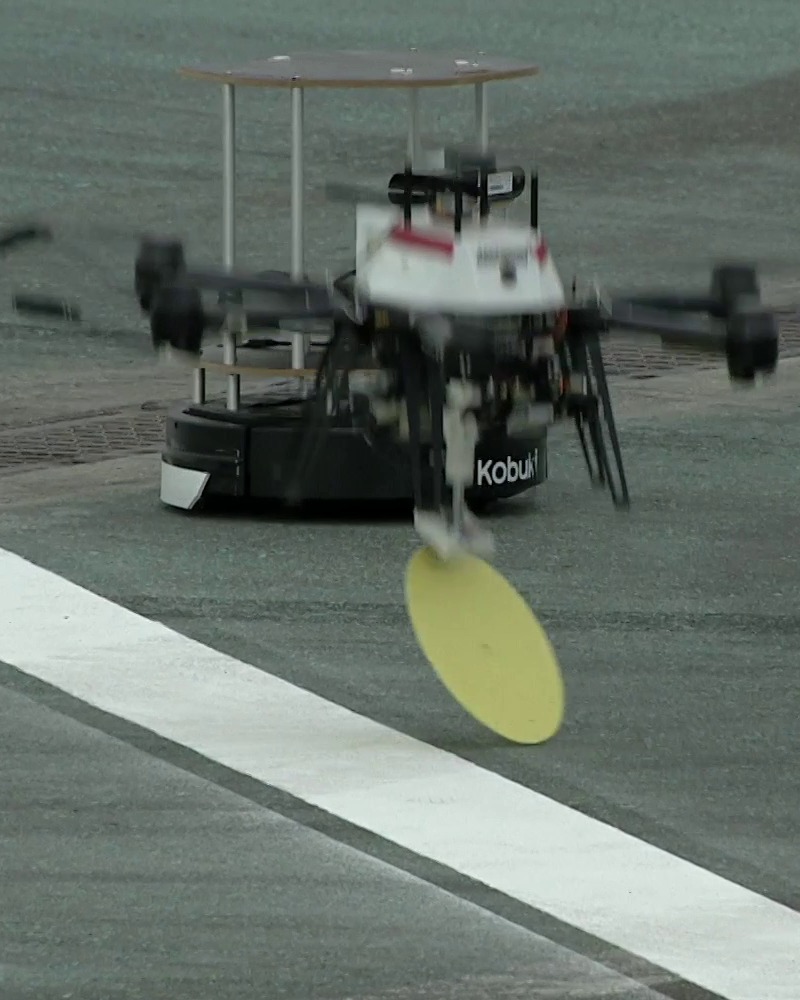}
   \caption{Weak magnet. \newline \newline}
   \label{fig:ch3_loss_magnet}
 \end{subfigure}\hfill
 \begin{subfigure}{0.19\textwidth}
   \centering
   \includegraphics[width=1.0\textwidth]{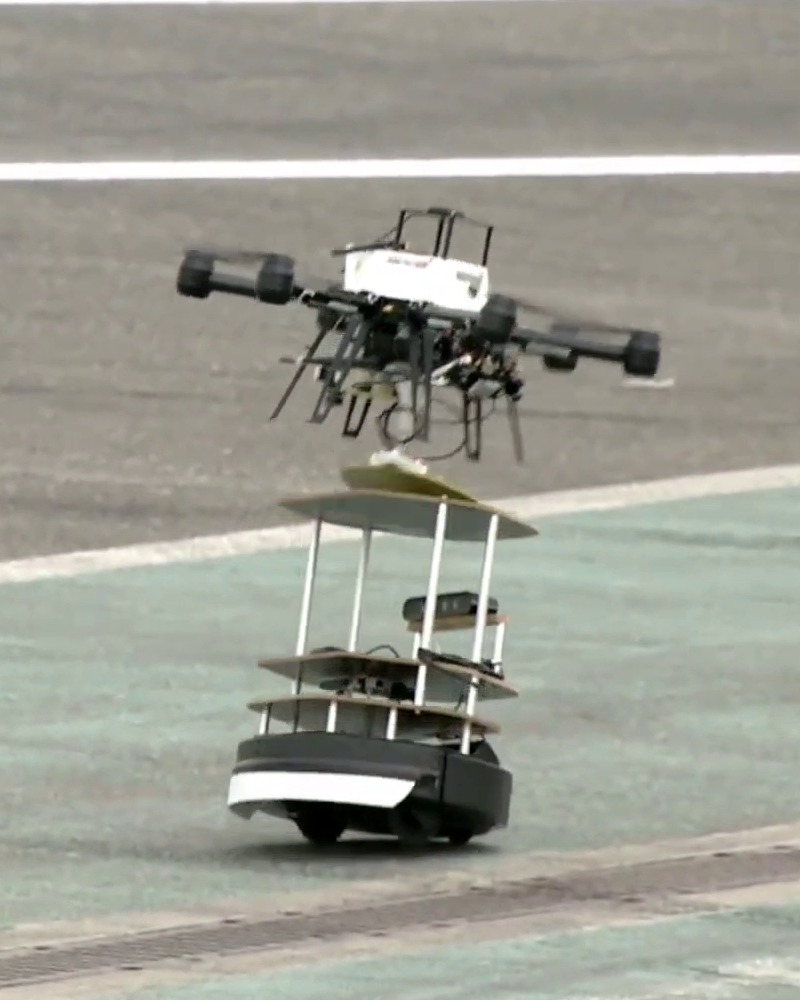}
   \caption{Shearing due to slow or erroneous Hall sensor response.}
   \label{fig:ch3_loss_sensor}
 \end{subfigure}\hfill
 \begin{subfigure}{0.19\textwidth}
   \centering
   \includegraphics[width=1.0\textwidth]{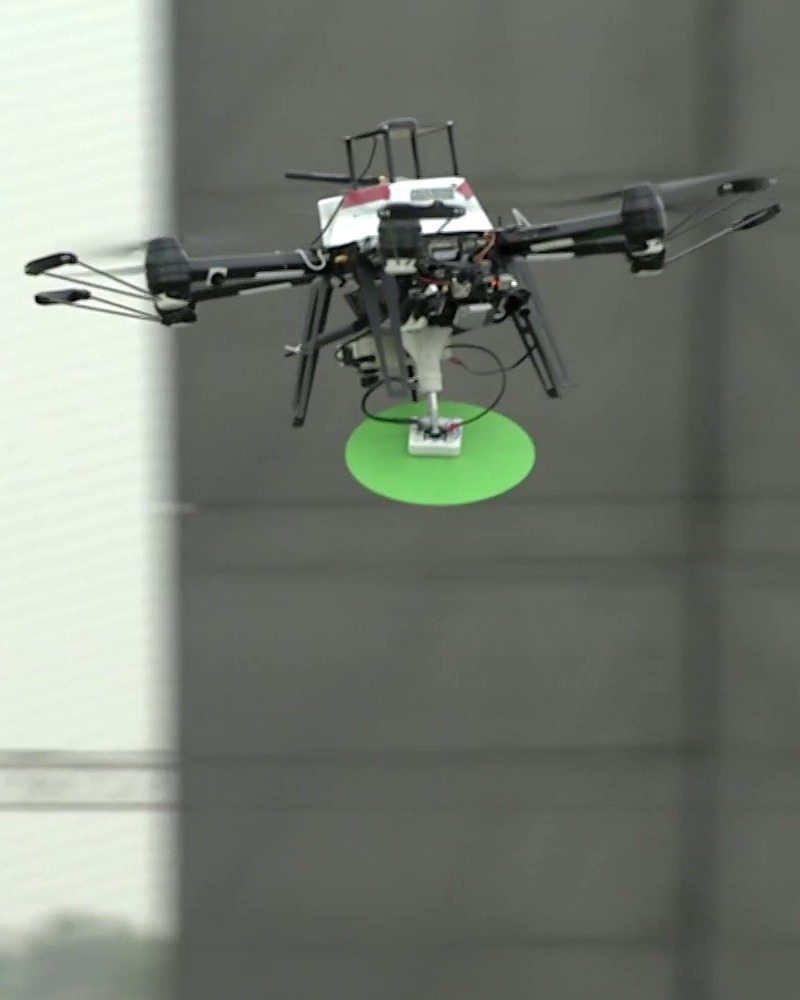}
   \caption{Object gripped but not sensed correctly.}
   \label{fig:ch3_loss_sensor_error}
 \end{subfigure}\hfill
 \begin{subfigure}{0.19\textwidth}
   \centering
   \includegraphics[width=1.0\textwidth]{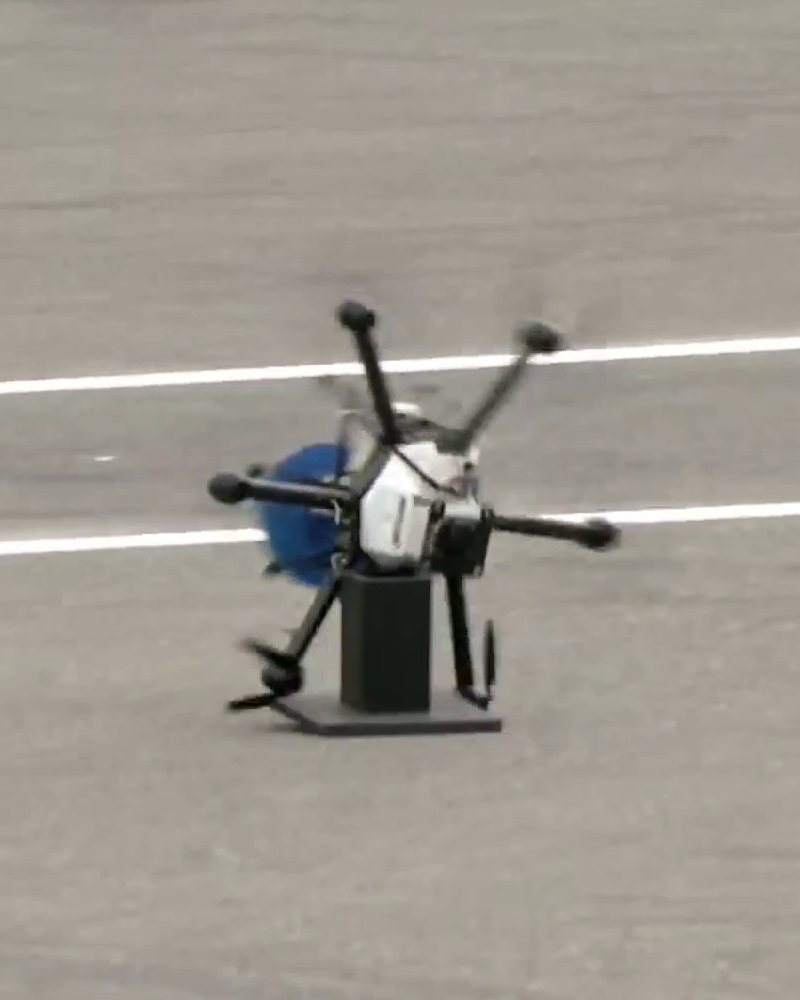}
   \caption{Missing design adaption for mechanical safety.}
   \label{fig:ch3_loss_mechanical_adaption}
 \end{subfigure}\hfill
 \begin{subfigure}{0.19\textwidth}
   \centering
   \includegraphics[width=1.0\textwidth]{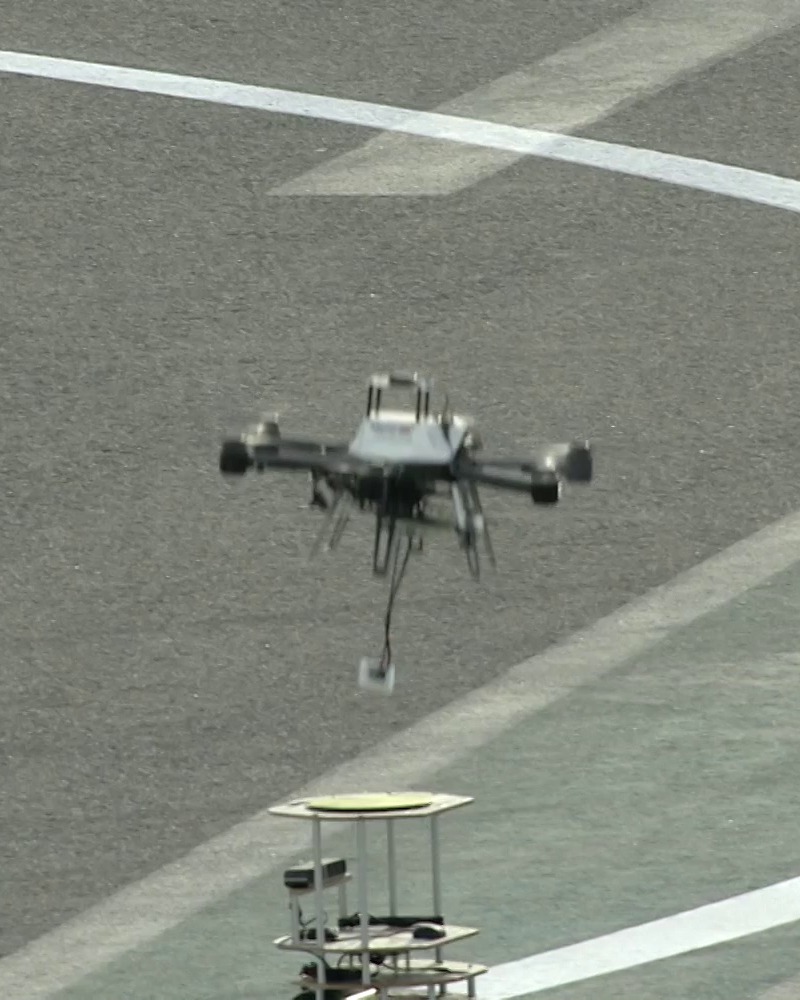}
   \caption{Lost gripper through missing mechanical robustness.}
   \label{fig:ch3_loss_mechanical_robustness}
 \end{subfigure}\hfill
 \centering
 \caption{Common aerial gripping failures sorted by descending frequency.}
 \label{fig:ch3_failures}
 \end{figure}

As shown in \autoref{fig:ch3_aerial_gripping_stats}, in the two individual challenge trials, our \acp{MAV} detected $14$ items, touched $13$ items with the magnet, picked up $6$ items, and delivered $4$ objects.
This corresponds to a servoing success rate of $93\%$ but a gripping rate of only $46\%$.
The gripping failures were mainly caused by the weak \ac{EPM} mentioned above or erroneous contact sensing.
The only missed servoing attempt resulted from a broken gripper connection (\autoref{fig:ch3_loss_mechanical_robustness}).
The two object losses after successful gripping were once caused by erroneous contact sensing (\autoref{fig:ch3_loss_sensor_error}) and once due to a crash after grabbing (\autoref{fig:ch3_loss_mechanical_adaption}).

The limited flight time was mainly due to severe CPU overload through serialization of the high resolution nadir camera image.
As the state estimation and control ran on the same processor, an overloaded CPU sometimes caused divergence of the \ac{MAV}.
Under these circumstances and given limited safety pilot capabilities, it was difficult to engage multiple \acp{MAV} simultaneously.

\subsubsection{Grand Challenge Trials}
Once we gained more experience in the system setup, we were able to increase the flight time in the Grand Challenge.
In the last trial, when we could also afford risk, we even used all three \acp{MAV} simultaneously.
Furthermore, during the Grand Challenge we deactivated the Hall sensing as it has shown to be unreliable in the individual challenge trials.
Instead we always inferred successful object pick up after servoing.
This resulted in a maximum number of four delivered objects and a maximum score that surpassed the individual contest winner's score (see \autoref{fig:ch3_score} and \autoref{fig:ch3_num_items}).

\subsection{Lessons Learned}
In general, we created two very successful robotic solutions from established techniques.
Except for the hardware and low-level control system, our entire framework was developed and integrated in-house.
We believe that especially the \ac{VIO}-based state estimation and \ac{NMPC}-pipeline differentiated our system from other competing systems which were mostly relying on Pixhawk or DJI \ac{GPS}-\ac{IMU}-fusion and geometric or PID tracking control \cite{loianno2018localization,nieuwenhuisen2017collaborative,beul2017fast,baca2017autonomous}.
However, the advantage of a potentially larger flight-envelope came with the cost of integrating and testing a more complex system and greater computational load.
Eventually, this resulted in less testing time of the actual challenge components and the complete setup in the preparation phase.

Obviously, this made \ac{SIL}-testing even more valuable.
The simulator described in \autoref{sec:testing_and_data_collection} was a great tool to test new features before conducting expensive field tests.
\ac{SIL} also allowed changing parameters safely between challenge trials where actual flying was prohibited.
Still, hardware and outdoor-tests were indispensable.
On the one hand, they reveal unmodelled effects, e.g., wind, lightning conditions, delays, noise, mechanical robustness, and computational load.
On the other hand, they lead to improvements in the system infrastructure, such as developing GUIs, simplifying tuning, and automatizing startup-procedures.
A lack of testing time induced some severe consequences in the competition.
In Challenge 1 a lot of testing was only conducted during the actual trials, where we first did not take off, then crashed, then aborted landing several times before finally showing a perfect landing (see \autoref{sec:ch1_results}).
In Challenge 3 the detection and Hall sensors failures, as described in \autoref{sec:ch3_ict_results}, could have been revealed in more extensive prior testing.

Also setting up a testing area was key in making a good transition between the home court and the competition arena.
One should put great consideration into which parts to reproduce from the challenge specifications.
On the one hand, one does not want to lose time on overfitting to the challenge, on the other hand challenge criteria has to be met.
While we spent a critical amount of time on reproducing the eight-shaped testing environment for Challenge 1, as described in \autoref{sec:testing_and_data_collection}, we missed some critical characteristics, e.g., object weight, rostrum, and arbitrary paint thickness, in testing for Challenge 3 which eventually led to failed object gripping, and missing mechanical adaptation (see \autoref{sec:ch3_ict_results}).

Furthermore, when developing the systems, we benefited from working with a modular architecture.
Every module could be developed, tested, and debugged individually.
However, this can also hide effects, that only occur when running the full system.
For example, in Challenge 3 we faced state estimation errors and logging issues due to computational overload in a (too) late phase of the development (see \autoref{sec:ch3_ict_results}).
Also, whereas testing only one platform instead of all three reduced testing effort and led to a great aerial gripping pipeline, it did not improve multi-agent engagement.

Finally, having proprietary platforms from \ac{AscTec} which were unfortunately not supported anymore, caused problems when either replacing hardware or facing software issues in the closed-source low level controller of the platform.
Eventually, this lead to the hardware failure in the first trial of Challenge 1 in \autoref{sec:ch1_ict_results} and made maintaining the three identical systems in Challenge 3 even more difficult.

\section{Conclusion}
\label{sec:conclusion}
In conclusion, our team demonstrated the autonomous capabilities of our different flying platforms by competing for multiple days both in Challenge 1, Challenge 3 and in the Grand Challenge.
The \acp{MAV} were able to execute complex tasks while handling outdoor conditions, such as wind gusts and high temperatures.
The team was able to operate multiple \acp{MAV} concurrently, owing to a modular and scalable software stack, and could gain significant experience in field robotics operations.
This article presented the key considerations we addressed in designing a complex robotics infrastructure for outdoor applications.
Our results convey valuable insights towards integrated system development
and highlight the importance of experimental testing in real physical environments.

The modules developed for \ac{MBZIRC} will be further developed and improved to become a fundamental part of the whole software framework of \ac{ASL}.
At the moment, work on manipulators attached to \acp{MAV} and cooperation strategies with a \ac{UGV} is in progress,
which will contribute towards dynamic interactions with the environment.

\ac{MBZIRC} was a great opportunity not only to train and motivate young engineers entering the field of robotics,
but also to cultivate scientific research in areas of practical application.
We think our system is a good reference for future participants.
\ac{ASL} plans to join the next \ac{MBZIRC} in \num{2019}, in which we will try to include outdoor testing earlier in the development phase.

\subsubsection*{Acknowledgments}
This work was supported by the Mohamed Bin Zayed International Robotics Challenge 2017,
the European Community’s Seventh Framework Programme (FP7) under grant agreement n.608849 (EuRoC), and
the European Union’s Horizon 2020 research and innovation programme under grant agreement n.644128 (Aeroworks) and grant agreement n.644227 (Flourish).
The authors would like to thank Abel Gawel and Tonci Novkovic for their support in the hardware design of the gripper,
Zachary Taylor for his help in field tests,
Fadri Furrer, Helen Oleynikova and Michael Burri for their thoughtful coffee breaks and code reviews,
Zeljko and Maja Popovi\'c for their warm welcome in Abu Dhabi and their filming during the challenge,
and the students Andrea Tagliabue and Florian Braun for their active collaboration.

\bibliographystyle{apalike}
\bibliography{bibliografy}

\begin{thebibliography}{}

\bibitem[Achtelik et~al., 2011]{achtelik2011onboard}
Achtelik, M., Achtelik, M., Weiss, S., and Siegwart, R. (2011).
\newblock Onboard imu and monocular vision based control for mavs in unknown
  in- and outdoor environments.
\newblock In {\em Conference on Robotics and Automation (ICRA)}. IEEE.

\bibitem[Akinlar and Topal, 2011]{edlines}
Akinlar, C. and Topal, C. (2011).
\newblock Edlines: A real-time line segment detector with a false detection
  control.
\newblock {\em Pattern Recogn. Lett.}, 32(13):1633--1642.

\bibitem[Baca et~al., 2017]{baca2017autonomous}
Baca, T., Stepan, P., and Saska, M. (2017).
\newblock Autonomous landing on a moving car with unmanned aerial vehicle.
\newblock In {\em Mobile Robots (ECMR), 2017 European Conference on}, pages
  1--6. IEEE.

\bibitem[B{\"a}hnemann et~al., 2017]{bahnemann2017decentralized}
B{\"a}hnemann, R., Schindler, D., Kamel, M., Siegwart, R., and Nieto, J.
  (2017).
\newblock A decentralized multi-agent unmanned aerial system to search, pick
  up, and relocate objects.
\newblock In {\em International Symposium on Safety, Security, and Rescue
  Robotics (SSRR)}. IEEE.

\bibitem[Beul et~al., 2017]{beul2017fast}
Beul, M., Houben, S., Nieuwenhuisen, M., and Behnke, S. (2017).
\newblock Fast autonomous landing on a moving target at mbzirc.
\newblock In {\em Mobile Robots (ECMR), 2017 European Conference on}, pages
  1--6. IEEE.

\bibitem[Blanco, 2014]{blanco2014nanoflann}
Blanco, J.~L. (2014).
\newblock nanoflann: a {C}++ header-only fork of {FLANN}, a library for nearest
  neighbor ({NN}) wih kd-trees.
\newblock \url{https://github.com/jlblancoc/nanoflann}.

\bibitem[Bloesch et~al., 2015]{rovio}
Bloesch, M., Omari, S., Hutter, M., and Siegwart, R. (2015).
\newblock Robust visual inertial odometry using a direct ekf-based approach.
\newblock {\em 2015 IEEE/RSJ International Conference on Intelligent Robots and
  Systems (IROS)}, pages 298--304.

\bibitem[Carius et~al., 2018]{mbzirc_ugv_rsl}
Carius, J., Wermelinger, M., Rajasekaran, B., Holtmann, K., and Hutter, M.
  (2018).
\newblock The {ETH-UGV} team in the {MBZ International Robotics Challenge}.
\newblock {\em Journal of Field Robotics (JFR)}.
\newblock (Under Review).

\bibitem[Chameleon 2 Datasheet, 2017]{chameleon_mono_camera}
Chameleon 2 Datasheet (2017).
\newblock Retrieved July 12, 2017, from
  \url{https://eu.ptgrey.com/chameleon-usb2-cameras}.

\bibitem[Chameleon 3 Datasheet, 2017]{chameleon_color_camera}
Chameleon 3 Datasheet (2017).
\newblock Retrieved October 09, 2017, from
  \url{https://www.ptgrey.com/support/downloads/10578}.

\bibitem[Furgale et~al., 2013]{furgale2013unified}
Furgale, P., Rehder, J., and Siegwart, R. (2013).
\newblock Unified temporal and spatial calibration for multi-sensor systems.
\newblock In {\em International Conference on Intelligent Robots and Systems
  (IROS)}. IEEE.

\bibitem[Furrer et~al., 2016]{Furrer2016}
Furrer, F., Burri, M., Achtelik, M., and Siegwart, R. (2016).
\newblock {\em Robot Operating System (ROS): The Complete Reference (Volume
  1)}, chapter RotorS---A Modular Gazebo MAV Simulator Framework, pages
  595--625.
\newblock Springer International Publishing, Cham.

\bibitem[Gadeyne, 2001]{bfl-url}
Gadeyne, K. (2001).
\newblock {BFL}: {B}ayesian {F}iltering {L}ibrary.
\newblock \url{http://www.orocos.org/bfl}.

\bibitem[Gawel et~al., 2017]{Gawel16}
Gawel, A., Kamel, M., Novkovic, T., Widauer, J., Schindler, D., von Altishofen,
  B.~P., Siegwart, R., and Nieto, J.~I. (2017).
\newblock Aerial picking and delivery of magnetic objects with mavs.
\newblock In {\em Conference on Robotics and Automation (ICRA)}. IEEE.

\bibitem[Gazebo, 2017]{gazebo_web}
Gazebo (2017).
\newblock Retrieved July 17, 2017, from \url{http://gazebosim.org/}.

\bibitem[Kalibr Github, 2017]{kalibr_github}
Kalibr Github (2017).
\newblock Retrieved July 17, 2017, from
  \url{https://github.com/ethz-asl/kalibr}.

\bibitem[Kamel et~al., 2017a]{kamel2017nonlinear}
Kamel, M., Alonso-Mora, J., Siegwart, R., and Nieto, J. (2017a).
\newblock Nonlinear model predictive control for multi-micro aerial vehicle
  robust collision avoidance.
\newblock In {\em International Conference on Intelligent Robots and Systems
  (IROS)}.

\bibitem[Kamel et~al., 2017b]{kamel2016linear}
Kamel, M., Burri, M., and Siegwart, R. (2017b).
\newblock Linear vs nonlinear mpc for trajectory tracking applied to rotary
  wing micro aerial vehicles.
\newblock In {\em International Federation of Automatic Control (IFAC)},
  volume~50, pages 3463--3469. Elsevier.

\bibitem[Kaplan, 2005]{Kaplan2005}
Kaplan, E. (2005).
\newblock {\em Understanding GPS - Principles and applications}.
\newblock Artech House.

\bibitem[Kuhn, 1955]{kuhn1955hungarian}
Kuhn, H.~W. (1955).
\newblock The hungarian method for the assignment problem.
\newblock In {\em Naval Research Logistics (NRL)}, volume~2. Wiley.

\bibitem[Lidar Datasheet, 2017]{lidar_garmin}
Lidar Datasheet (2017).
\newblock Retrieved July 12, 2017, from
  \url{http://buy.garmin.com/en-US/US/p/557294}.

\bibitem[Loianno et~al., 2018]{loianno2018localization}
Loianno, G., Spurny, V., Baca, T., Thomas, J., Thakur, D., Hert, D., Penicka,
  R., Krajnik, T., Zhou, A., Cho, A., et~al. (2018).
\newblock Localization, grasping, and transportation of magnetic objects by a
  team of mavs in challenging desert like environments.
\newblock {\em IEEE Robotics and Automation Letters}.

\bibitem[Lynen et~al., 2013]{lynen13robust}
Lynen, S., Achtelik, M., Weiss, S., Chli, M., and Siegwart, R. (2013).
\newblock A robust and modular multi-sensor fusion approach applied to mav
  navigation.
\newblock In {\em International Conference on Intelligent Robots and Systems
  (IROS)}. IEEE.

\bibitem[MAV Control Github, 2017]{control_github}
MAV Control Github (2017).
\newblock Retrieved October 12, 2017, from
  \url{https://github.com/ethz-asl/mav_control_rw}.

\bibitem[MSF Github, 2017]{msf_github}
MSF Github (2017).
\newblock Retrieved July 12, 2017, from
  \url{https://github.com/ethz-asl/ethzasl_msf}.

\bibitem[Mueller et~al., 2015]{mueller2015computationally}
Mueller, M.~W., Hehn, M., and D'Andrea, R. (2015).
\newblock A computationally efficient motion primitive for quadrocopter
  trajectory generation.
\newblock {\em IEEE Transactions on Robotics}, 31(6):1294--1310.

\bibitem[NicaDrone OpenGrab EPM Datasheet v3, 2017]{gripper_datasheet_v3}
NicaDrone OpenGrab EPM Datasheet v3 (2017).
\newblock Retrieved October 09, 2017, from
  \url{http://nicadrone.com/index.php?id_product=66&controller=product}.

\bibitem[Nieuwenhuisen et~al., 2017]{nieuwenhuisen2017collaborative}
Nieuwenhuisen, M., Beul, M., Rosu, R.~A., Quenzel, J., Pavlichenko, D., Houben,
  S., and Behnke, S. (2017).
\newblock Collaborative object picking and delivery with a team of micro aerial
  vehicles at mbzirc.
\newblock In {\em Mobile Robots (ECMR), 2017 European Conference on}, pages
  1--6. IEEE.

\bibitem[Nikolic et~al., 2014]{nikolic2014synchronized}
Nikolic, J., Rehder, J., Burri, M., Gohl, P., Leutenegger, S., Furgale, P.~T.,
  and Siegwart, R. (2014).
\newblock A synchronized visual-inertial sensor system with fpga pre-processing
  for accurate real-time slam.
\newblock {\em Robotics and Automation (ICRA), 2014 IEEE International
  Conference on}, pages {431--437}.

\bibitem[Node Manager, 2017]{node_manager_web}
Node Manager (2017).
\newblock Retrieved July 17, 2017, from
  \url{http://wiki.ros.org/multimaster_fkie}.

\bibitem[Olson, 2011]{olson2011tags}
Olson, E. (2011).
\newblock {AprilTag}: A robust and flexible visual fiducial system.
\newblock In {\em Proceedings of the {IEEE} International Conference on
  Robotics and Automation ({ICRA})}, pages 3400--3407. IEEE.

\bibitem[Piksi Accuracy, 2017]{piksi_accuracy}
Piksi Accuracy (2017).
\newblock Retrieved July 17, 2017, from
  \url{https://support.swiftnav.com/customer/en/portal/articles/2492803-understanding-gps-rtk-technology}.

\bibitem[Piksi Datasheet V2, 2017]{piksi_datasheet_v2}
Piksi Datasheet V2 (2017).
\newblock Retrieved July 12, 2017, from
  \url{http://docs.swiftnav.com/pdfs/piksi_datasheet_v2.3.1.pdf}.

\bibitem[Piksi Github, 2018]{piksi_github}
Piksi Github (2018).
\newblock Retrieved February 02, 2018, from
  \url{https://github.com/ethz-asl/ethz_piksi_ros}.

\bibitem[Rovio Github, 2017]{rovio_github}
Rovio Github (2017).
\newblock Retrieved July 12, 2017, from
  \url{https://github.com/ethz-asl/rovio}.

\bibitem[Scaramuzza and Fraundorfer, 2011]{scaramuzza_tutorial_vio}
Scaramuzza, D. and Fraundorfer, F. (2011).
\newblock Visual odometry [tutorial].
\newblock {\em IEEE Robotics Automation Magazine}, pages {80--92}.

\bibitem[Smach, 2017]{smach_web}
Smach (2017).
\newblock Retrieved July 17, 2017, from \url{http://wiki.ros.org/smach}.

\bibitem[Tagliabue et~al., 2017a]{tagliabue2017robust}
Tagliabue, A., Kamel, M., Siegwart, R., and Nieto, J. (2017a).
\newblock Robust collaborative object transportation using multiple mavs.
\newblock {\em arXiv preprint arXiv:1711.08753}.

\bibitem[Tagliabue et~al., 2017b]{tagliabue2017}
Tagliabue, A., Kamel, M., Verling, S., Siegwart, R., and Nieto, J. (2017b).
\newblock Collaborative object transportation using mavs via passive force
  control.
\newblock In {\em Conference on Robotics and Automation (ICRA)}. IEEE.

\bibitem[Thrun, 2002]{thrun2002particle}
Thrun, S. (2002).
\newblock Particle filters in robotics.
\newblock In {\em Proceedings of the Eighteenth conference on Uncertainty in
  artificial intelligence}, pages 511--518. Morgan Kaufmann Publishers Inc.

\bibitem[USB 3.0 Interference, 2017]{intel_paper_usb3_interference}
USB 3.0 Interference (2017).
\newblock Retrieved July 17, 2017, from
  \url{https://www.intel.com/content/www/us/en/io/universal-serial-bus/usb3-frequency-interference-paper.html}.

\end{thebibliography}

\end{document}